\documentclass[10pt,twocolumn,letterpaper]{article}
\PassOptionsToPackage{numbers,sort,compress}{natbib}
\usepackage[pagenumbers]{cvpr} 
\usepackage{times}
\usepackage{epsfig}
\usepackage{graphicx}
\usepackage{amsmath}
\usepackage{amssymb}

\usepackage[utf8]{inputenc}
\usepackage[T1]{fontenc}
\usepackage{url}
\usepackage{amsfonts}
\usepackage{nicefrac}
\usepackage{microtype}
\usepackage{xcolor}
\usepackage{colortbl}
\usepackage{tabularx}
\usepackage[numbers,compress]{natbib}
\usepackage{multirow}
\usepackage{xspace}
\usepackage{booktabs}
\usepackage{makecell}
\usepackage{caption}
\usepackage{soul}

\definecolor{mylightgray}{gray}{0.88}
\definecolor{mylightcyan}{rgb}{0.86,1.0,1.0}

\definecolor{tblblue}{rgb}{0.1215,0.4666,0.7059}
\definecolor{tblorange}{rgb}{1.0,0.4980,0.0549}
\definecolor{tblgreen}{rgb}{0.1725,0.6275,0.1725}
\definecolor{tblred}{rgb}{0.8392,0.1529,0.1569}
\definecolor{tblpurple}{rgb}{0.5804,0.4039,0.7411}
\definecolor{cvprblue}{rgb}{0.21,0.49,0.74}
\usepackage[pagebackref,breaklinks,colorlinks,bookmarks=false,citecolor=cvprblue]{hyperref}

\newcommand{\ourdatasetfull}{Habitat Synthetic Scenes Dataset\xspace}
\newcommand{\ourdataset}{HSSD\xspace}
\newcommand{\ourwordnet}{WordNetCo\xspace}
\newcommand{\numscenes}{211\xspace}
\newcommand{\numobjects}{18,656\xspace}

\newcommand{\annname}[1]{\texttt{\small #1}}

\newcommand{\goal}[1]{`#1'}

\DeclareMathSymbol{@}{\mathord}{letters}{"3B}

\newcommand{\mypara}[1]{\noindent\textbf{#1}}

\newcommand\best[1]{\textbf{#1}}

\newcommand\stderr[1]{\footnotesize $\pm$ {#1}}

\newcolumntype{Y}{>{\centering\arraybackslash}X}

\newcommand\update[1]{\textcolor{blue}{#1}}
\renewcommand\update[1]{{#1}} 

\def\paperID{6083}
\def\confName{CVPR}
\def\confYear{2024}

\title{\ourdatasetfull (\ourdataset-200):\\An Analysis of 3D Scene Scale and Realism Tradeoffs for ObjectGoal Navigation}

\author{
\textbf{Mukul Khanna$^{1}$\thanks{Equal Contribution}, Yongsen Mao$^{2}$\footnotemark[1] , Hanxiao Jiang$^{2}$ , Sanjay Haresh$^{2}$ , Brennan Shacklett$^{3}$ ,}\\\textbf{Dhruv Batra$^{1,4}$ , Alexander Clegg$^{4}$ , Eric Undersander$^{4}$ , Angel X. Chang$^{2}$ , Manolis Savva$^{2}$}\\
$^{1}$Georgia Tech , $^{2}$Simon Fraser University , $^{3}$Stanford University , $^{4}$Meta AI\\
\url{https://3dlg-hcvc.github.io/hssd/}
}

\begin{document}

\twocolumn[{%
\renewcommand\twocolumn[1][]{#1}%
\maketitle
\begin{center}
\vspace{-0.75cm}
\captionsetup{type=figure}
\includegraphics[width=\linewidth]{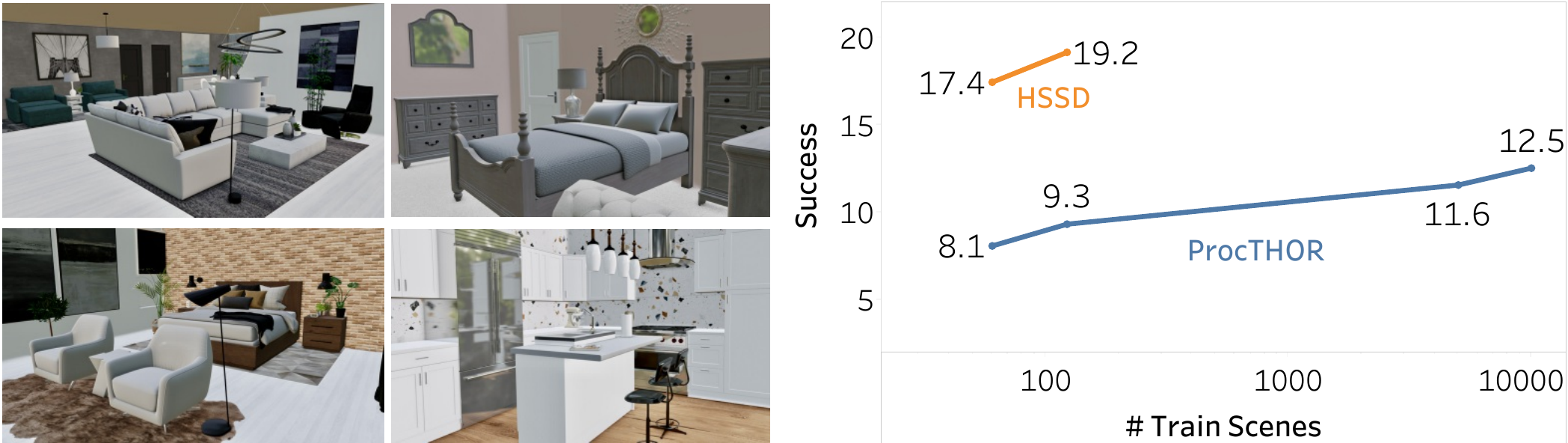}
\captionof{figure}{
\textbf{Left}: we contribute the \ourdatasetfull (\ourdataset-200), a new dataset of high-quality, human-authored synthetic 3D scenes.
\textbf{Right}: zero-shot ObjectNav performance on HM3DSem~\cite{yadav2022hm3dsem} for agents pretained on synthetic 3D scene datasets of different scale and quality.
Through a systematic analysis of scene dataset scale and realism our experiments show that the benefit of dataset scale saturates quickly, and scene realism and quality become the bottleneck for improved ObjectNav agent generalization to realistic scenes.
Concretely, we find that agents trained on 122 scenes from \ourdataset outperform agents trained on two orders of magnitude more scenes from the ProcTHOR~\cite{deitke2022procthor} dataset \update{(19.2 vs 12.5 success rate)}.
}
\label{fig:teaser}
\end{center}
}]

\maketitle

\begin{abstract}
We contribute the \ourdatasetfull (\ourdataset-200), a dataset of 211 high-quality 3D scenes, and use it to test navigation agent generalization to realistic 3D environments. Our dataset represents real interiors and contains a diverse set of \numobjects models of real-world objects. We investigate the impact of synthetic 3D scene dataset scale and realism on the task of training embodied agents to find and navigate to objects (ObjectGoal navigation). By comparing to synthetic 3D scene datasets from prior work, we find that scale helps in generalization, but the benefits quickly saturate, making visual fidelity and correlation to real-world scenes more important. Our experiments show that agents trained on our smaller-scale dataset can outperform agents trained on much larger datasets. Surprisingly, we observe that agents trained on just 122 scenes from our dataset outperform agents trained on 10,000 scenes from the ProcTHOR-10K dataset in terms of zero-shot generalization in real-world scanned environments.
\end{abstract}

\begin{figure*}
\centering
\includegraphics[width=0.328\linewidth]{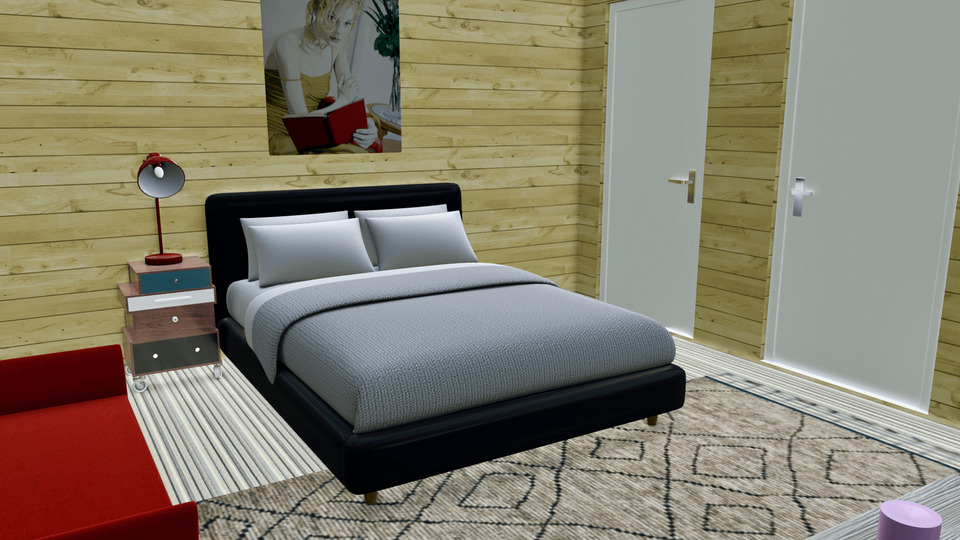}
\includegraphics[width=0.328\linewidth]{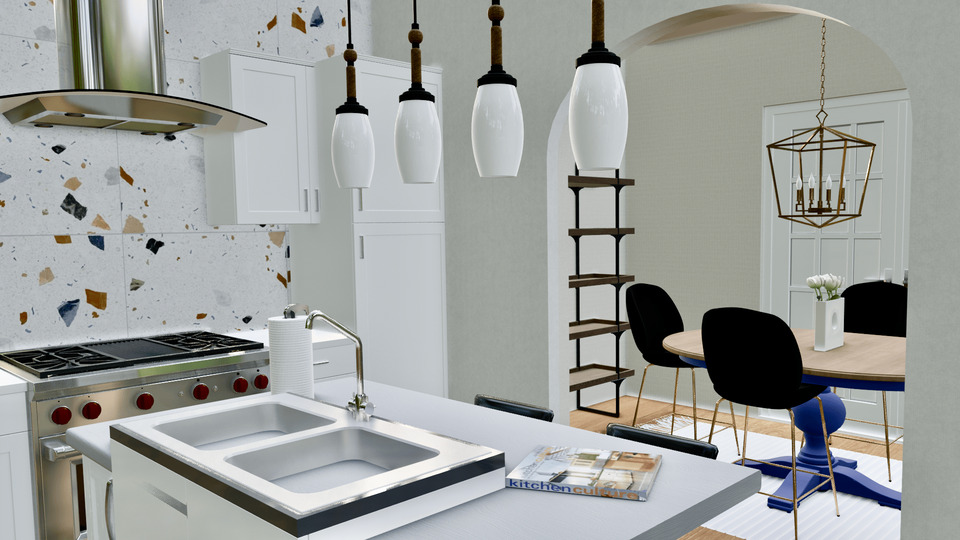}
\includegraphics[width=0.328\linewidth]{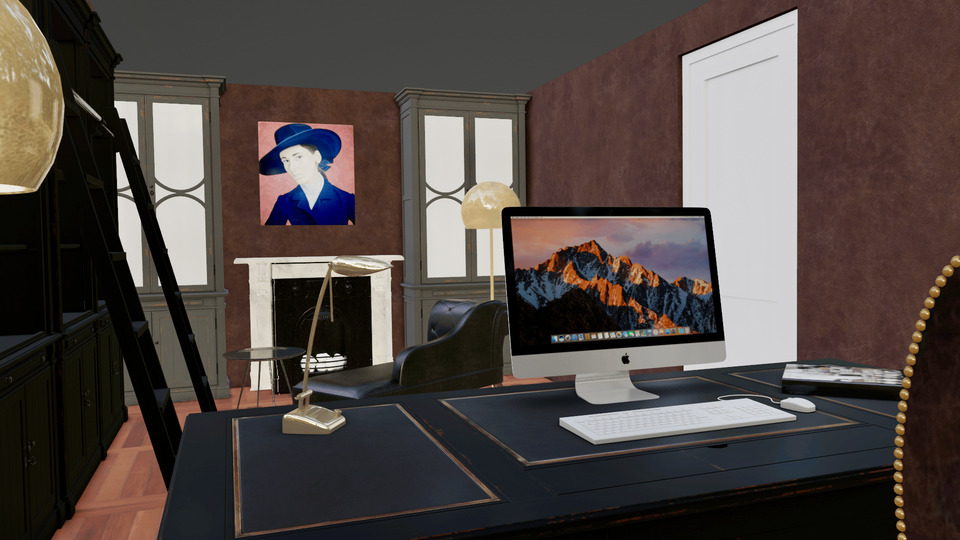}\\[1pt]
\includegraphics[width=0.328\linewidth]{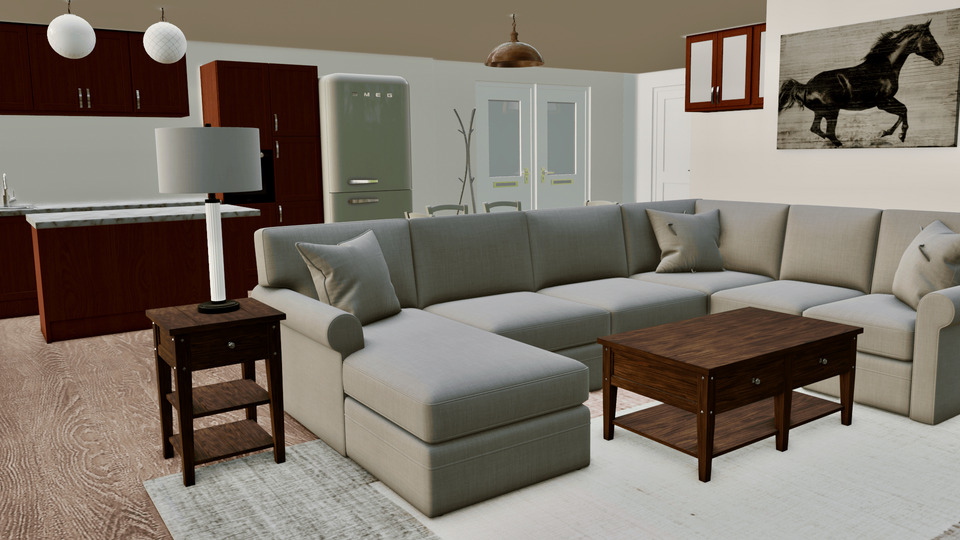}
\includegraphics[width=0.328\linewidth]{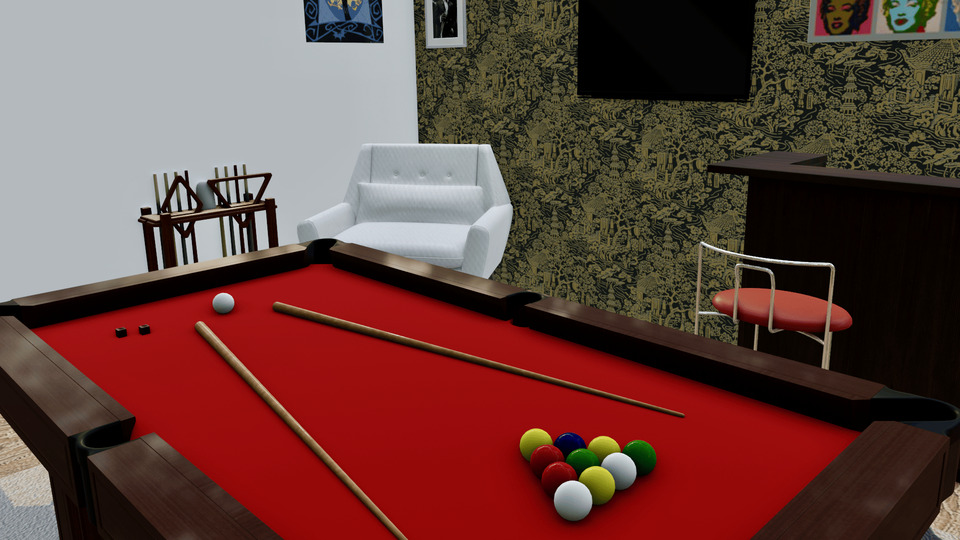}
\includegraphics[width=0.328\linewidth]{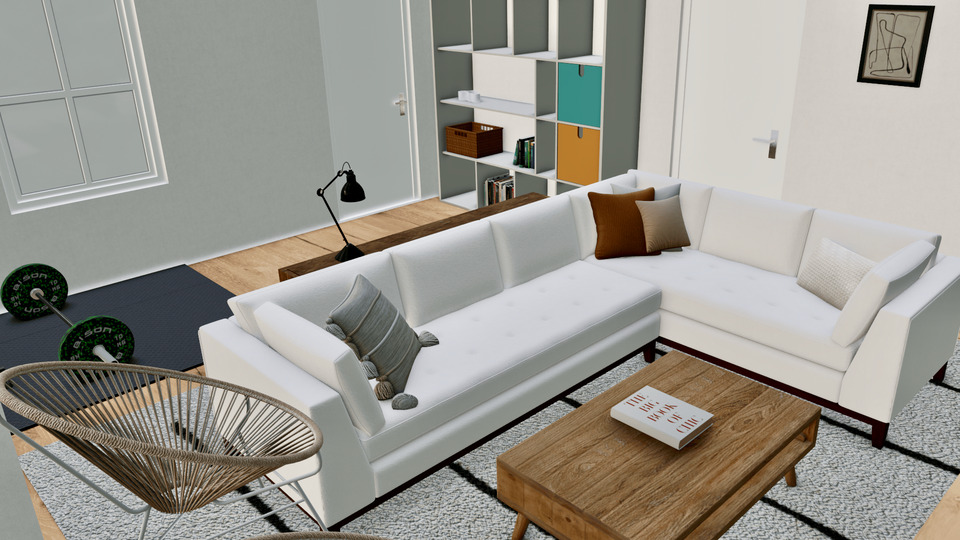}\\[1pt]
\includegraphics[width=0.328\linewidth]{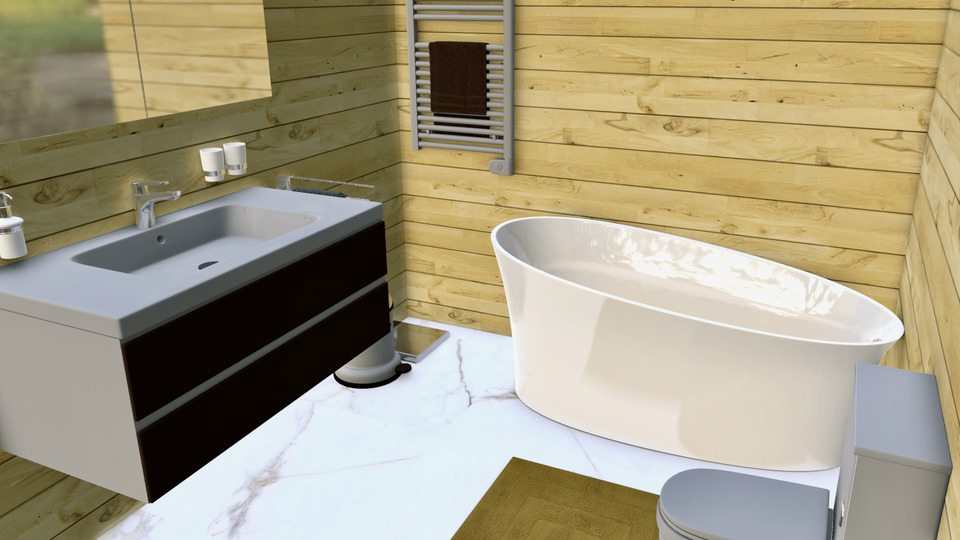}
\includegraphics[width=0.328\linewidth]{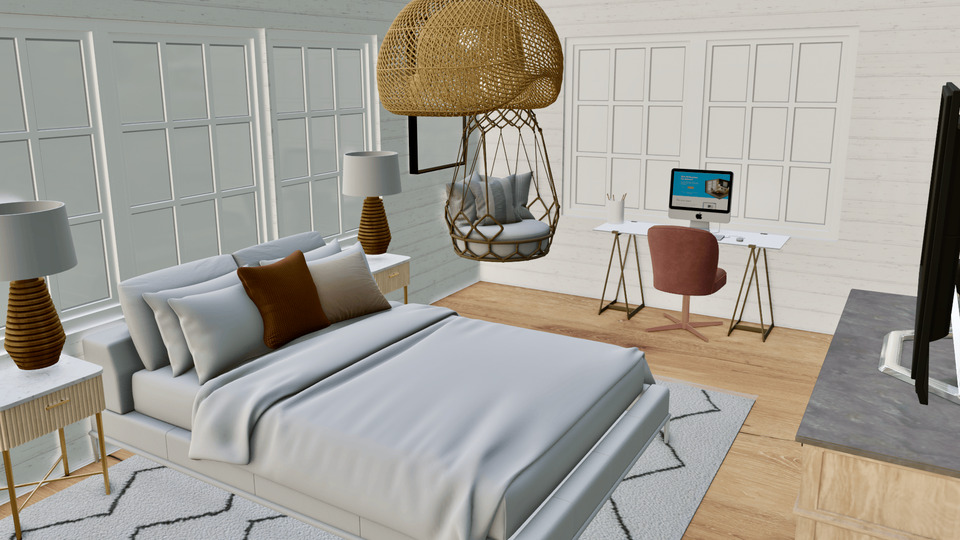}
\includegraphics[width=0.328\linewidth]{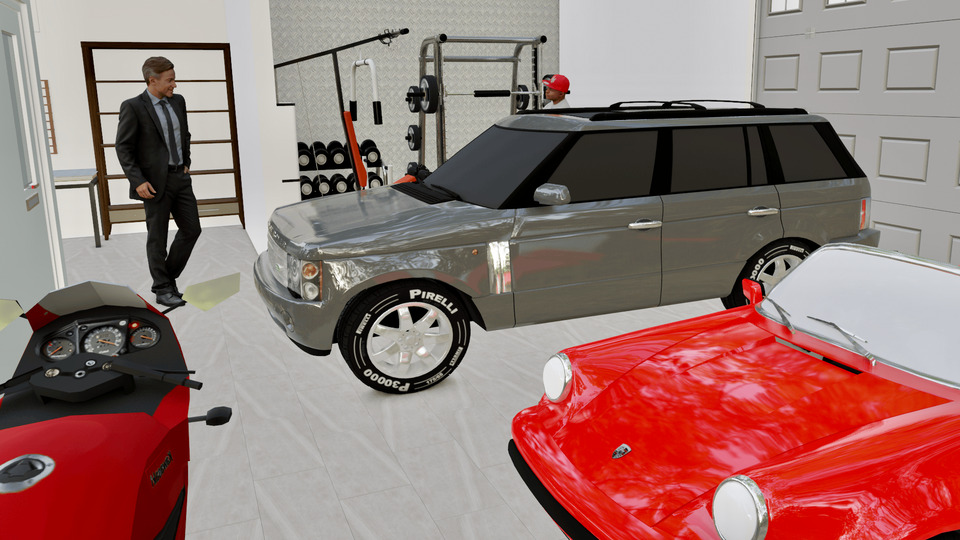}
\caption{
\textbf{Example scenes from \ourdataset-200.}
Our dataset provides high-quality 3D interiors that are fully human-authored. The scenes are densely annotated with object semantic information and assets are prepared to enable performant embodied AI experiments. This dataset will be open-sourced and distributed under a permissive academic research license (free of cost).
}
\label{fig:example-scenes}
\end{figure*}

\section{Introduction}
\label{sec:intro}

Recent years have brought considerable progress in embodied AI agents that navigate in realistic scenes, follow language instructions~\cite{anderson2018vision}, find and rearrange objects \cite{chaplot2020object, szot2021habitat, weihs2021visual, batra2020rearrangement}, and perform other tasks involving embodied sensing, planning, and acting~\cite{chen2020soundspaces, chaplot2020neural, min2022object}.
This progress is supported by simulation platforms that enable systematic, safe, and scalable training and evaluation of embodied AI agents before deployment to the real world~\cite{kadian2020sim2real,truong2021bi,gervet2022navigating,kaufmann2023learning,truong2022rethinking}.

Much of the success of simulation for embodied AI relies on \update{3D scene datasets that mimic the real world.
To this end, the community has leveraged 3D reconstructions and synthetic datasets composed of arrangements of human-designed 3D objects.
Though reconstruction datasets~\cite{matterport3d2017,xiazamirhe2018gibsonenv,ramakrishnan2021hm3d,yadav2022hm3dsem} capture the diversity and complexity of real-world arrangements, the reconstructed scenes are often noisy with missing geometry for thin structures or shiny surfaces.
Other prevalent artifacts include ``holes'' on walls and other surfaces, as well as partially reconstructed object, which can adversely impact the training of agents and lead to overt specialization tailored to artifacts.
In addition, the acquisition and annotation of reconstructions is a significant undertaking and is hard to scale.
Furthermore,} 3D reconstructions are ``monolithic'' scene representations and do not easily allow manipulations such as addition, removal, or state changes of constituent objects (e.g., opening drawers).
Such manipulations are critical in tasks that require interaction with the environment~\cite{batra2020rearrangement}.  \update{This has led to a recent trend in embodied AI to lean more on synthetic 3D scenes which represent real-world environments through composition of human-authored 3D objects~\cite{kolve2017ai2,szot2021habitat,deitke2022procthor}}.

Despite the ubiquitous use of synthetic 3D scene datasets in embodied AI experiments,
there has been no systematic analysis of the tradeoffs between dataset scale (number of scenes and total scene physical size) and dataset realism (visual fidelity and correlation to real-world statistics).
Prior work has largely treated extant datasets as ``black boxes'' for training and evaluation, even in settings where generalization to real-world setups is important.
Moreover, procedural scene generation~\cite{deitke2022procthor} has enabled near-infinite dataset scale but the value of such scale to task performance has not been investigated in a focused manner.

In this paper, we contribute \ourdatasetfull (\ourdataset-200): a human-authored 3D scene dataset that more closely mirrors real scenes than prior datasets.
Our dataset consists of recreations of real houses modeled using a diverse set of \numobjects unique, high-quality 3D models of real objects.
This dataset will be open-sourced and distributed under a permissive academic research license free of charge.
We compare \ourdataset with prior synthetic datasets to show it is closer to real-world scenes in terms of visual fidelity, scene dimensions, and object occurrence statistics.
We then perform a systematic study of scale vs realism with this dataset and other synthetic scene datasets from prior work.
Our experiments show that the smaller-scale but higher-quality \ourdataset dataset leads to ObjectGoal navigation (ObjectNav) agents that \update{outperform} agents trained on significantly larger datasets.
In ObjectNav, an embodied agent is spawned at a random location and orientation and has to efficiently navigate to an instance of a goal category (bed, tv, chair, etc.)~\cite{batra2020objectnav, habitatchallenge2022}.
Surprisingly, we find that we can train navigation agents with better generalization to real-world 3D reconstructed scenes using two orders of magnitude fewer scenes from \ourdataset than from prior datasets.
\Cref{fig:teaser} shows that training on 122 \ourdataset scenes leads to agents that generalize better to HM3DSem~\cite{yadav2022hm3dsem} and MP3D \cite{matterport3d2017} real-world scenes than agents trained on 10,000 ProcTHOR~\cite{deitke2022procthor} scenes.
\update{Beyond training navigation agents, \ourdataset enables work in object manipulation and rearrangement~\cite{yenamandra2023homerobot,homerobotovmmchallenge2023,puig2023habitat}.
This is possible due to the compositionality of synthetic scenes allowing easy removal or addition of objects, operations that are challenging in reconstructions as they require 3D object segmentation and infilling of holes left by moved objects.}

\begin{table*}
\centering
\resizebox{\linewidth}{!}
{
\begin{tabular}{@{} l rrr rrrrrr @{}}
\toprule
 & \multicolumn{4}{c}{Total} & \multicolumn{5}{c}{Average per scene} \\
 \cmidrule(l{2pt}r{2pt}){2-5} \cmidrule(l{2pt}r{0pt}){6-10}
Dataset & Scenes & Objects & Categories & Nav area & Nav area & Nav comp & Clutter & Categories & Instances \\
\midrule
HM3DSem (scans) ~\cite{yadav2022hm3dsem} & 181 & 59,269 & 1,533 & 20.7K & 114.4  & 9.7 & 4.0    & 103.9  & 327.5 \\
MP3D (scans) ~\cite{matterport3d2017} & 90 & 50,851 & 1,658 & 29.6K & 328.4  & 12.8 & 2.8    & 95.5  & 565.0 \\
Gibson tiny (scans) ~\cite{xiazamirhe2018gibsonenv} & 35 & 2,397 & 35 & 4.9K & 139.4  & 8.8 & 4.2    &  15.5  & 68.5  \\
\midrule
ReplicaCAD~\cite{szot2021habitat} & 90 & 92 & 39& 4.5K & 49.8 & 7.2 & 4.8 & 14.3 & 25.5 \\
iTHOR~\cite{kolve2017ai2} & 120 & 3,748 & 112& 2.0K & 17.1 & 2.4 & 9.6 & 28.2 & 46.2 \\
RoboTHOR~\cite{deitke2020robothor} & 75 & 652 & 47& 1.9K & 25.9 & 8.1 & 5.8 & 28.8 & 38.4 \\
ProcTHOR~\cite{deitke2022procthor} & 12K & 1,547 & 95& 808.4K & 67.4 & 10.0 & 5.2 & 40.5 & 74.7 \\
\ourdataset-200 (ours) & 211 & \numobjects & 466 & 53.2K & 252.2 & 13.7 & 5.9 & 61.5 & 329.7  \\
\bottomrule
\end{tabular}
}
\vspace{1pt}
\caption{
\textbf{Scene dataset statistics.}
From left: number of scenes, number of unique objects, number of object categories, total navigable area in m$^2$, average navigable area per scene, navigation complexity as defined by \citet{ramakrishnan2021hm3d} with threshold of points 1m or more apart, mean categories per scene, clutter as defined by \citet{ramakrishnan2021hm3d}, object categories per scene, and instances per scene.
Note that numbers for RoboTHOR and HM3DSem do not include the hidden test sets, and we exclude architectural objects (e.g., walls).
}
\label{tab:dataset_stats}
\end{table*}

\section{Related work}
\label{sec:related}

Progress in embodied AI has been driven by the availability of 3D scene datasets that can be used with simulation platforms to train and evaluate agents.
We focus on analyzing the transfer performance of ObjectNav agents trained on synthetic data to real-world 3D scene scans.

\paragraph{ObjectGoal navigation.}
There are three families of approaches to this task: end-to-end reinforcement learning (RL), imitation learning (IL), and modular learning (ML).
RL methods learn policies that map visual observations directly to action probabilities through hidden states of minimal recurrent neural networks (RNNs) that serve as memory \cite{maksymets2021thda, ye2021auxiliary, mousavian2019visual}.
These methods learn from taking actions and getting rewards based on the progress they make towards the goal object.  On the other hand, IL methods learn navigation from large-scale human demonstrations~\cite{ramrakhya2022habitatweb}.
Modular approaches typically build and leverage scene semantic maps to navigate to the goal object \cite{chaplot2020object, gervet2022navigating}.
Recent work has tackled ObjectNav using modular approaches~\cite{al2022zero} and using self-supervised training~\cite{min2022object}.
We choose ObjectNav as the focus of our investigation since it can serve as a building block for more complex or longer-horizon tasks, including rearrangement~\cite{batra2020rearrangement} where the agent must first navigate in order to manipulate an object or move it to another location.
Moreover, the ObjectNav task requires semantic information about the objects present in the scene and involves reasoning about spatial arrangement patterns of common object categories.
Thus, the characteristics of the 3D scene datasets in terms of visual fidelity and correlation to real-world object occurrence and spatial arrangement patterns are important.

\paragraph{3D scene datasets.}
ObjectNav experiments have been conducted in both scanned real-world environments~\cite{xiazamirhe2018gibsonenv,matterport3d2017,ramakrishnan2021hm3d,yadav2022hm3dsem} and synthetic datasets~\cite{kolve2017ai2,deitke2020robothor,szot2021habitat,deitke2022procthor}.
The performance of ObjectNav agents depends on the scale of the training data as well as the complexity of the environments used for training and testing.
\update{There is a trend of training with increasing number of environments, from small single room environments~\cite{kolve2017ai2} to ever larger multi-room environments~\cite{xiazamirhe2018gibsonenv,matterport3d2017,yadav2022hm3dsem}.
Despite the desire to train on larger datasets, the number of available annotated semantically annotated real-world scans remains limited.
While HM3D~\cite{ramakrishnan2021hm3d} has 1000 scenes, only 216 have been semantically annotated and can be used to train semantically-aware agents.
In addition, \citet{ramakrishnan2021hm3d} pointed out issues with training ObjectNav agents on these scanned environments due to holes in walls, ceilings, and floors.}
Most recently, \citet{deitke2022procthor} demonstrated improved performance on ObjectNav and transfer to real-world scans by leveraging procedurally generated scenes.

Despite this, there has been limited work investigating \update{the scale vs quality tradeoff of synthetic scenes for training ObjectNav agents.
This is partly due to the limited number of synthetic datasets available.}
Existing synthetic scene datasets tend to be either limited to single rooms~\cite{kolve2017ai2}, number of scenes~\cite{deitke2020robothor,szot2021habitat}, or incomplete~\cite{fu20203dfront}.
For instance, 3D-FRONT~\cite{fu20203dfront}, while large, does not have populated kitchens or bathrooms, and has interpenetration issues due to use of algorithmic object asset replacement.
\update{In this work, we investigate aspects of synthetic 3D scene datasets that are important for transfer to real-world scans.
Concretely, we ask the following question for synthetic 3D scene datasets: does large scale suffice, or is realism (match to real-world scenes) also important?
To do so, we contribute \ourdataset-200, a high-quality 3D scene dataset that better matches real-world scenes than prior synthetic datasets.
While the number of scenes is small compared to ProcTHOR, we show that the high quality of this dataset allows for better transfer to navigation in real-world environments.
}

\section{\ourdataset-200: \ourdatasetfull}

To enable our investigation, we develop and contribute a new dataset providing high-quality synthetic 3D interiors.
The \ourdatasetfull (\ourdataset-200) consists of \numscenes houses containing \numobjects objects across 466 semantic categories.
These scenes were designed using the Floorplanner\footnote{\url{https://floorplanner.com} -- data licensed from Floorplanner and made available for academic use.} web interior design interface.
The layouts are predominantly re-creations of real houses by realtors.
Individual objects are created by professional 3D artists and in most cases match specific brands of real-world furniture and appliances.
\ourdataset is distinguished from prior work along several axes: i) high-quality, fully human-authored 3D interiors; ii) fine-grained semantic categorization corresponding to WordNet ontology; iii) asset compression to enable high-performance embodied AI simulation.
See \Cref{fig:example-scenes} for example scenes from the dataset, and \Cref{tab:dataset_stats} for a comparison against other synthetic scene datasets in terms of overall statistics.

The preparation of this dataset involved several stages: object extraction, decomposition, alignment, semantic categorization, and asset compression.
We describe each here.

\mypara{Object extraction.}
The original scenes are exported as a single glTF asset from the Floorplanner database.
We decompose each of these scene assets into constituent objects, as well as architectural elements (walls, floors, ceilings) and openings (doors or windows).
We extract all objects and deduplicate into unique glTF assets to create a shared 3D object model database.
This database of \numobjects objects includes a variety of objects which we semantically annotate.

\mypara{Object decomposition.}
In some cases, a source 3D model of an object represented multiple semantically-distinct objects (e.g., a dining table with chairs, plates, and silverware).
Four graduate students annotated all 3D models into single object or ``multiple object'' classes.
A model has multiple objects if there are distinct nameable components that can be easily detached.
We identified 1,791 such 3D models.
We used an interface based on \citet{mao2022multiscan} web UI to segment these models.
First, we use connected component analysis on the 3D mesh topology to obtain an initial segmentation.
Then, the four annotators decomposed 1,662 models resulting in 11,153 object parts.
We then extracted submeshes, identified duplicate geometry and computed alignment transforms to correspond instances of the same object (e.g., chairs around a table).
We did this by fitting an oriented bounding box (OBB) on each extracted part, initializing alignment using the OBB parameters and then running ICP~\cite{besl1992method} for one iteration.

\mypara{Object alignment.}
All objects including single objects are then aligned to have semantically consistent orientations.
We aligned the objects to have a consistent up and front orientation based on the interface of \citet{mao2022multiscan}.
A total of 2,883 object models required such manual alignment.

\mypara{Semantic categorization.}
Each object was then annotated with a semantic category label.
The labels are from an augmented set of WordNet~\cite{miller1995wordnet} synsets that we call \ourwordnet (WordNet common objects).
WordNet is a popular taxonomy used to organize popular datasets such as ImageNet~\cite{deng2009imagenet} and ShapeNet~\cite{shapenet2015}) but it is lacking in granularity for common objects and modern devices (e.g., iPad, USB stick).
We augmented WordNet with synsets for several common objects (e.g., potted plant, wall lamp) and initialize the object labels by mapping internal tags provided by Floorplanner to \ourwordnet synsets.
We then asked the annotators to manually check, correct, and refine the linked WordNet synsets.
The annotators also specified in what room category a particular object is typically found (e.g., bed in bedrooms).
Unlike other datasets that use heuristics to estimate the real-world size of the objects~\cite{deitke2022objaverse}, our objects are already modeled with real-world dimensions and consistently scaled.

\begin{table}
\centering
\resizebox{\linewidth}{!}
{%
\begin{tabular}{@{} l rrr rrr @{}}
\toprule
 & \multicolumn{3}{c}{ProcTHOR-S (FPS $\uparrow$)} & \multicolumn{3}{c}{ProcTHOR-L (FPS $\uparrow$)} \\ %
\cmidrule(l{2pt}r{2pt}){2-4}
\cmidrule(l{2pt}r{0pt}){5-7}
Simulator & 1 Proc & 1 GPU & 8 GPU & 1 Proc & 1 GPU & 8 GPU \\
\midrule

AI2-THOR~\cite{kolve2017ai2} & \makecell{240\\{\footnotesize $\pm$69}} & \makecell{1,427\\{\footnotesize $\pm$74}} & \makecell{8,599\\{\footnotesize $\pm$359}} & \makecell{115\\{\footnotesize $\pm$19}} & \makecell{6,280$^\ast$\\{\footnotesize $\pm$40}} & \makecell{3,208\\{\footnotesize $\pm$127}} \\
\makecell[l]{Habitat~\cite{savva2019habitat,szot2021habitat}\\(uncompressed)} & \makecell{2,297\\{\footnotesize $\pm$447}} & \makecell{6,374\\{\footnotesize $\pm$798}} & \makecell{57,160\\{\footnotesize $\pm$4,917}} & \makecell{1,007\\{\footnotesize $\pm$187}} & \makecell{5,237\\{\footnotesize $\pm$1,130}} & \makecell{39,510$^\dagger$\\{\footnotesize $\pm$6,345}} \\

\makecell[l]{Habitat~\cite{savva2019habitat,szot2021habitat}\\(compressed)} & \makecell{\best{2,523}\\{\footnotesize $\pm$300}} & \makecell{\best{7,363}\\{\footnotesize $\pm$394}} & \makecell{\best{58,947}\\{\footnotesize $\pm$1,804}} & \makecell{\best{1,233}\\{\footnotesize $\pm$224}} & \makecell{\best{6,508}\\{\footnotesize $\pm$495}} & \makecell{\best{46,674}\\{\footnotesize $\pm$5,640}} \\

\bottomrule
\end{tabular}
}
\vspace{1pt}
\caption{
\textbf{Benchmark of navigation FPS.}
We optimize ProcTHOR scenes similarly to \ourdataset and compare performance when the assets are loaded in Habitat to the original scenes in AI2-THOR.
We observe close to an order of magnitude of improvement of simulation performance across most setups.
$^\ast$After correspondence with \citet{deitke2022procthor}, the authors confirmed this number is a typo and the true number is unknown.
$^\dagger$ProcTHOR-L 8-GPU benchmark conducted with 7 processes per GPU instead of 15 due to memory bottlenecks.
}
\label{tab:fps_benchmark}
\end{table}

\mypara{Asset compression.}
After the above annotation was performed, all object assets were compressed to reduce GPU memory requirements during experimentation.
We performed quadric mesh simplification~\cite{garland1997surface} to an error threshold of $1e^{-3}$, reduced texture resolution to a maximum dimension of $256$, and compressed all textures using the Basis\footnote{\url{https://github.com/BinomialLLC/basis_universal}} supercompression algorithm.
This compression resulted in a 12.4x reduction of on-disk size and comparable on-GPU memory consumption reduction.

\mypara{Comparison with Habitat-optimized ProcTHOR assets.}
To demonstrate the performance benefits of this pipeline we converted and compressed all ProcTHOR~\cite{deitke2022procthor} scenes in the same way as \ourdataset.
\Cref{tab:fps_benchmark} benchmarks these ProcTHOR assets in Habitat and compares with performance of the original assets in AI2-THOR.
We use both the small and large ProcTHOR sets from \citet{deitke2022procthor}, randomly sample 50 scenes from each and take 2K steps in each scene with a random navigation agent, rendering $224\times224\times3$ RGB images. The 1 GPU benchmark uses 15 simulation processes and the 8 GPU benchmark uses 15 processes per GPU.
Benchmarking is done on a server with 8 NVIDIA RTX Quadro 4000 GPUs.
Note that the AI2-THOR benchmark numbers were using RTX Quadro 8000 GPUs which have higher memory and more CUDA cores.
As we can see, the asset compression and the use of the Habitat simulation platform enable much higher performance.
These optimized ProcTHOR assets are independently valuable by enabling faster training and evaluation on other embodied AI tasks.

\section{Dataset analysis}
\label{sec:dataset-analysis}

Before carrying out experiments with \ourdataset, we characterize its properties by comparison against datasets from prior work.
We compare mainly against iTHOR~\cite{kolve2017ai2} and ProcTHOR~\cite{deitke2022procthor}, and use the HM3DSem~\cite{yadav2022hm3dsem} and MP3D \cite{matterport3d2017} real scan datasets as references since they are the basis of our ObjectNav transfer experiments.
We characterize the synthetic datasets and \ourdataset along three axes: \emph{scale}, \emph{realism}, and \emph{complexity}.

\begin{figure}
\includegraphics[width=\linewidth]
{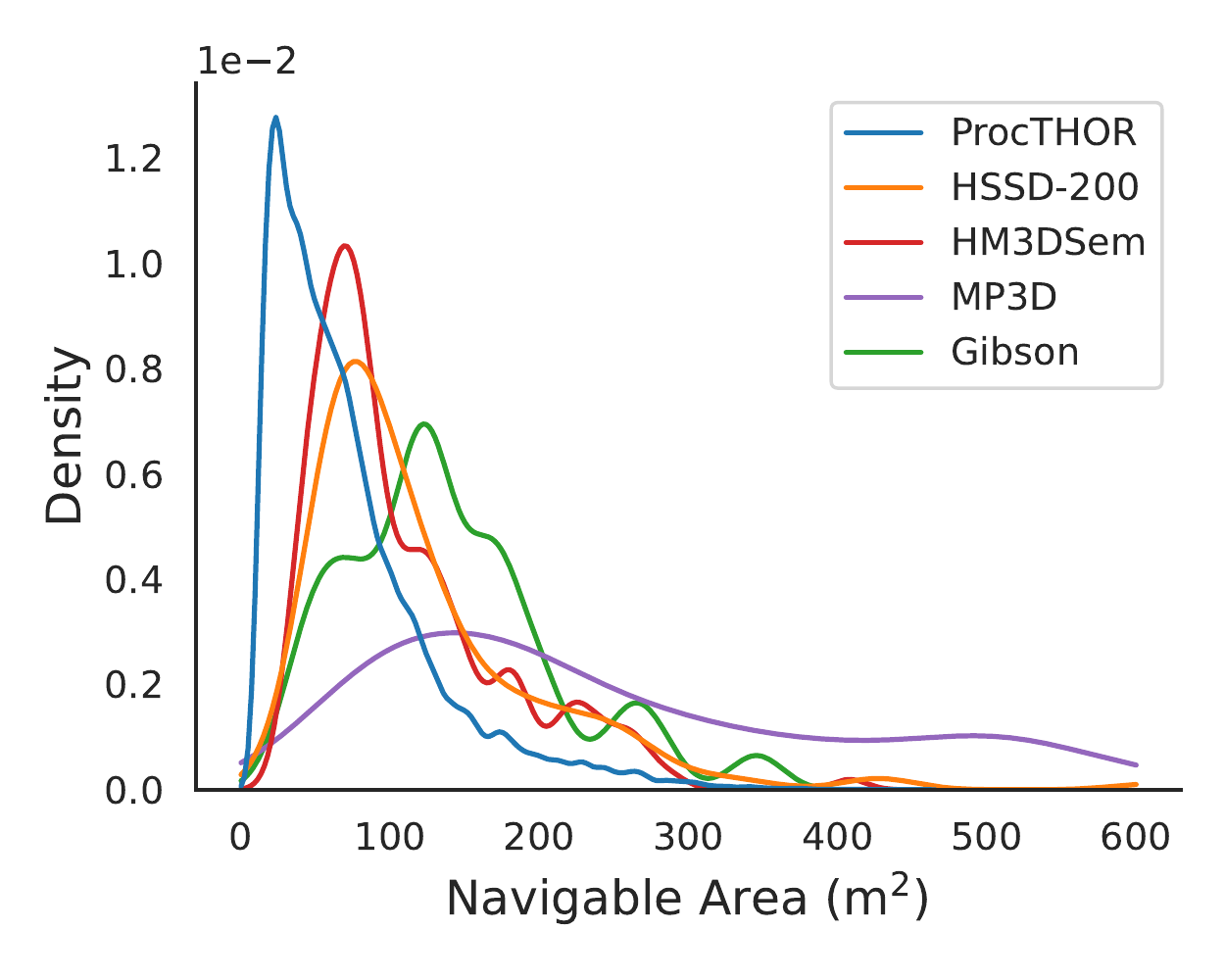} \\
\caption{
\textbf{Per-scene navigable area distribution in ProcTHOR, \ourdataset and HM3DSem.}
\ourdataset (orange) more closely matches real-world scans in HM3DSem, Gibson, and MP3D than ProcTHOR (blue).
\update{Note that the MP3D area distribution is broader as it contains larger-scale public and commercial spaces in addition to residences.}
}
\label{fig:navarea_dist}
\end{figure}

\mypara{Scale.}
The volume of scene data available for experiments is naturally an important characteristic.
We measure scale in terms of scene and object counts, as well as navigable area since that is a key attribute for the ObjectNav task.
\Cref{tab:dataset_stats} summarizes these statistics.
\ourdataset has a relatively small number of scenes but a high number of unique objects (more than 18K), and significantly higher average navigable area per scene (252.2 m$^2$) than prior datasets.
The average number of object categories and object instances per scene is also significantly higher compared to prior synthetic 3D scene datasets, and closer to the real-world scenes in HM3DSem~\cite{yadav2022hm3dsem}, Gibson~\cite{xiazamirhe2018gibsonenv}, and MP3D~\cite{matterport3d2017}.
In \Cref{fig:navarea_dist} we plot the distribution of navigable area per scene for ProcTHOR and \ourdataset, and compare against the distributions for these scan datasets.
We see that ProcTHOR has a high peak and narrow distribution with scenes that have relatively small total navigable areas.
In contrast, \ourdataset matches the per-scene navigable area distribution of the scanned real-world environments more closely.


\begin{table}
\centering
\resizebox{\linewidth}{!}
{
\begin{tabular}{@{} lrrrrrr @{}}
\toprule
 & \multicolumn{2}{c}{HM3D} & \multicolumn{2}{c}{Gibson} & \multicolumn{2}{c}{MP3D} \\
 \cmidrule(l{0pt}r{2pt}){2-3} \cmidrule(l{2pt}r{2pt}){4-5} \cmidrule(l{2pt}r{2pt}){6-7}
Dataset & FID $\downarrow$ & KID $\downarrow$ & FID $\downarrow$ & KID $\downarrow$ & FID $\downarrow$ & KID $\downarrow$ \\
\midrule
ProcTHOR & $88.5$ & $78.0\pm1.6$ & $95.6$ & $82.6\pm1.8$ & 87.2 & 63.7 $\pm1.3$ \\
\ourdataset & \textbf{65.2} & \textbf{57.0} $\pm1.4$ & $\textbf{73.0}$& $\textbf{61.7}\pm1.6$ & \textbf{61.6} & \textbf{43.3} $\pm1.3$ \\
\midrule
HM3D & --- & --- & $18.4$ & $12.7\pm0.8$ & 25.2 & 18.3 $\pm0.7$ \\
Gibson & $18.4$ & $12.8\pm0.8$ & --- & --- & 38.4 & 26.2 $\pm1.2$ \\
MP3D & 25.3 & 18.3$\pm0.7$ & $38.5$ & $26.2\pm1.2$ & --- & --- \\
\bottomrule
\end{tabular}
}
\vspace{1pt}
\caption{
\textbf{Visual fidelity against real images from HM3D, Gibson, and MP3D.}
We render images from ProcTHOR and \ourdataset using Habitat, and compute FID and KID to HM3D and Gibson images.
\ourdataset is closer to the real-world image datasets.
See the supplement for qualitative visuals.
}
\label{tab:visual_fidelity}
\end{table}
\begin{figure*}
\setkeys{Gin}{width=\linewidth}
\begin{tabularx}{\linewidth}{@{}YYYY@{}}
\toprule
 ProcTHOR & \ourdataset & HM3DSem & MP3D \\
\includegraphics[]{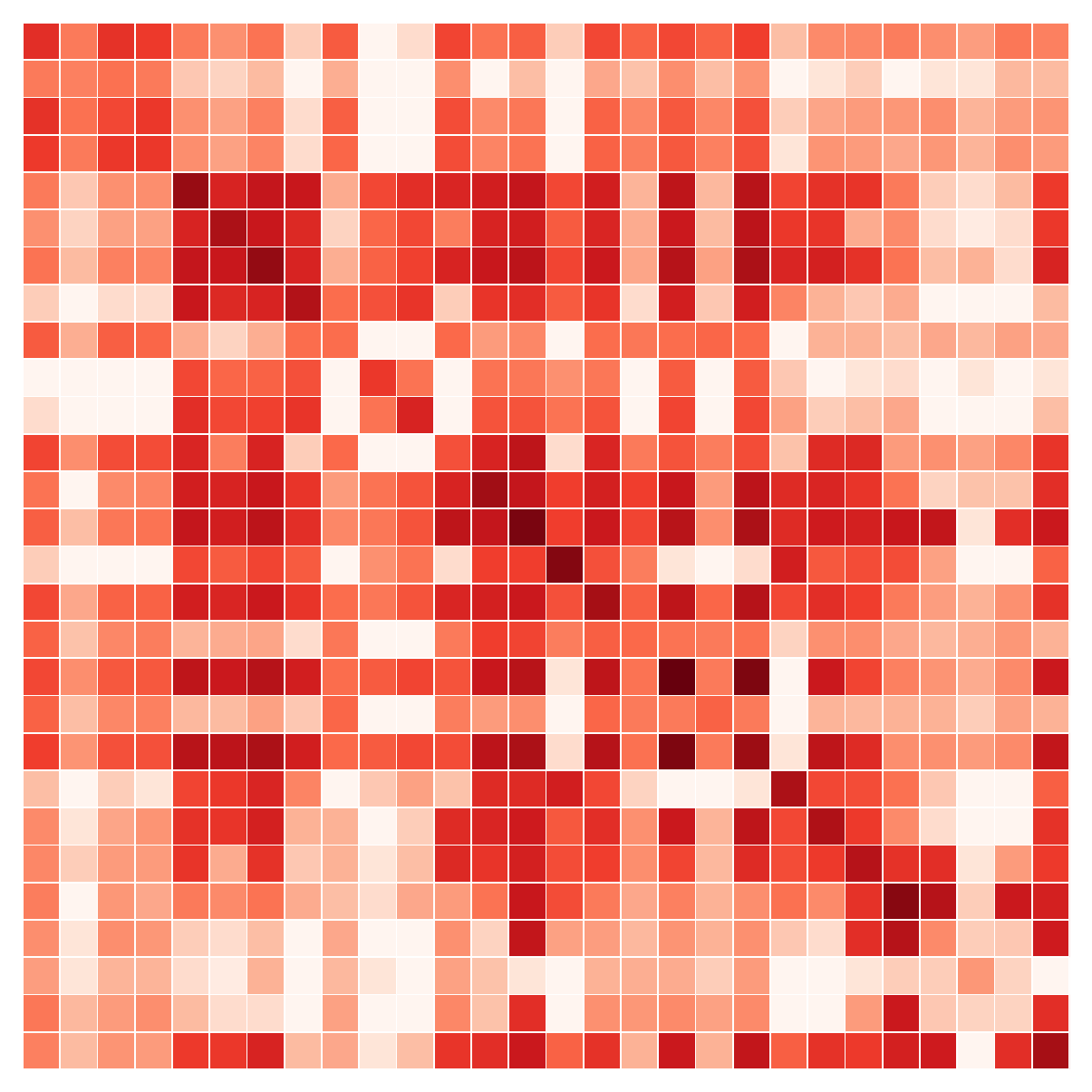} & \includegraphics[]{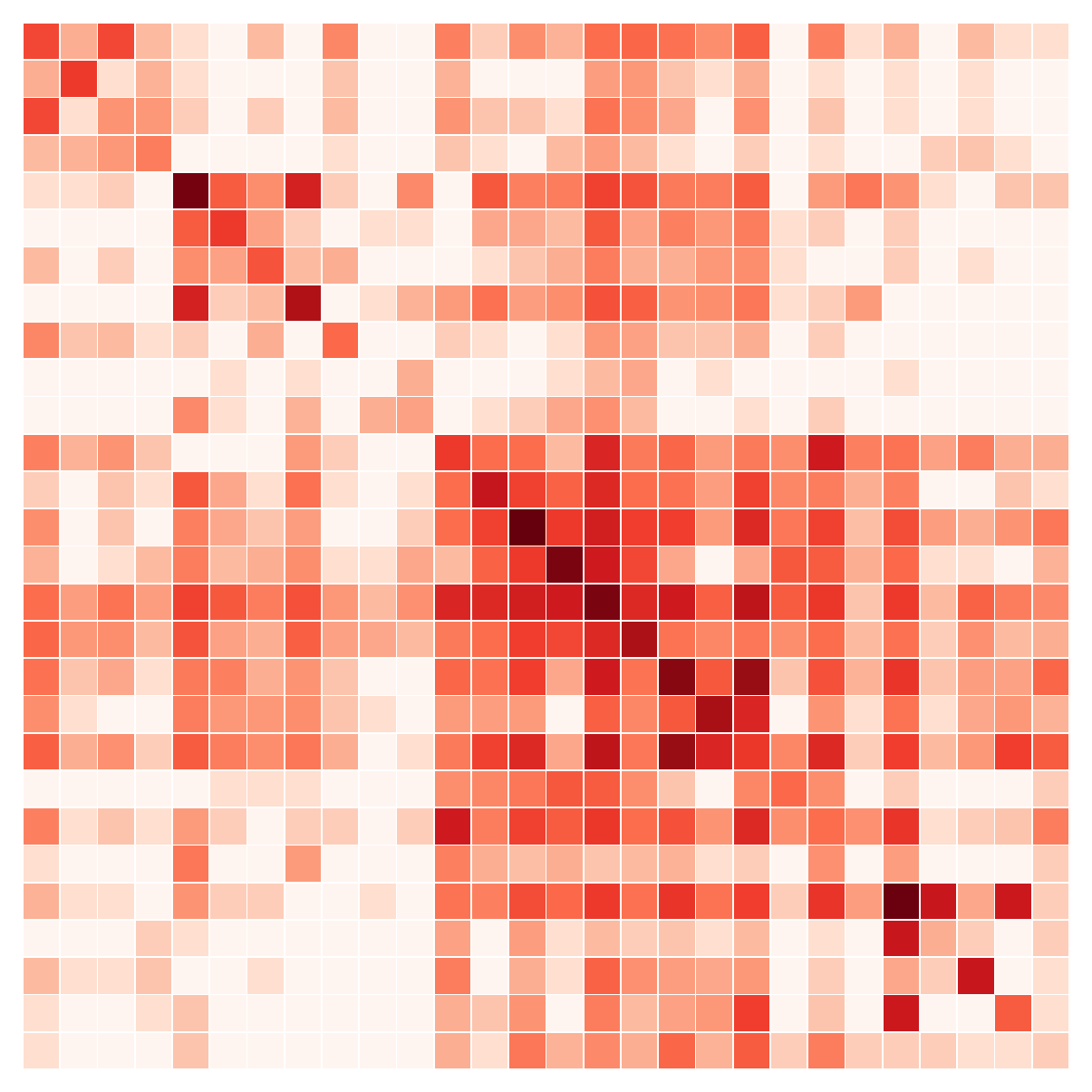} & \includegraphics[]{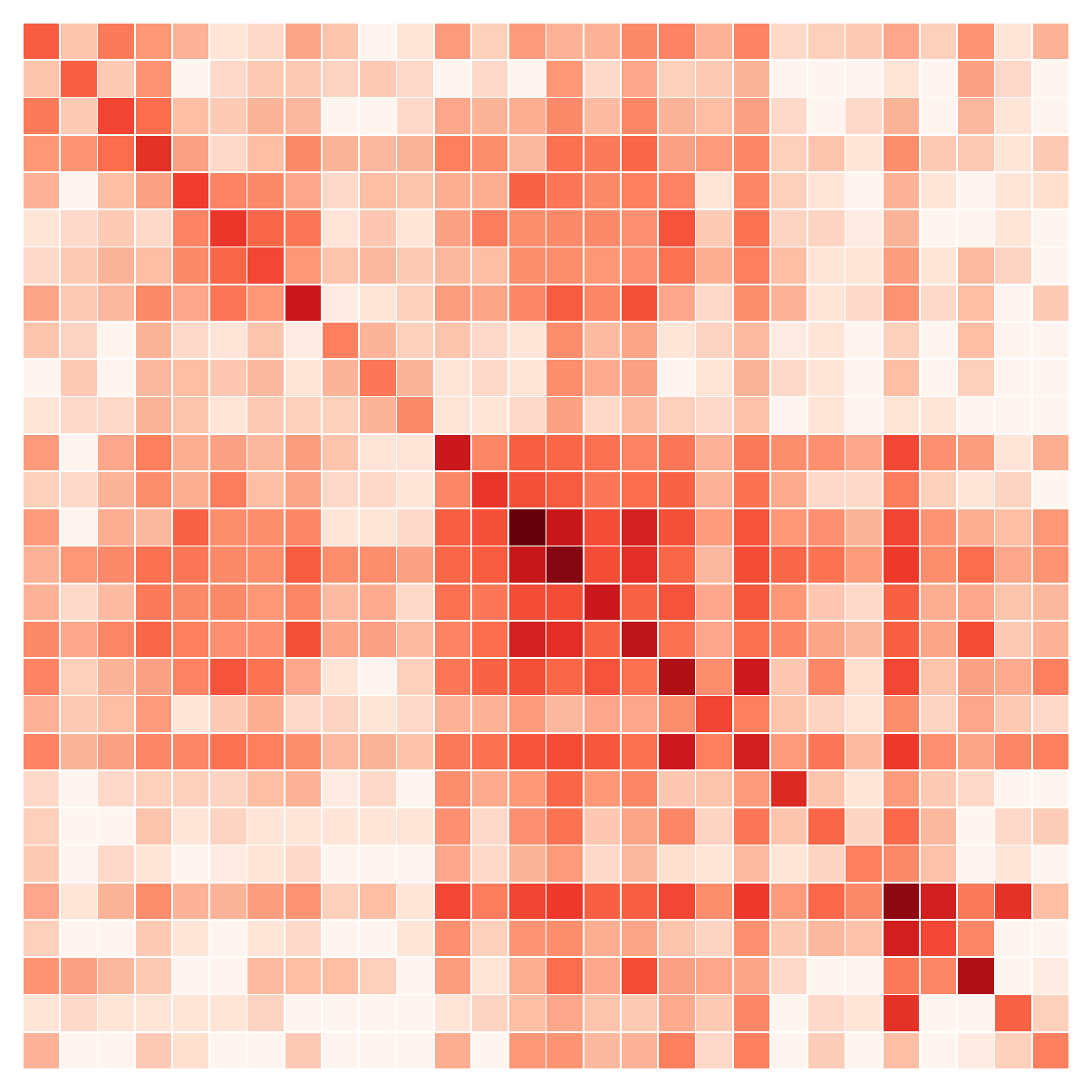} & \includegraphics[]{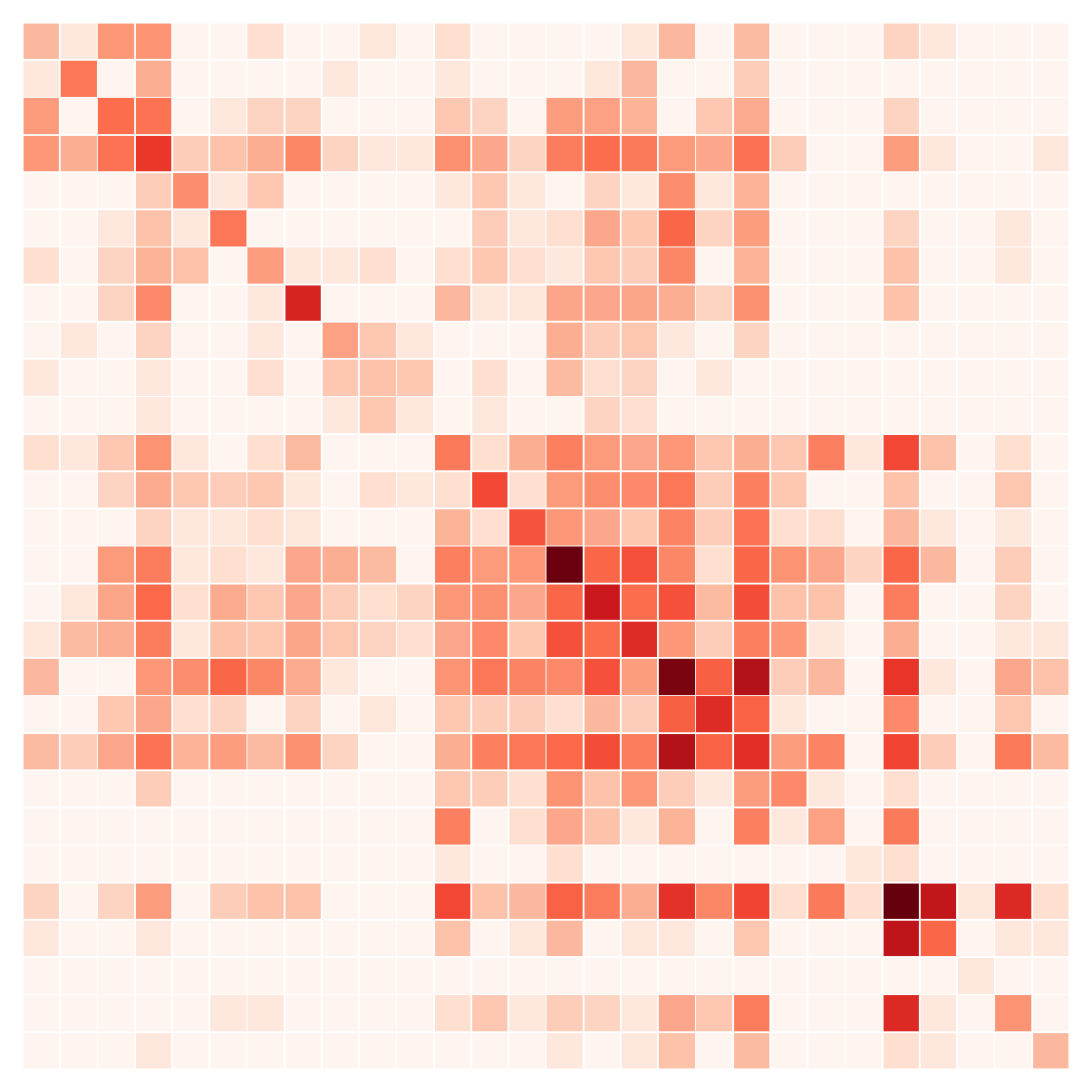} \\
$s_\text{THOR-HM3D}=0.14$ & $s_\text{\ourdataset-HM3D}=0.40$ & $s_\text{HM3D-HM3D}=1.00$ & $s_\text{MP3D-HM3D}=0.72$ \\
$s_\text{THOR-MP3D}=0.07$ & $s_\text{\ourdataset-MP3D}=0.28$ & $s_\text{HM3D-MP3D}=0.45$ & $s_\text{MP3D-MP3D}=1.00$ \\
\bottomrule
\end{tabularx}
\vspace{1pt}
\caption{
\textbf{Object category co-occurrence heatmap visualizations for 28 object categories in ProcTHOR, \ourdataset, and HM3DSem.}
We observe that object co-occurrences in \ourdataset are closer to the real-world HM3DSem scenes relative to ProcTHOR scenes.
We also quantify this observation using a similarity metric measuring the mean Jaccard Index between cluster pairs of hierarchically clustered co-occurrences (higher is better).
\update{The bottom two rows of values report the value of this metric with HM3DSem and MP3D as the reference point.}
}
\label{fig:co-occurrence-ordered}
\end{figure*}


\mypara{Realism.}
The realism of scenes impacts agent perception.
Following \citet{ramakrishnan2021hm3d}, we measure visual realism using the FID~\cite{heusel2017gans} and KID~\cite{binkowski2018demystifying} metrics.
Both metrics measure the perceptual similarity between two distributions of images.
\update{We compare rendered images from the synthetic datasets against images from real scenes.
\Cref{tab:visual_fidelity} shows that \ourdataset is closer to real-world images from HM3D, Gibson~\cite{xiazamirhe2018gibsonenv}, and MP3D~\cite{matterport3d2017}.}
Note that these metrics measure realism of the dataset--renderer combination, and we are using a simple, efficient rasterization-based renderer in Habitat.
Future improvements to rendering quality can lead to improvements in visual fidelity.
Moreover, this measure of visual fidelity does not directly capture the higher-level ``semantic'' realism of the scenes.

To measure semantic realism, we compute object co-occurrence statistics.
We select 28 common object categories found in all datasets (see supplement).
These objects span a range of sizes and vary in their placement on different surfaces (e.g., floor, countertop, tabletop, shelf).
We consider two objects to co-occur if the Euclidean distance of the object centroids is less than 1m.
\Cref{fig:co-occurrence-ordered} shows max-normalized co-occurrence heatmaps for the 28 object categories.
\update{We see that the \ourdataset co-occurrence patterns are qualitatively more similar to real-world scenes from HM3DSem and MP3D than ProcTHOR.
To quantify these trends we also compute overall co-occurrence similarity metrics.
The Spearman's rank correlation coefficient for the pairwise co-occurrences between \ourdataset and HM3DSem is $\rho_\text{\ourdataset-HM3D}=0.219$ compared to $\rho_\text{THOR-HM3D}=0.083$ (higher is better).
The correlation coefficient with MP3D is also higher for our dataset ($\rho_\text{\ourdataset-MP3D}=0.103$) than ProcTHOR ($\rho_\text{THOR-MP3D}=0.017$).}
To further quantify the co-occurrences we also compute a hierarchical clustering on the co-occurrence matrices and extract clusters at threshold 0.8.
\update{We then compute the mean Jaccard Index score between cluster pairs (maximum score pairs chosen with the Hungarian algorithm).
This gives a value of $s_\text{\ourdataset-HM3D}=0.40$ compared to $s_\text{THOR-HM3D}=0.14$ for HM3D, and value of $s_\text{\ourdataset-MP3D}=0.28$ compared to $s_\text{THOR-MP3D}=0.07$ for MP3D (higher is better).}

\mypara{Complexity.}
Most real-world houses are cluttered with furniture items and other objects packed in relatively small spaces.
We characterize this aspect of real-world scene complexity using object and scene statistics.
In \Cref{tab:dataset_stats} we see that \ourdataset has more than 4x object instances per scene than the next highest dataset (329.7 vs 74.7 for ProcTHOR), and about 4x more object categories in total (466 vs 112 for iTHOR).
\update{These statistics are much closer to those of real-world scenes from HM3DSem (327.5 object instances on average and 1,533 categories total) and MP3D (565 object instances on average and 1,658 categories total).}

\section{Experimental setup}

We investigate the role of dataset scale and realism in training and evaluating agents for ObjectNav, focusing on the transfer setting.

\mypara{Task.}
We adopt the Habitat ObjectNav 2022 challenge setup~\cite{habitatchallenge2022}, which we briefly summarize here.
There are 6 goal categories: \goal{bed}, \goal{chair}, \goal{sofa}, \goal{tv}, \goal{plant}, \goal{toilet}.
The agent is successful if it predicts STOP action within 0.1m of a ``viewpoint'' around each goal.
The agent is a LoCoBot with base radius of 0.18m and height of 0.88m and possesses an RGB sensor plus a compass and GPS sensor.
The agent action space consists of: \annname{STOP}, \annname{MOVE\_FORWARD}, \annname{TURN\_LEFT}, \annname{TURN\_RIGHT}, \annname{LOOK\_UP}, \annname{LOOK\_DOWN}, with forward step of 0.25m and turn angle of 30 degrees.

\begin{figure}
\centering
\setkeys{Gin}{width=\linewidth}
\noindent
\includegraphics[]{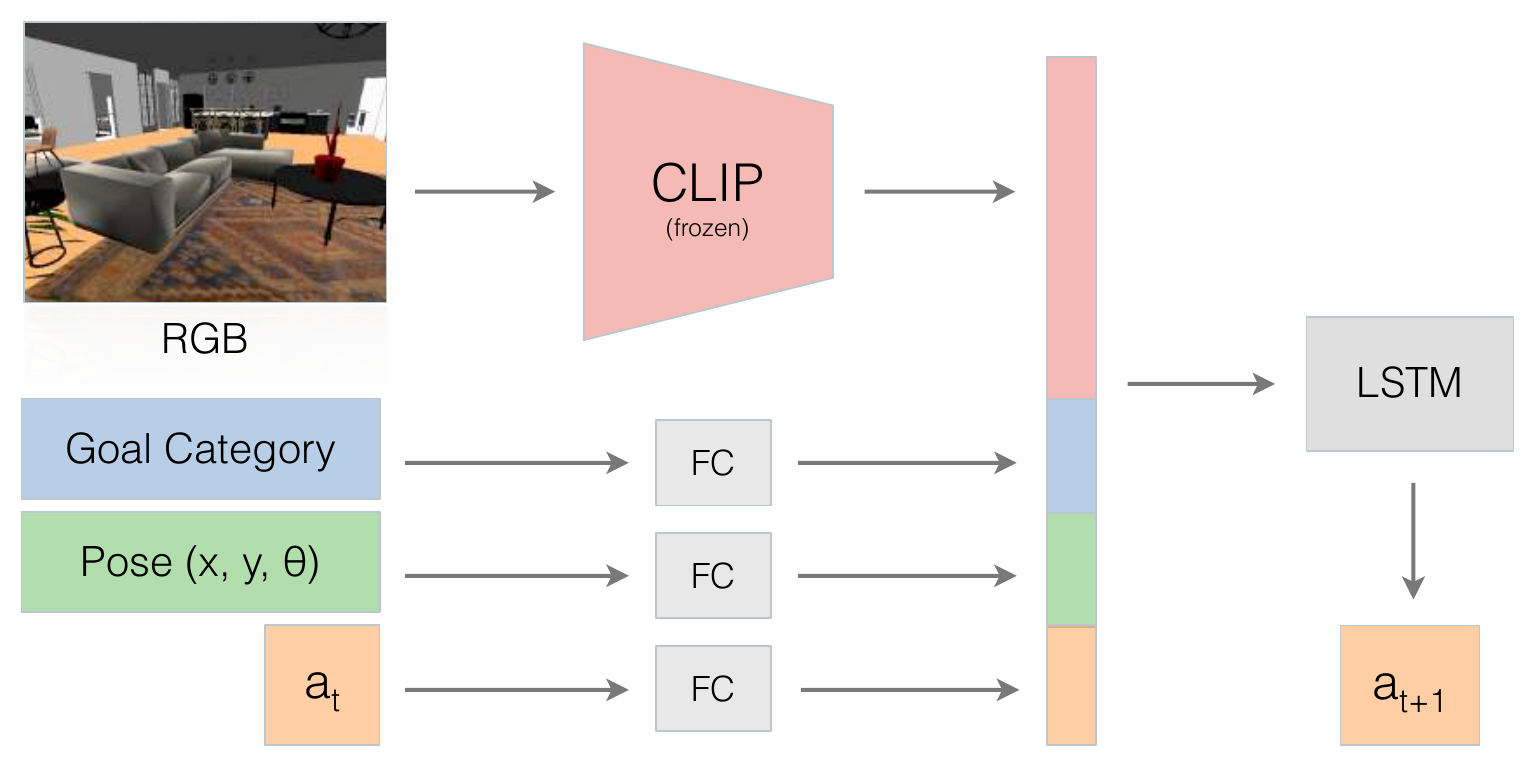} 
\caption{
\textbf{Architecture of the ObjectNav agent.}
The RGB observation is passed into a frozen CLIP backbone to produce a feature embedding.
The goal category, agent pose and previous action are embedded into 32-dim vectors.
All embeddings are fed to an LSTM to predict the next action.
}
\label{fig:arch_rl}
\end{figure}

\begin{table}

\centering
\resizebox{\linewidth}{!}
{%
\begin{tabular}{@{} l ccc ccc @{}}
\toprule
& \multicolumn{3}{c}{train} & \multicolumn{3}{c}{val} \\ 
Dataset & S & E/S & T & S & E/S & T\\
\midrule
iTHOR~\cite{kolve2017ai2} & 64 & 2K & 128K & 13 & 30 & 390 \\ ProcTHOR~\cite{deitke2022procthor} & 10K & 1K & 10M & 1K & 10 & 10K \\ 
\ourdataset & 122 & 1K & 122K & 40 & 30 & 1.2K \\
\update{MP3D} \cite{matterport3d2017} & 56 & 16.5K & 925K  & 10 & 65.8 & 658 \\
HM3DSem~\cite{yadav2022hm3dsem} & 145 & 50K & 7.25M & 36 & 30 & 1.08K \\
\bottomrule
\end{tabular}%
}
\vspace{1pt}
\caption{
\textbf{Train/val scenes (S), episodes-per-scene (E/S) and total (T) episode statistics.}
\update{We generate ObjectNav episodes for the 6 object categories used in our experiments across available train and val set scenes in each dataset.}}
\label{tab:split-stats}
\end{table}

\mypara{Agent architecture.}
At each step, the agent has access to RGB observations, GPS and compass sensors, the object category of the goal to navigate to as input.
The RGB frames are of resolution $224 \times 224$.
The GPS and compass sensors are specified as in \citet{yadav2022hm3dsem}.
The input goal is an integer $o$ corresponding to the object category and used to index into an embedding matrix $O$ to produce a 32-dimensional embedding.
We use an architecture similar to \citet{khandelwal2022simple} (see \Cref{fig:arch_rl}).
The RGB frames are passed through a frozen CLIP pre-trained ResNet 50 visual encoder to get a $2048$ dimension feature vector.
The agent also takes the previous action as input.
Both the goal and action embedding matrices ($O, A$) are learned during training.
These embeddings are concatenated along with the outputs from the GPS and compass sensors and passed through a 2-layer LSTM network.
The LSTM outputs are then passed through a linear layer to produce action probabilities for the next step.
We train agents using DDPPO~\cite{wijmans2019dd} with VER~\cite{wijmans2022ver} on 4 NVIDIA A40 GPUs and 24 environment workers per GPU.


\begin{table*}
\centering
\resizebox{1\linewidth}{!}{
{%
\newcolumntype{g}{>{\columncolor{mylightgray}}c}
\begin{tabular}{@{} lcccccc gggg}
\toprule
Eval dataset $\rightarrow$ & \multicolumn{2}{c}{iTHOR} & \multicolumn{2}{c}{ProcTHOR} &
  \multicolumn{2}{c}{\ourdataset} & \multicolumn{2}{c}{MP3D} & \multicolumn{2}{c}{HM3DSem} \\
\cmidrule(l{0pt}r{2pt}){2-3} \cmidrule(l{2pt}r{2pt}){4-5} \cmidrule(l{2pt}r{2pt}){6-7} \cmidrule(l{2pt}r{0pt}){8-9} \cmidrule(l{2pt}r{0pt}){10-11}
Train dataset $\downarrow$ & Success $\uparrow$ & SPL $\uparrow$ & Success $\uparrow$ & SPL $\uparrow$ & Success $\uparrow$ & SPL $\uparrow$ & Success $\uparrow$ & SPL $\uparrow$ & Success $\uparrow$ & SPL $\uparrow$ \\
\midrule

iTHOR & 78.06 \stderr{1.31} & 53.05 \stderr{0.59} & 29.8 \stderr{0.08} & 15.61 \stderr{0.08} & 13.43 \stderr{0.29} & 4.75 \stderr{0.11} & 6.18 \stderr{0.41} & 2.13 \stderr{0.27} & 6.16 \stderr{0.19} & 2.38 \stderr{0.22} \\

ProcTHOR-10K & 67.04 \stderr{4.31} & 36.32 \stderr{4.24} & 80.72 \stderr{0.43} & 46.44 \stderr{0.33} & 31.54 \stderr{1.42} & 12.94 \stderr{0.7} & 8.26 \stderr{0.73} & 2.96 \stderr{0.46} & 12.53 \stderr{0.49} & 5.26 \stderr{0.2} \\

\ourdataset-122 & 27.5 \stderr{3.36} & 12.34 \stderr{2.08} & 9.57 \stderr{0.5} & 3.86 \stderr{0.24} & 54.81 \stderr{0.28} & 24.12 \stderr{0.14} & \best{14.44 \stderr{1.29}} & \best{5.17 \stderr{0.48}} & \best{19.15 \stderr{0.95}} & \best{7.71 \stderr{0.47}}  \\

\midrule

MP3D & & & & & & & 32.07 \stderr{1.22} & 14.73 \stderr{0.99} & 31.95 \stderr{1.17} & 13.13 \stderr{0.45} \\
HM3DSem &  & & & & & & 30.8 \stderr{1.82} & 13.99 \stderr{0.89} & 48.1 \stderr{1.54} & 22.16 \stderr{0.05} \\


\bottomrule
\end{tabular}
}}
\caption{
\textbf{ObjectNav zero-shot generalization across datasets.}
Agents are trained on the training set of the dataset indicated in each row and evaluated on the val set of the datasets in the columns.
We report the average across three independent training runs, and the standard error on this average.
As expected, agents perform well when evaluated on the dataset on which they were trained.
When looking at the generalization trends, we observe that surprisingly \ourdataset-122 achieves better generalization performance on both MP3D and HM3DSem compared to agents trained with the much larger scale ProcTHOR-10K.
}
\label{tab:zero_6_28}
\end{table*}

\mypara{Episode dataset generation.}
\update{
We split \ourdataset into train/val/test following a ratio 60/20/20 ratio.
We use the standard splits for ProcTHOR~\cite{deitke2022procthor}, and HM3DSem~\cite{yadav2022hm3dsem}.
For iTHOR~\cite{kolve2017ai2}, we start with the standard split (80 train, 20 val scenes) but filter out scenes that do not contain any of the 6 goal categories or are too small for navigation.
Following prior work, we generate 2K training episodes per scene for this dataset~\cite{kolve2017ai2}.
For ProcTHOR and \ourdataset, we use 1K training episodes per scene.
For HM3DSem, we follow prior work and use 50K training episodes per scene for the 145 HM3DSem train scenes.
For MP3D, we again follow prior work but remove episodes with target objects that do not belong to the 6 goal categories.
This results in a dataset of 56 scenes with around 16.5K episodes each for training and 10 scenes with 658 total episodes for validation.}

For the val set we use 30 episodes per scene in \ourdataset, iTHOR, and HM3DSem scenes, and 10 episodes per scene for ProcTHOR (see \Cref{tab:split-stats}).
The total number of episodes per scene are uniformly divided across all object categories and object instances within each category in each scene.
See the supplement for details on episode generation. 

\section{Results}

Our goal is to compare and contrast agents in terms of generalization performance to HM3DSem and MP3D scenes when trained on synthetic 3D scene datasets of different scale and realism.

We evaluate generalization of agents trained on iTHOR~\cite{kolve2017ai2}, ProcTHOR-10k~\cite{deitke2022procthor}, and \ourdataset.
\update{We also train agents on HM3DSem and MP3D to provide a comparison point of agents trained on reconstruction scene datasets.}
The trained agents for each dataset are then evaluated on the validation sets of all datasets.

\mypara{Zero-shot generalization.}
\update{We train agents until convergence on the train set and report the performance of the checkpoints with highest val set SPL averaged across three training runs in \Cref{tab:zero_6_28}. The supplement provides training plots for these experiments.} 
As expected the best performance in most cases is achieved by the agent trained on the same dataset.
Note that iTHOR and ProcTHOR scenes use the same object \update{assets} and therefore agents tend to transfer relatively well between them.
\update{Despite the much smaller overall dataset size of \ourdataset compared with ProcTHOR-10K, zero-shot evaluation on HM3DSem results in higher success (19.15\% vs 12.53\%) and SPL (7.71 vs 5.26).
Similarly, zero-shot evaluation on MP3D scenes also results in higher success (12.56\% vs 8.26\%) and SPL (4.56 vs 2.96).}
This finding indicates that the much higher scale of ProcTHOR-10K does not translate to improved zero-shot ObjectNav agent generalization.

\begin{figure*}[t]
\includegraphics[width=\linewidth]{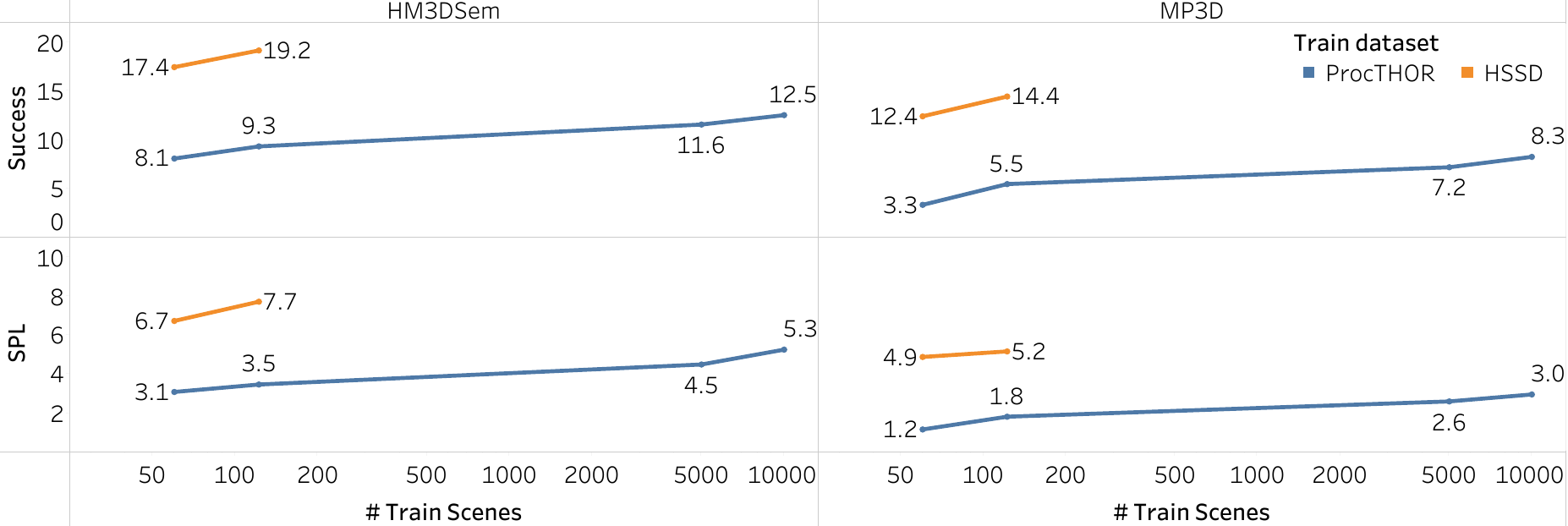}
\caption{
\textbf{ObjectNav zero-shot generalization performance for agents trained with different dataset scales.}
The plots show zero-shot performance of agents on the HM3DSem and MP3D val set.
We see that agents trained on \ourdataset-60 and \ourdataset-122 outperform agents trained on ProcTHOR-60 and ProcTHOR-122.
Moreover, the much larger ProcTHOR-10K dataset does not lead to significant improvements in ObjectNav generalization performance, with agents trained on it generalizing less well than ones trained with far fewer HSSD scenes.
}
\label{fig:dataset_scaling}
\end{figure*}

\mypara{Fine-tuned agents.}
\update{
In addition to the zero-shot generalization performance of agents, we carried out a number of experiments where agents were fine-tuned on the target dataset's train split prior to evaluation on that dataset's val split.
As expected, after finetuning the differences between agents initialized from different training sets are reduced, and agents converge to more similar levels of performance on the target dataset.
After finetuning, agents trained on \ourdataset-122 achieve 48.23\% success and 23.1 SPL compared to ProcTHOR-10K agents achieving 48.32\% success and 21.8 SPL.
The gap in combined efficiency and success (as measured by SPL) remains but is smaller compared to the zero shot setting.
This finding stands in contrast to the appreciable gap in the zero shot generalization performance which is more representative of real world deployment to previously unseen environments.
See the supplemental materials for a more complete summary of these fine-tuning experiments.
}

\mypara{Disentangling scene dataset scale and realism.}
To disentangle the roles of scene dataset scale and scene dataset realism in agent generalization we construct scale-matched datasets for \ourdataset and ProcTHOR by varying the number of scenes and total navigable area.
We consider different total dataset scales for \ourdataset (60 scenes, and all 122 training scenes) and ProcTHOR (60, 122, and all 10K scenes).
We select the 60 and 122 ProcTHOR scene subsets such that the distribution of their navigable area is similar to \ourdataset.
Note that all the aforementioned datasets have 1K episodes per scene.
We refer to these subset as ProcTHOR-60 and ProcTHOR-122 (see supplement for details).
Then, we compare the zero-shot generalization performance of agents on the MP3D and HM3DSem val sets.
All agents are trained until convergence (i.e. until training and validation success saturate).
The results are
plotted in \Cref{fig:dataset_scaling}.
We see that agents trained on \ourdataset-60 and \ourdataset-122 outperform agents trained on the scale-matched ProcTHOR-60 and ProcTHOR-122 in terms of both success rate and SPL.
\update{The difference in the 122 scene setting is particularly pronounced with \ourdataset-122 vs ProcTHOR-122 agents achieving 19.15 vs 9.33 success and 7.71 vs 3.47 SPL, respectively on the HM3DSem val set (see \Cref{fig:dataset_scaling} left column).
The trend is even more pronounced when evaluating generalization to the MP3D val set: 14.44 vs 5.47 success and 5.17 vs 1.83 SPL, for \ourdataset-122 vs ProcTHOR-122 agents respectively (see \Cref{fig:dataset_scaling} right column).
}
\update{Note that agents trained on ProcTHOR-10K perform only marginally better than agents trained on the much smaller ProcTHOR-122 indicating the limited value of larger scale by itself.}

\mypara{Limitations.}
Our investigation is limited to one type of agent: monolithic pixels-to-actions agents trained end-to-end in an RL fashion.
A broader investigation including agents designed using modular approaches~\cite{gervet2022navigating} or imitation learning~\cite{ramrakhya2022habitatweb} would offer insights into the comparative trends between these families of approaches.

\section{Conclusion}

Our goal was to investigate the impact of scene scale and realism on the generalization ability of ObjectGoal navigation agents.
To this end, we constructed \ourdataset: a high-quality, human-authored synthetic 3D scene dataset.
We carried out a systematic analysis of how agents trained in this scene dataset and other synthetic 3D scene datasets from prior work generalize to realistic 3D scenes.
We found that a smaller number of higher-quality synthetic 3D scenes leads to better generalization compared to larger numbers of procedurally generated 3D scenes or lower-quality scenes.
We hope our dataset and our findings on the tradeoffs between scale and realism in synthetic 3D scene datasets help to enable future work on Embodied AI agents for visual navigation and related tasks.

\mypara{Acknowledgments.}
The research team members at SFU were supported by a Canada CIFAR AI Chair grant, a Canada Research Chair grant, an NSERC Discovery Grant and a research grant by Meta AI. Experiments at SFU were enabled by support from the \href{https://alliancecan.ca/en}{Digital Research Alliance of Canada}.
We thank Karmesh Yadav, Chris Paxton, Mrinal Kalakrishnan, Sonia Raychaudhuri, Qirui Wu, and Xiaohao Sun for useful discussions and feedback on early drafts of this paper. We also thank Ram Ramrakhya for useful discussions and help with the ProcTHOR experiments, and John Turner and Vladim\'{i}r Vondru\v{s} for help with 3D asset compression.

{\small
\bibliographystyle{plainnat}
\setlength{\bibsep}{0pt}
\bibliography{main}
}

\clearpage
\newpage

\appendix

\begin{figure*}
\centering
\includegraphics[width=\linewidth]{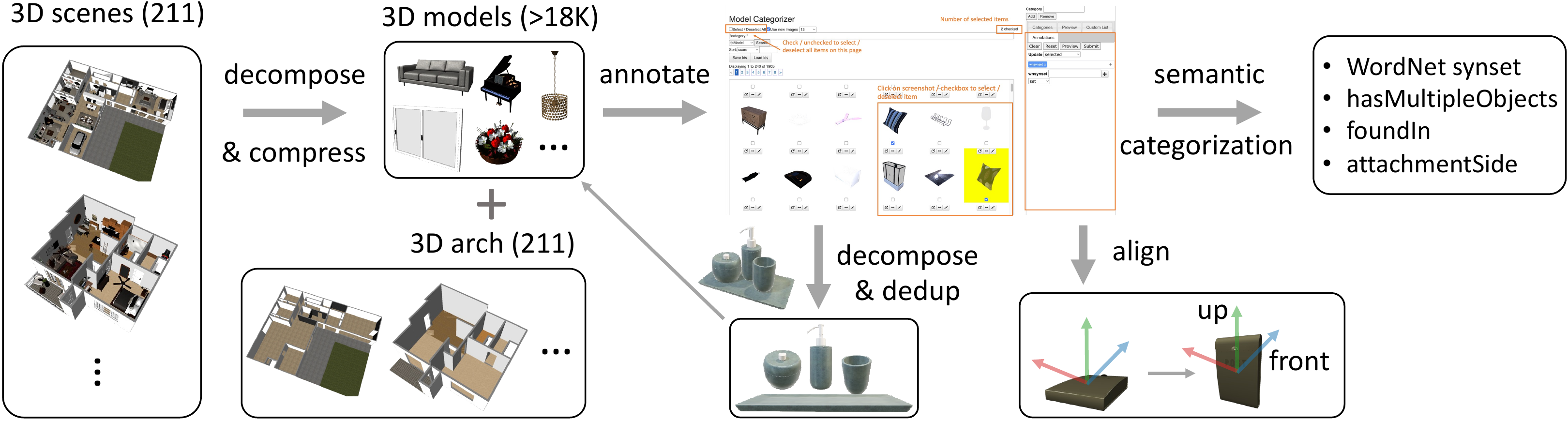}
\caption{
\textbf{Overview of \ourdataset construction pipeline.}
We first decompose the original 211 3D scenes into more than 18K individual per-object 3D models, and architectural geometry for each scene.
The per-object models are then annotated with a variety of semantics including WordNet synsets.
The objects are also decomposed and semantically aligned so that they have a consistent up and front orientation.
}
\label{fig:fp-data-pipeline}
\end{figure*}

\begin{figure*}
\centering
\includegraphics[width=\linewidth]{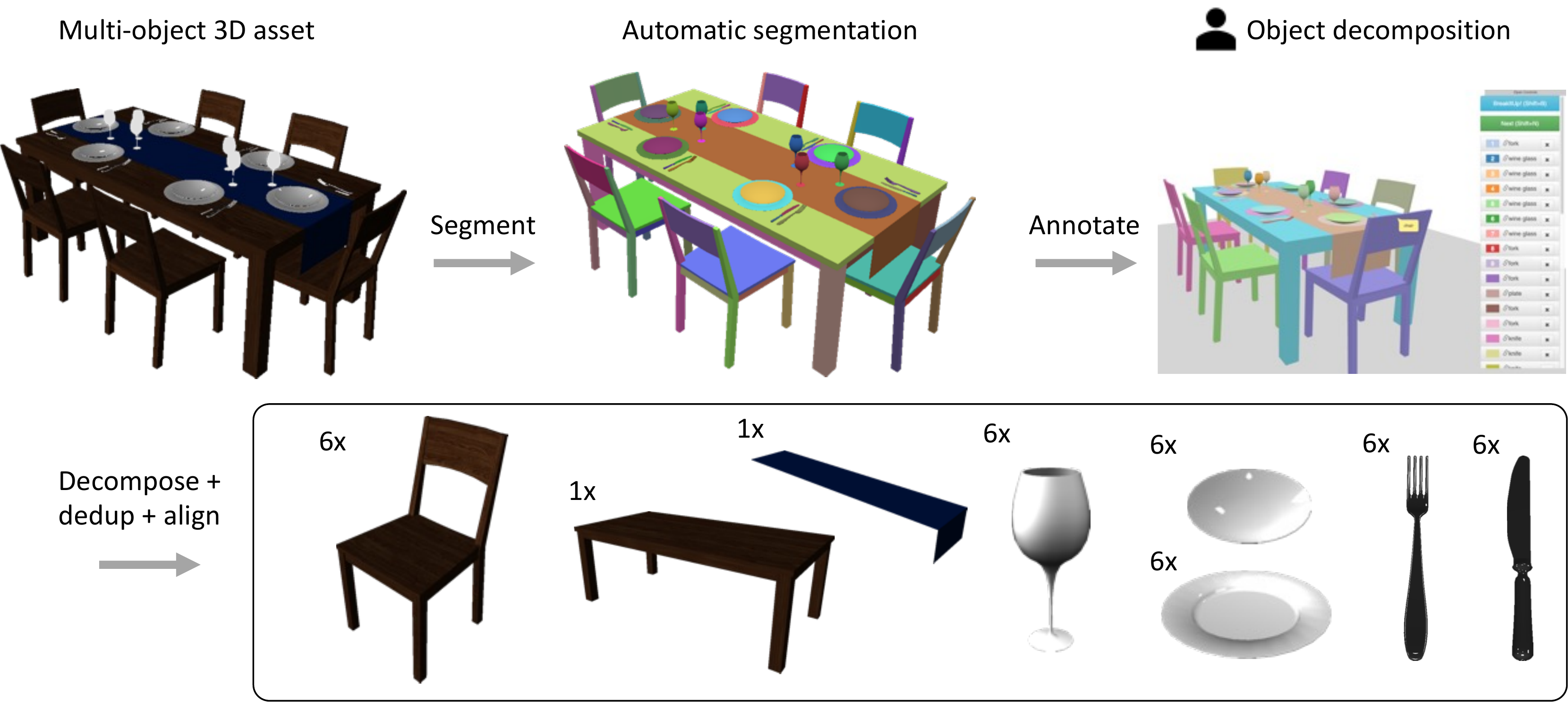}
\caption{
\textbf{Illustration of decomposition process for multi-object assets.}
We create independent object assets for arrangements such as the dining table seen at the left.
The process involves an automatic segmentation, manual annotation to group segments into individual objects, and finally an algorithmic decomposition, deduplication, and alignment of all extracted object instances.
}
\label{fig:fp-decompose-multiobject}
\end{figure*}

In this supplement, we provide more details about our datasets (\Cref{sec:supp-data}), episode generation procedure (\Cref{sec:supp-episode-gen}), analysis \& experiment details (\Cref{sec:supp-expr-details}), \update{training plots and finetuning results (\Cref{sec:train-and-ft-results})}, and agent failure case analysis (\Cref{sec:supp-expr-error}).

\section{Dataset details}
\label{sec:supp-data}

\subsection{\ourdataset dataset construction details}

We show the process for constructing \ourdataset in \Cref{fig:fp-data-pipeline}.
Starting with the glTF format assets representing the 211 scenes from Floorplanner, we use node information of the underlying asset IDs to decompose and extract over 18K unique 3D assets representing furniture and other objects.
The architectural layout for each of the 211 scenes is what remains after this extraction from each scene.
The object assets and the architecture are then compressed as described in the main paper. 

The 3D assets are used to create a 3D model database which we clean and annotate with semantic information.
We use a UI that allows us to select multiple 3D models and tag them with semantic attributes such as WordNet synset, room that the object is typically found in, and on what side the object typically attaches to other objects (bottom vs vertical vs top).
For the semantic categorization step, we start with tags that are provided by Floorplanner and refine and correct the categories.
The interface allows annotators to pull up a 3D view of each object and closely examine it.
The 3D view also provides an interface for semantically annotating the up and front orientation of each object so we have semantically aligned objects.
We find most objects already have consistent alignment, and only re-align objects that are not consistently aligned (typically wall objects).

In addition, we also mark whether the 3D asset represents multiple objects.
For 3D assets that are marked as being composed of multiple objects, we follow the process depicted in \Cref{fig:fp-decompose-multiobject} to decompose the 3D asset into multiple objects.
We first obtain an automatic segmentation using connected component analysis, and then have users manually ``paint'' and ``label'' the objects.
Due to the clean geometry, we can obtain clean segmentations.
Our interface allows for easy marking of object parts, displaying of the bounding box of annotated objects, copying of labels, undo/redo operations etc.
The annotated objects are algorithmically extracted, deduplicated and aligned.  

At the scene level, we also identify floater objects and exterior doors.
Floater objects are objects that are outside of the scenes, and should not be part of the scene.
We remove such objects.
For ObjectNav experiments, we also remove interior doors but keep exterior doors (to prevent the agents from wandering outside).
In addition, we also remove animate objects (animals and humans) from our scenes for all ObjectNav experiments.

\begin{figure}
\includegraphics[width=\linewidth]{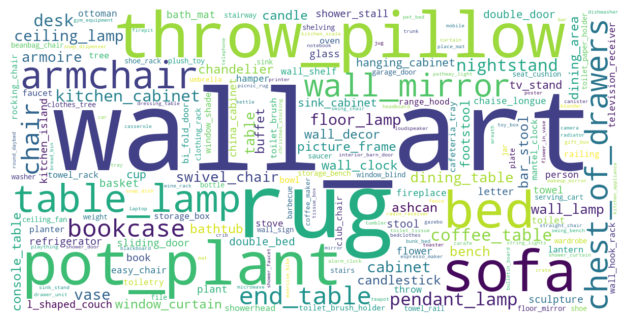}
\caption{
\textbf{Word cloud of object categories in \ourdataset.}
Font sizes indicate unique instance count per category.
The \ourdataset contains a rich variety of categories, with many instances especially for categories that exhibit diversity in the real world (wall art, rugs etc.)
}
\label{fig:word_cloud}
\end{figure}
\begin{figure}
\includegraphics[width=\linewidth]{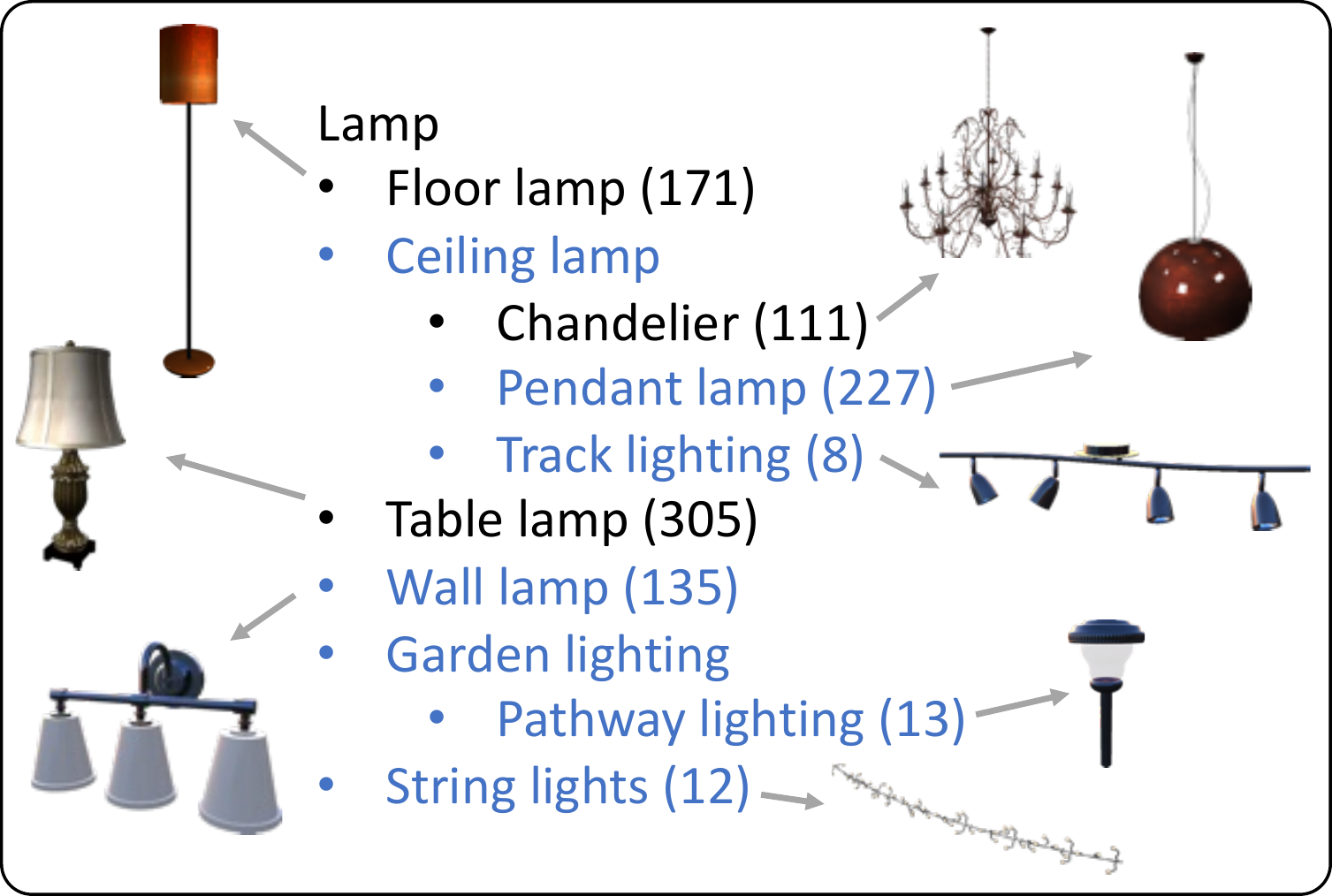}
\caption{
\textbf{Lamp object category hierarchy in \ourdataset.}
Our category hierarchy leverages an augmented taxonomy of WordNet synsets that we call WordNetCO (for WordNet Common Objects).
The figure shows synsets that we introduced in blue.
The count of object instances within each synset is indicated by the number, and an example instance is shown for select synsets.
}
\label{fig:fp-lamp-hierarchy}
\end{figure}

\begin{figure}
\includegraphics[width=\linewidth]{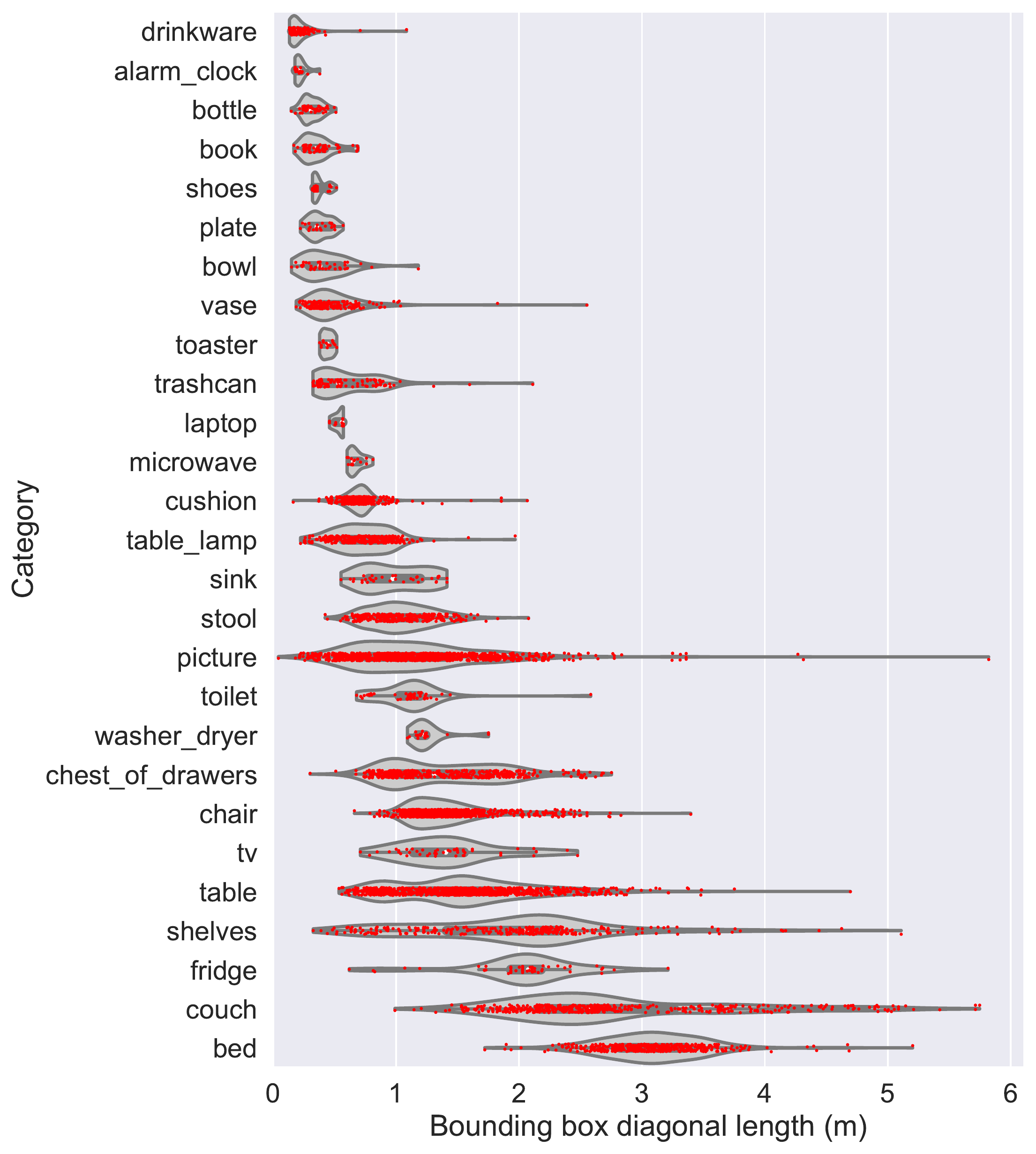}
\caption{
\textbf{Distributions of physical sizes for the 28 category set from \ourdataset.}
Distributions are on the diagonal length of the bounding box for each instance within the category.
Categories are sorted by average diagonal length from top (smallest) to bottom (largest).
The \ourdataset objects are scaled to have realistic physical sizes.
Shelves, tables, pictures, couches have a wide range of physical sizes.
In contrast, categories such as alarm clocks, bottles, plates, and shows have much narrower physical size distributions.
}
\label{fig:fp-object-sizes}
\end{figure}

\subsection{\ourdataset dataset statistics}
\Cref{fig:word_cloud} shows a word cloud visualization of categories in \ourdataset, with the text font size representing the total count of unique object instances in each category.
We see that our dataset contains a diverse set of object categories.
As described in the main paper, we annotate the objects in \ourdataset using a taxonomy based on WordNet, but extended to include additional common object categories.
\Cref{fig:fp-lamp-hierarchy} shows a sub-tree of this WordNetCO category hierarchy, focusing on lamp objects.
The breadth and fine-grained nature of the taxonomy allows for future experiments with embodied AI agents tackling scenarios requiring perception of objects closer to an open vocabulary setting.

The object co-occurrence analysis in the main paper is on the basis of a set of 28 common object types that are shared between ProcTHOR \cite{deitke2022procthor}, \ourdataset, and HM3DSem \cite{yadav2022hm3dsem}.
The complete list of these categories is: alarm\_clock, bed, book, bottle, bowl, chair, chest\_of\_drawers, couch, cushion, drinkware, fridge, laptop, microwave, picture, plate, potted\_plant, shelves, shoes, sink, stool, table, table\_lamp, toaster, toilet, trashcan, tv, vase, washer\_dryer.

We plot the size distribution (measured by the diagonal length of the bounding box in meters) of the 28 object categories in \Cref{fig:fp-object-sizes}.
We see that the \ourdataset objects exhibit realistic sizes, with some categories having fairly narrow size distributions (e.g., shoes) and some having fairly broad distributions (e.g., pictures, shelves and beds).

\begin{figure*}
\setkeys{Gin}{width=\linewidth}
\begin{tabularx}{\linewidth}{@{}rYYYYYYYY@{}}
\toprule

bed &
\includegraphics[]{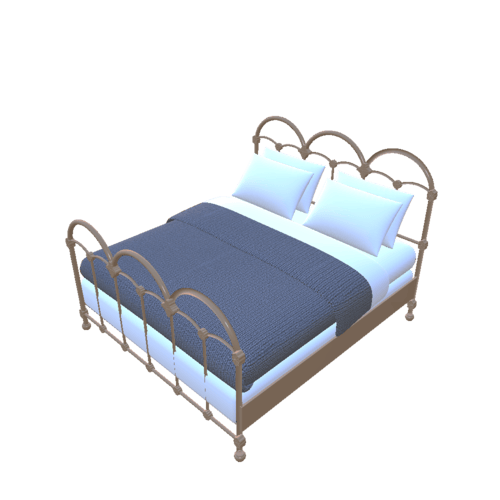} &
\includegraphics[]{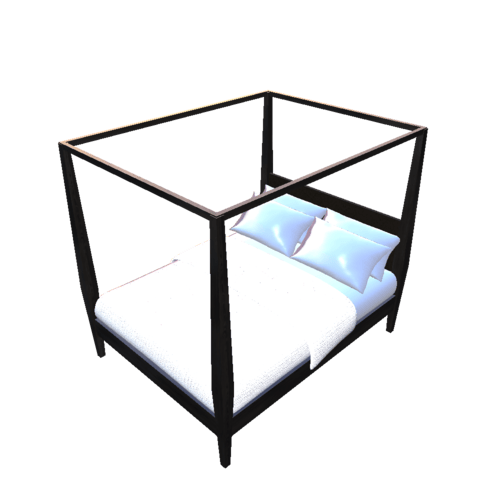} &
\includegraphics[]{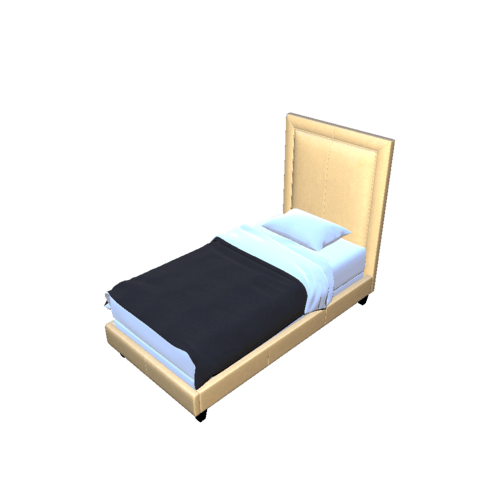} &
\includegraphics[]{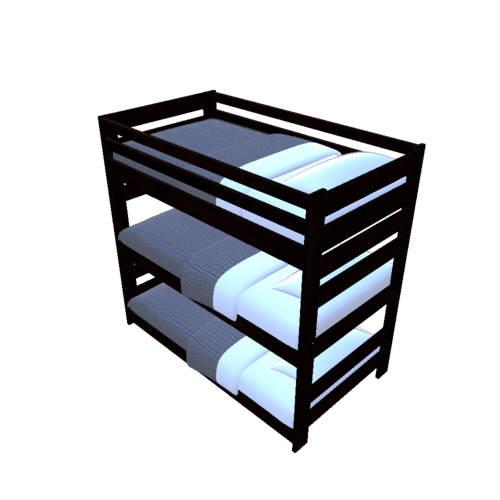} &
\includegraphics[]{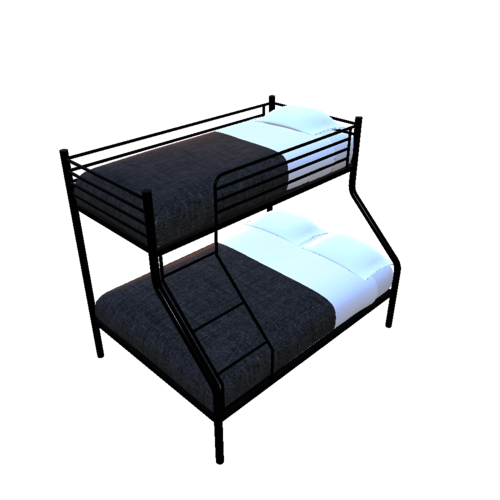} &
\includegraphics[]{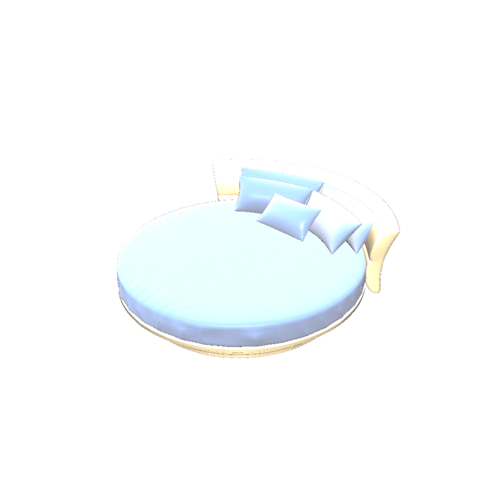} &
\includegraphics[]{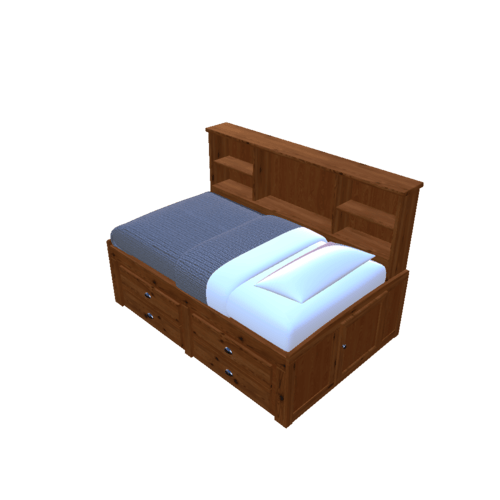} &
\includegraphics[]{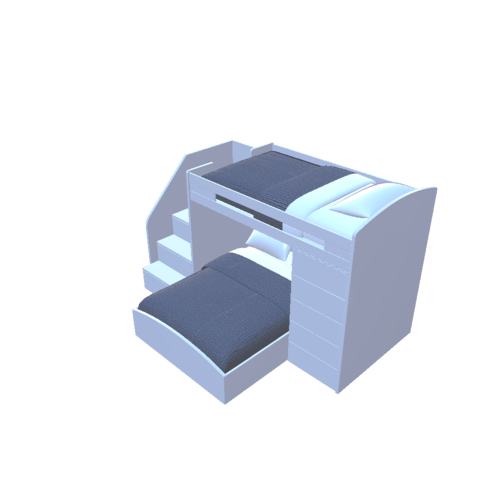} \\
shelving &
\includegraphics[]{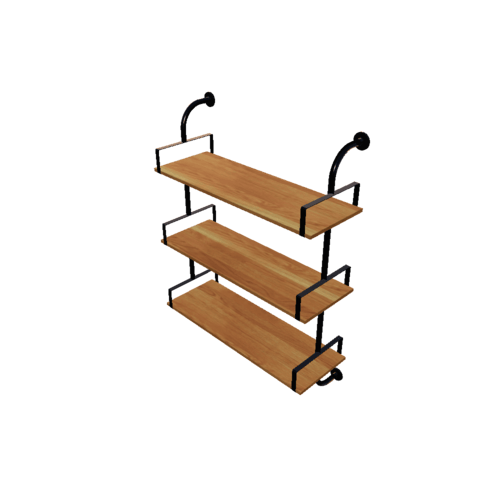} &
\includegraphics[]{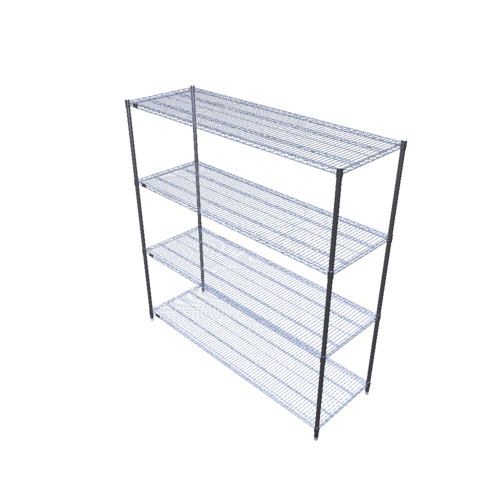} &
\includegraphics[]{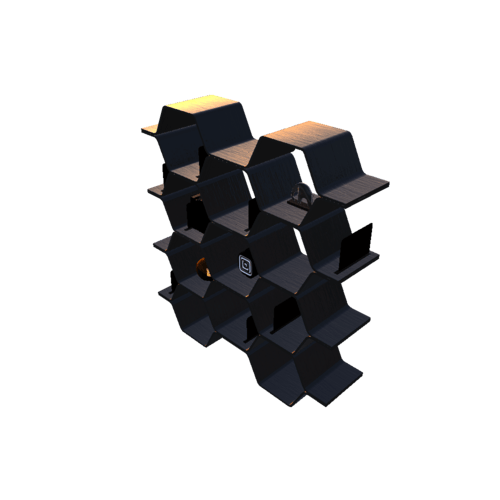} &
\includegraphics[]{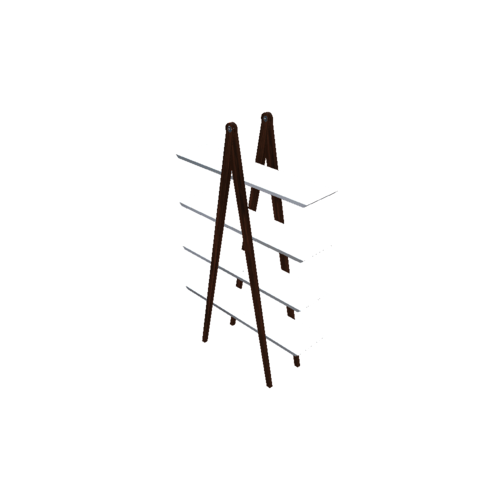} &
\includegraphics[]{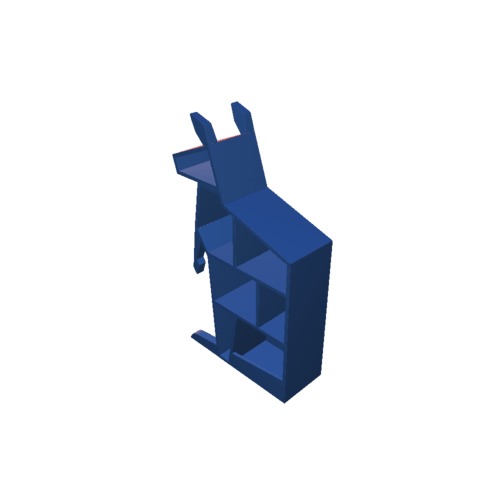} &
\includegraphics[]{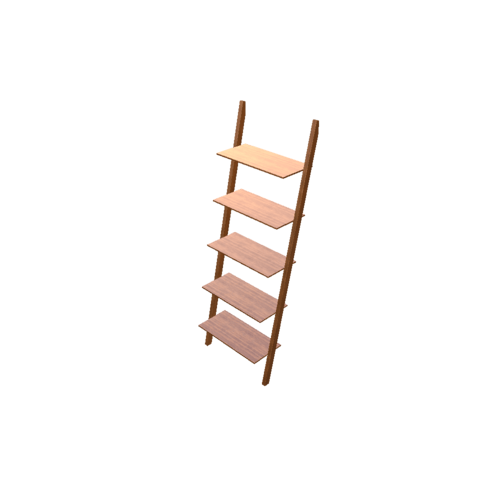} &
\includegraphics[]{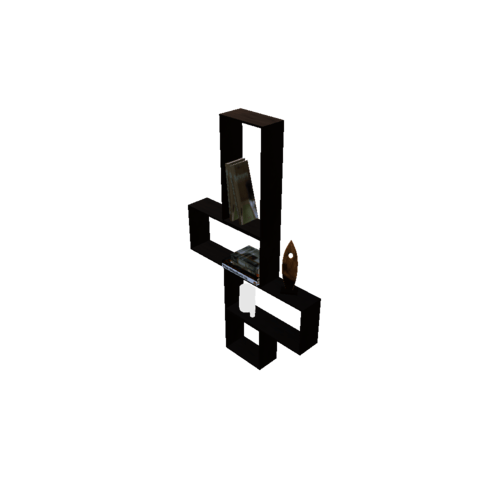} &
\includegraphics[]{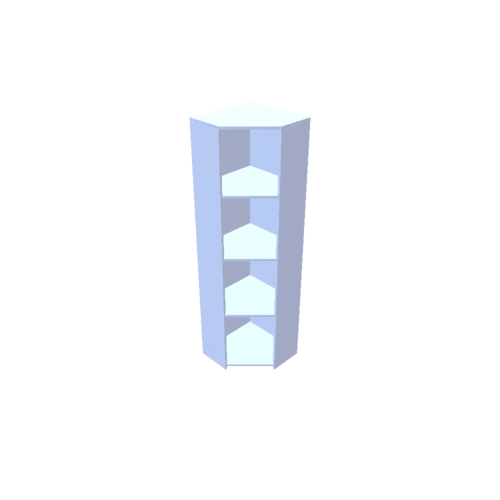} \\
chair &
\includegraphics[]{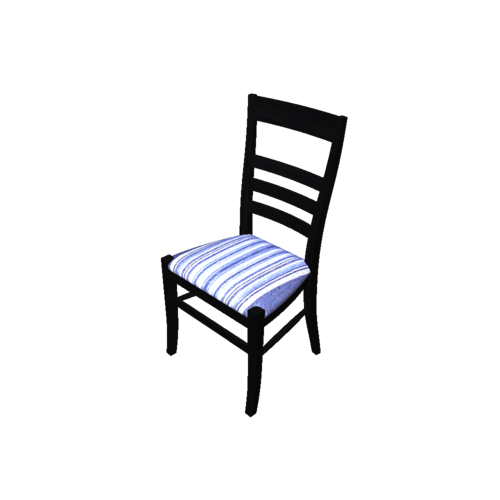} &
\includegraphics[]{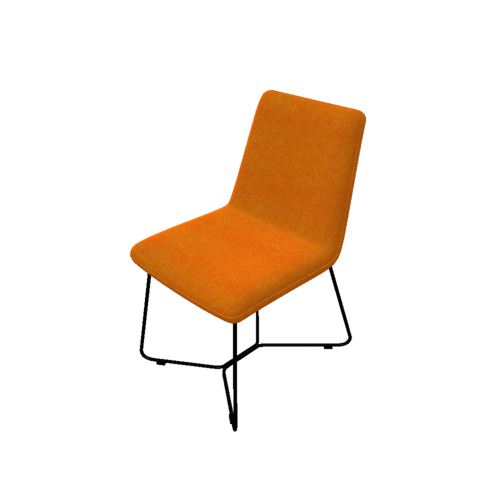} &
\includegraphics[]{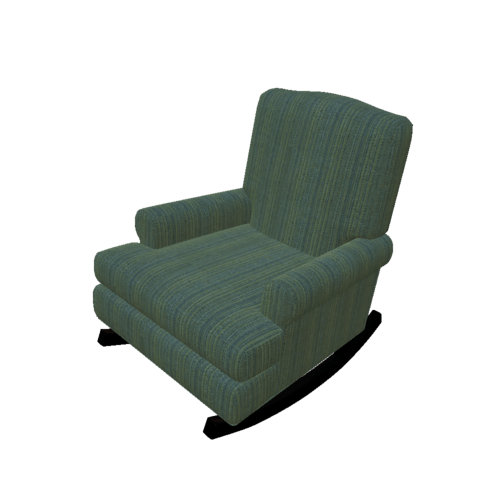} &
\includegraphics[]{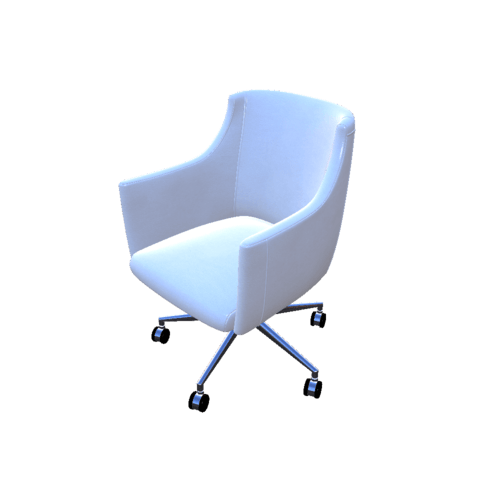} &
\includegraphics[]{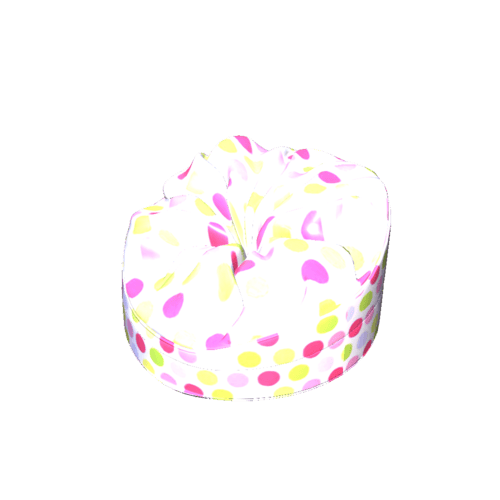} &
\includegraphics[]{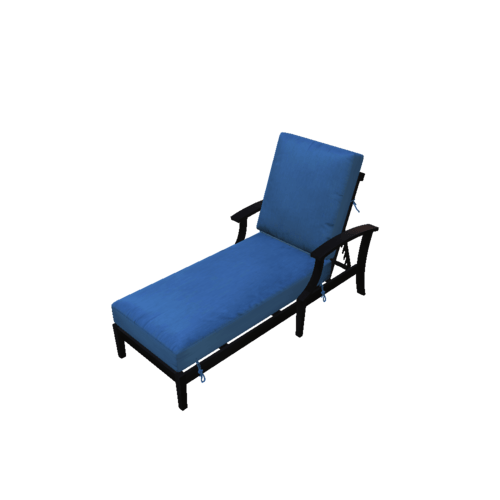} &
\includegraphics[]{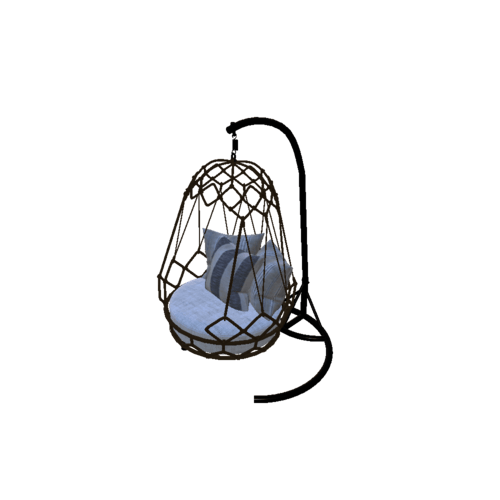} &
\includegraphics[]{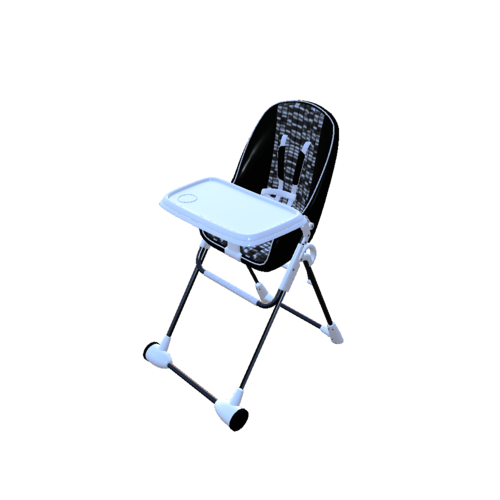} \\
table &
\includegraphics[]{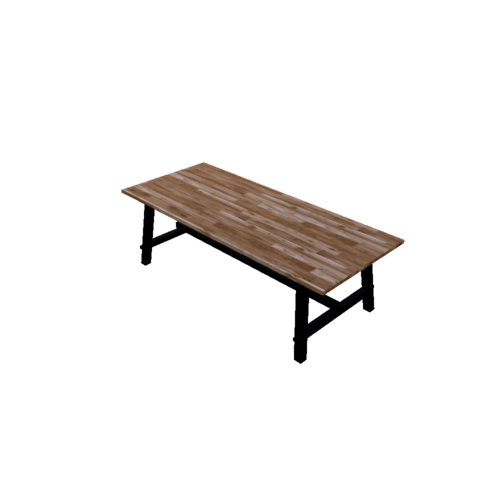} &
\includegraphics[]{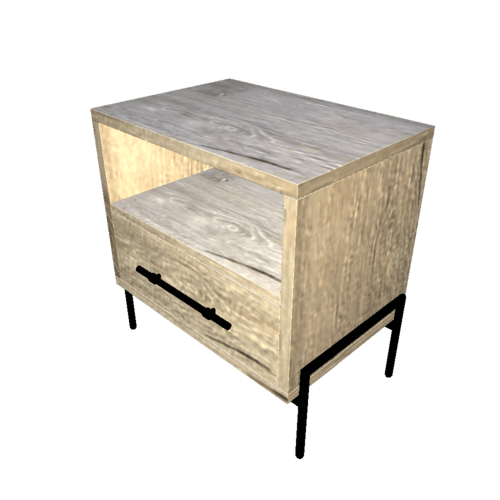} &
\includegraphics[]{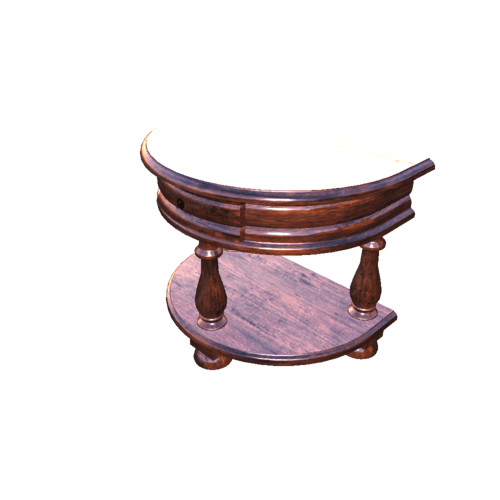} &
\includegraphics[]{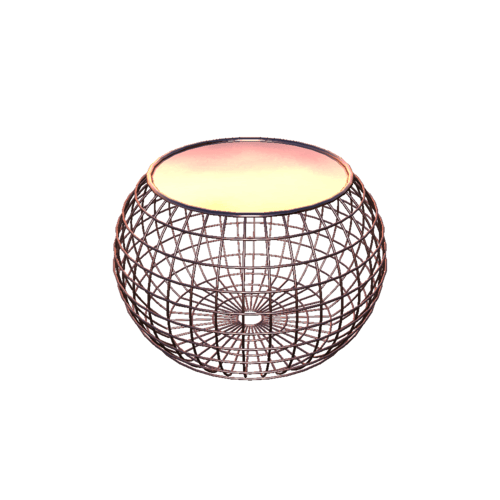} &
\includegraphics[]{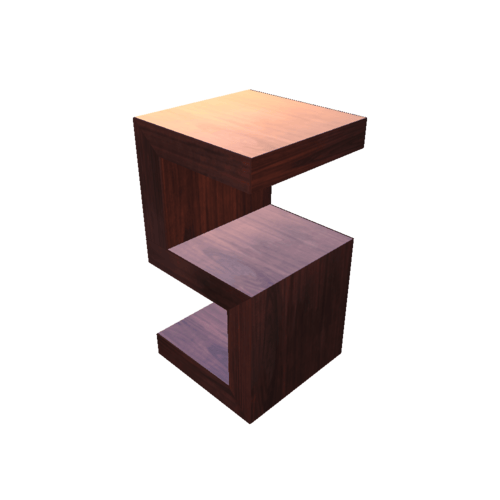} &
\includegraphics[]{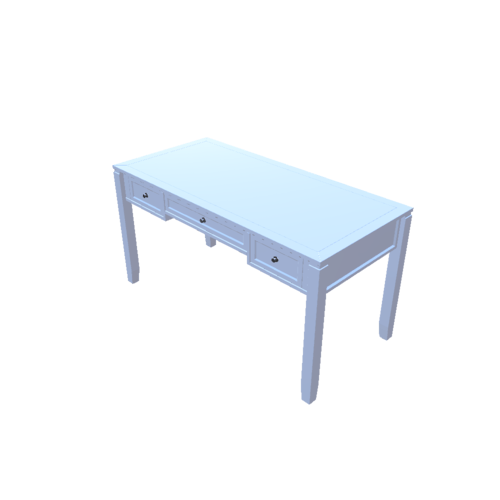} &
\includegraphics[]{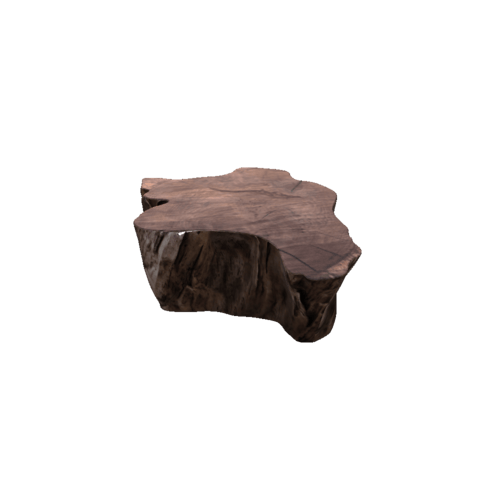} &
\includegraphics[]{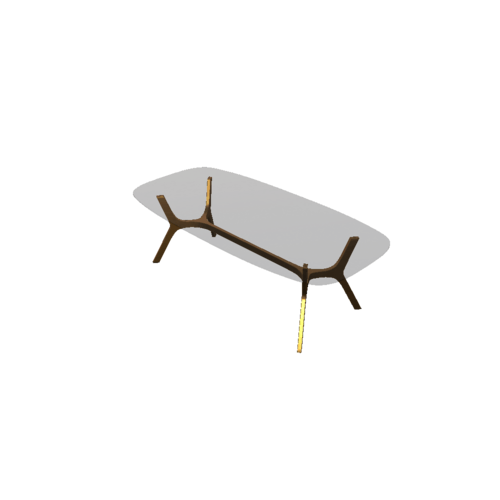} \\
table lamp &
\includegraphics[]{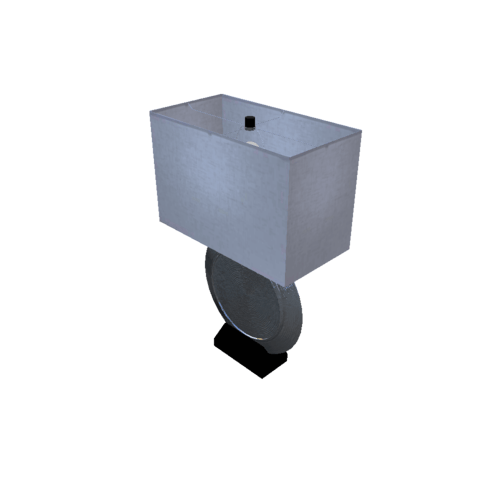} &
\includegraphics[]{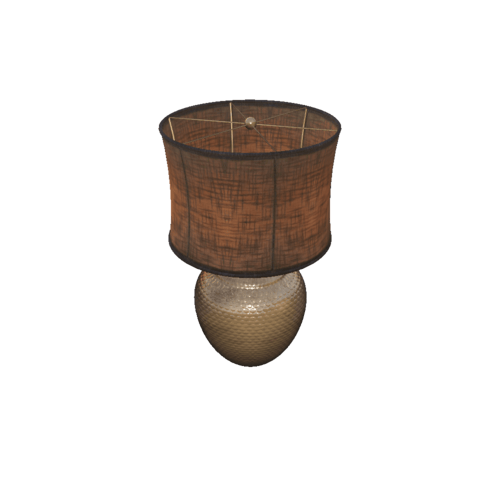} &
\includegraphics[]{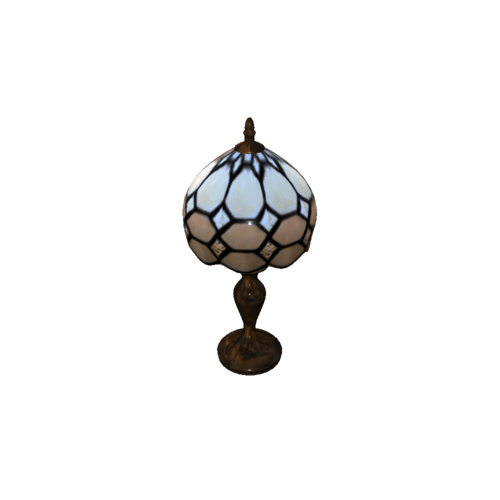} &
\includegraphics[]{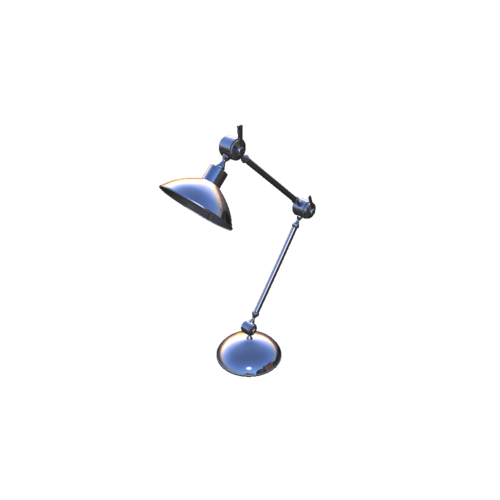} &
\includegraphics[]{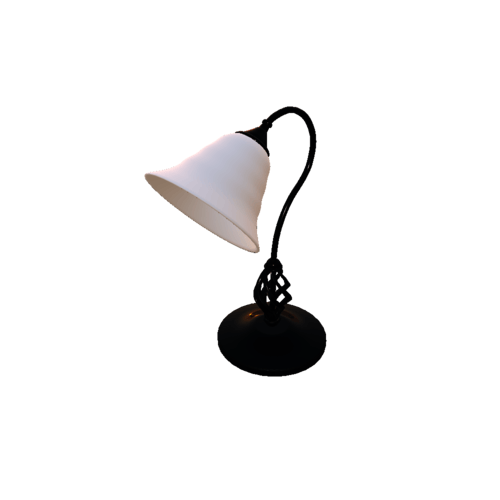} &
\includegraphics[]{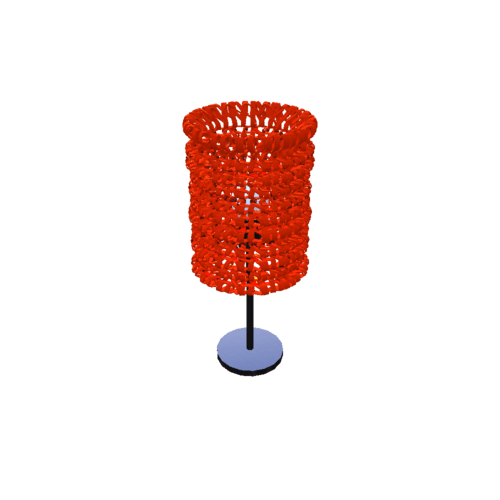} &
\includegraphics[]{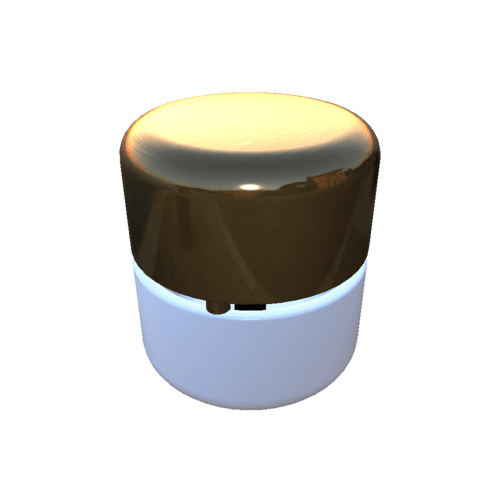} &
\includegraphics[]{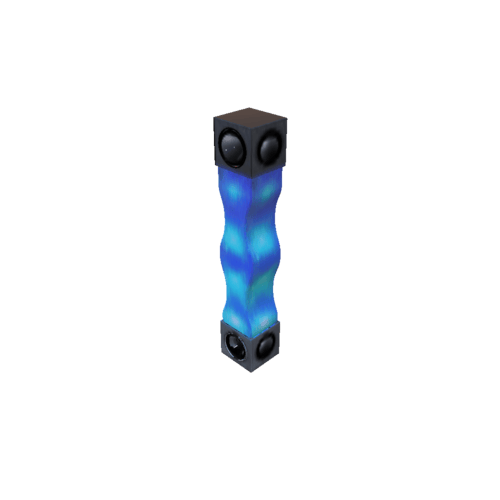} \\
floor lamp &
\includegraphics[]{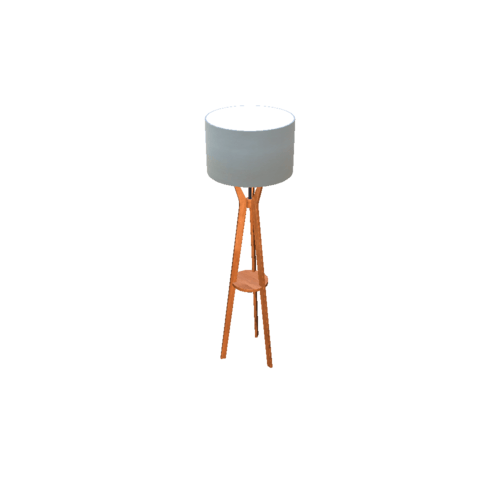} &
\includegraphics[]{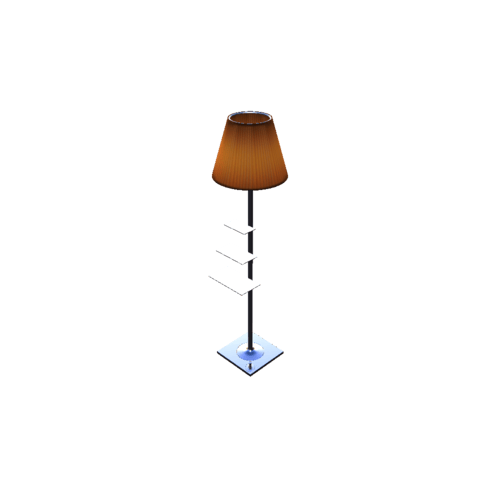} &
\includegraphics[]{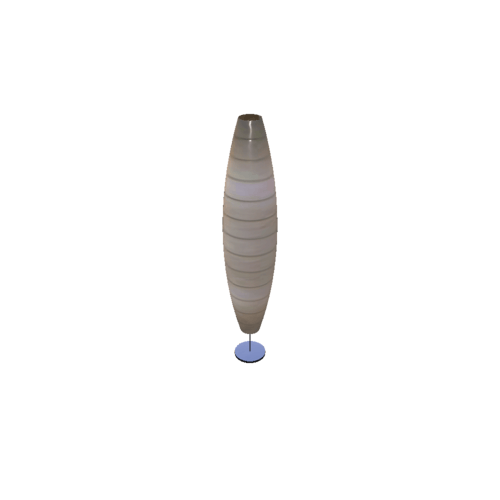} &
\includegraphics[]{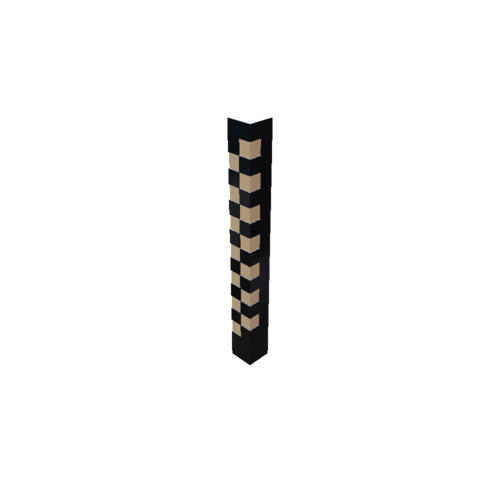} &
\includegraphics[]{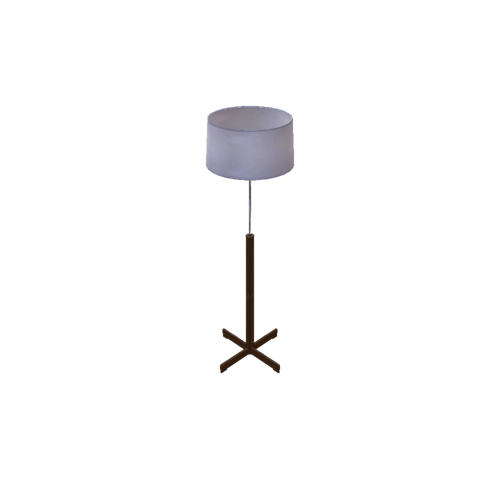} &
\includegraphics[]{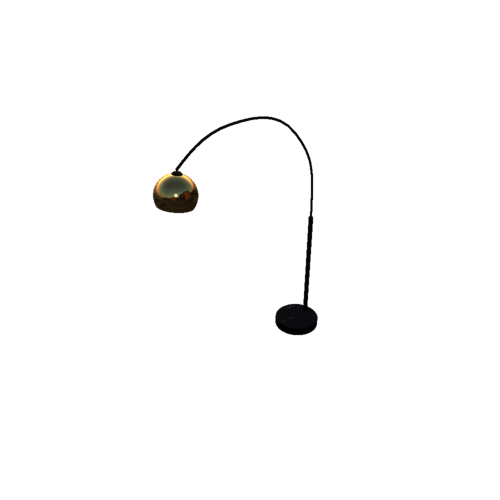} &
\includegraphics[]{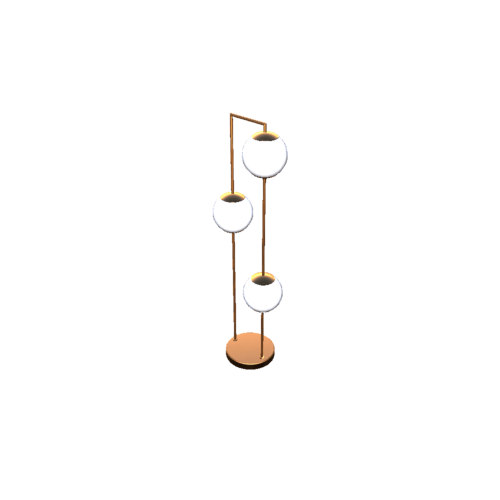} &
\includegraphics[]{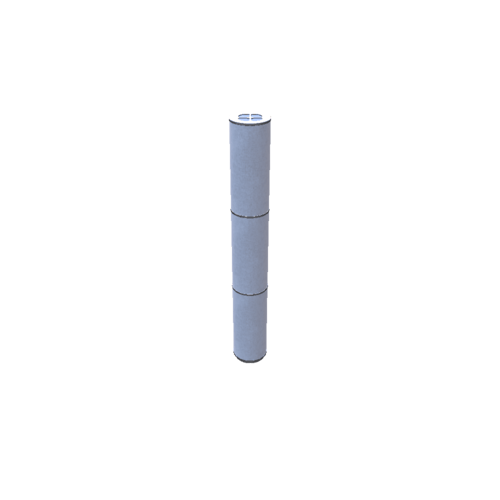} \\
wall lamp &
\includegraphics[]{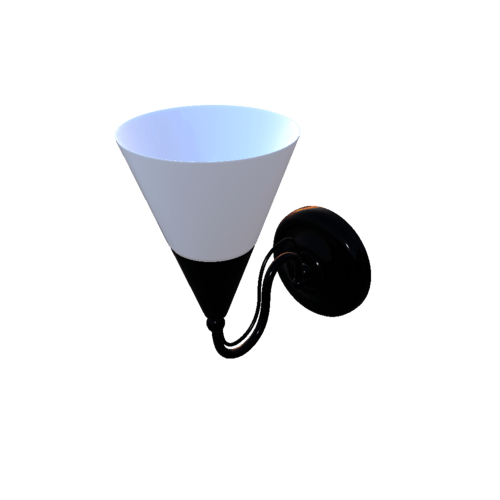} &
\includegraphics[]{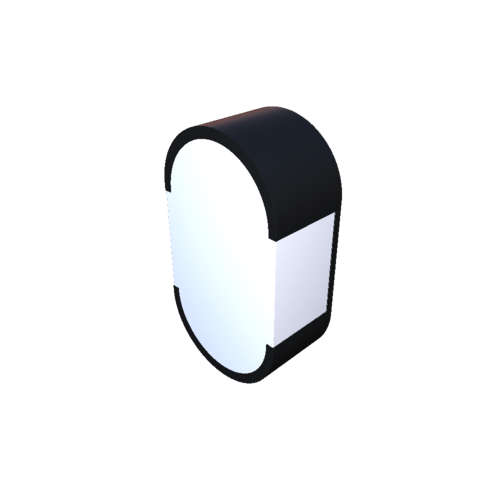} &
\includegraphics[]{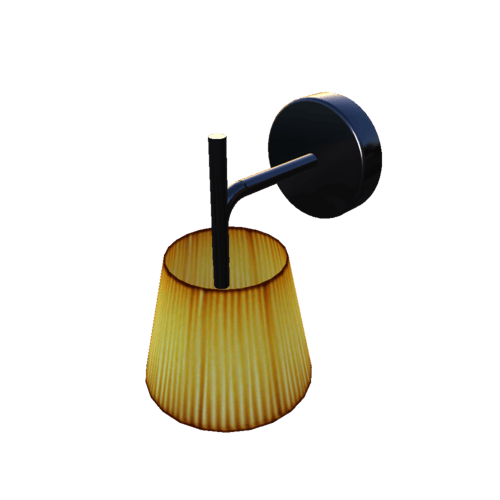} &
\includegraphics[]{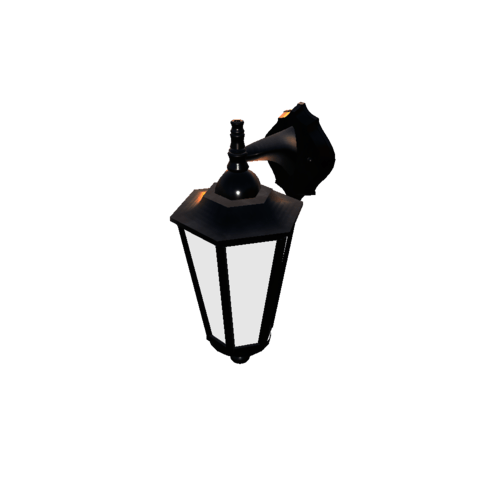} &
\includegraphics[]{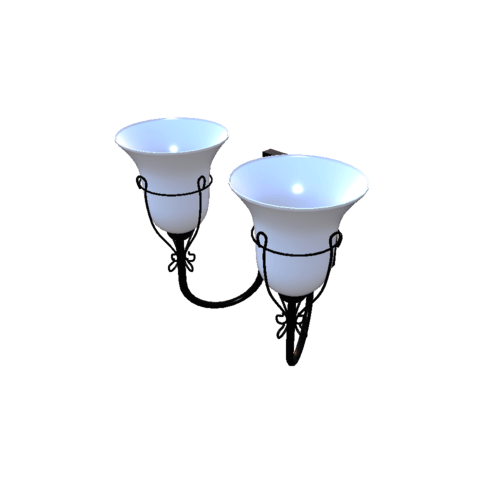} &
\includegraphics[]{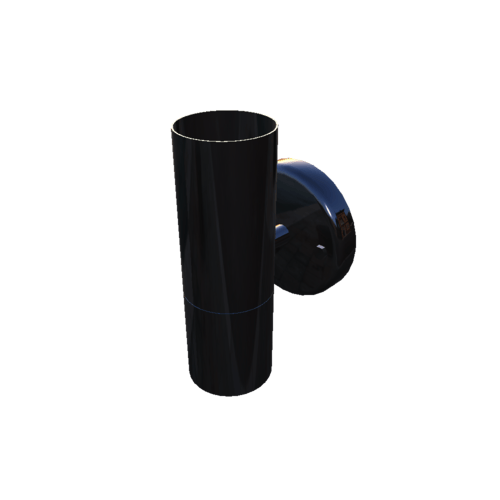} &
\includegraphics[]{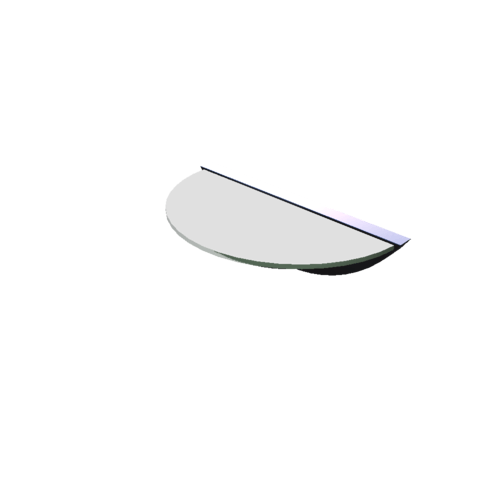} &
\includegraphics[]{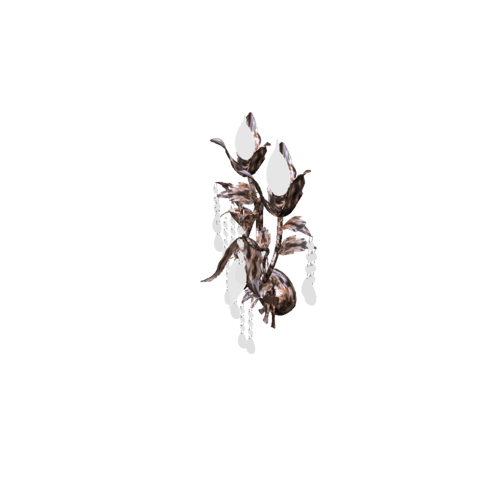} \\
fan &
\includegraphics[]{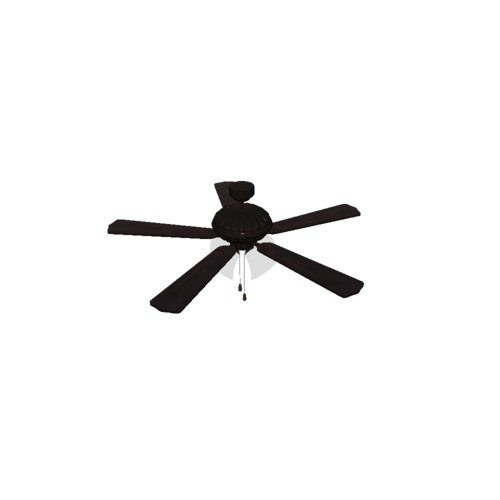} &
\includegraphics[]{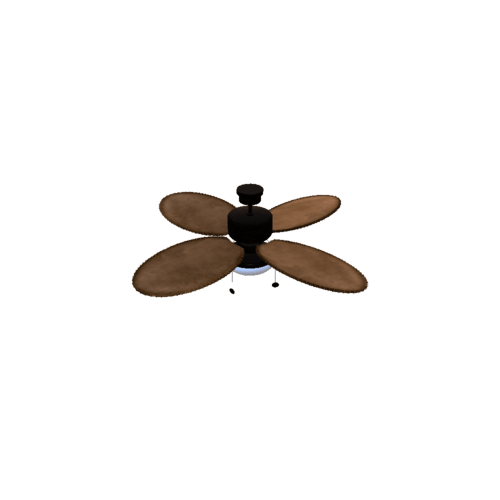} &
\includegraphics[]{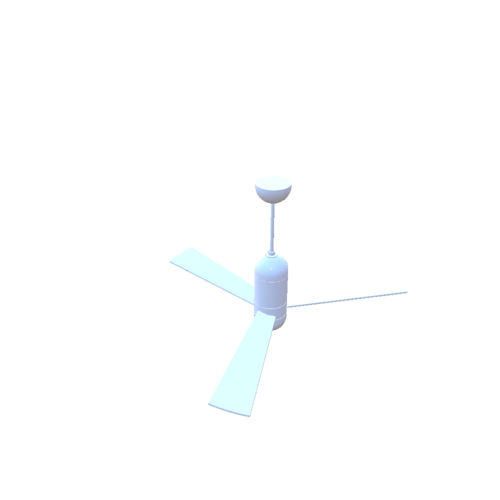} &
\includegraphics[]{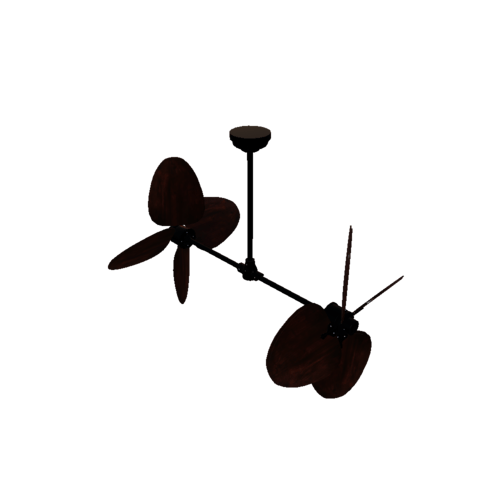} &
\includegraphics[]{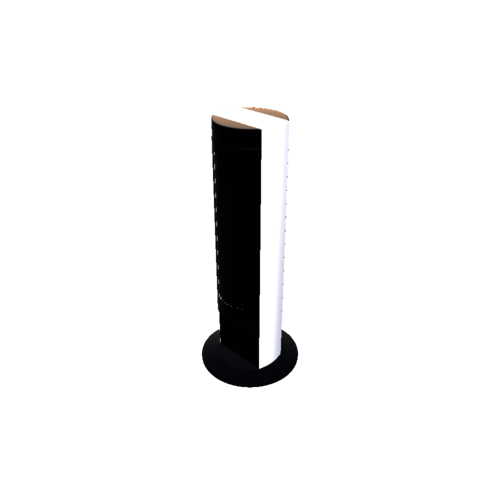} &
\includegraphics[]{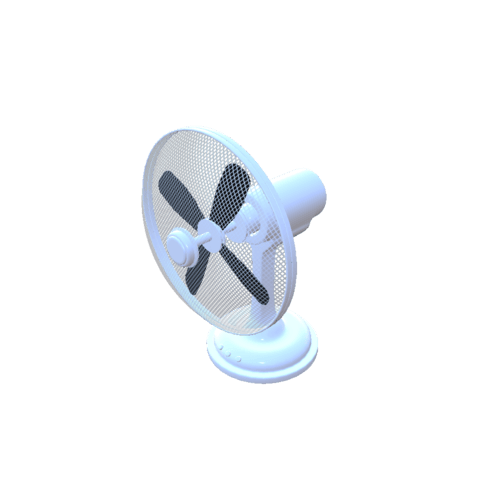} &
\includegraphics[]{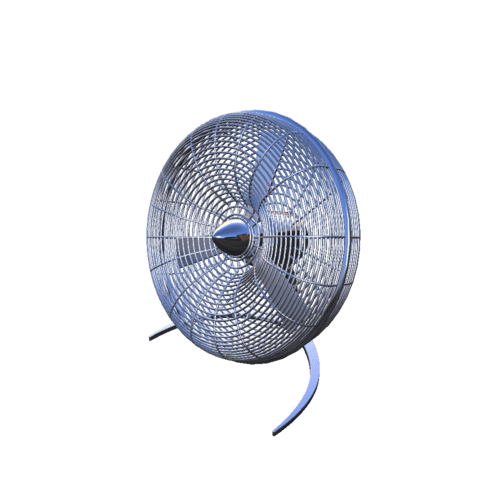} &
\includegraphics[]{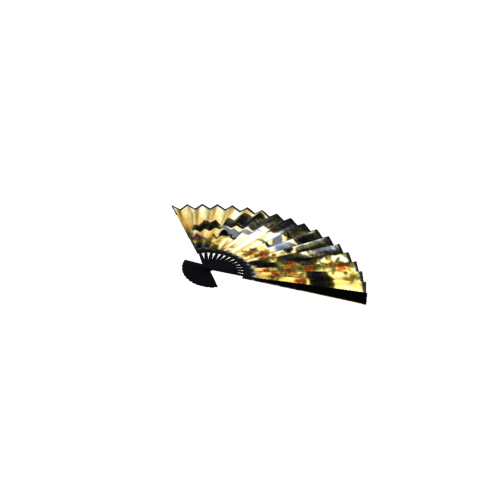} \\
toiletry &
\includegraphics[]{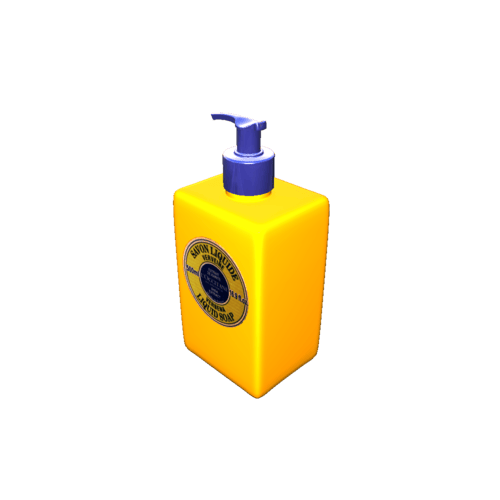} &
\includegraphics[]{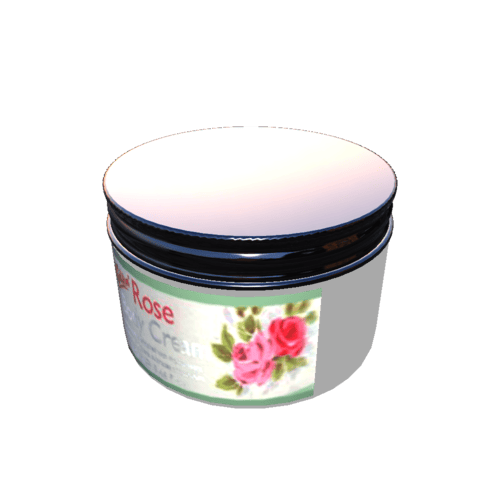} &
\includegraphics[]{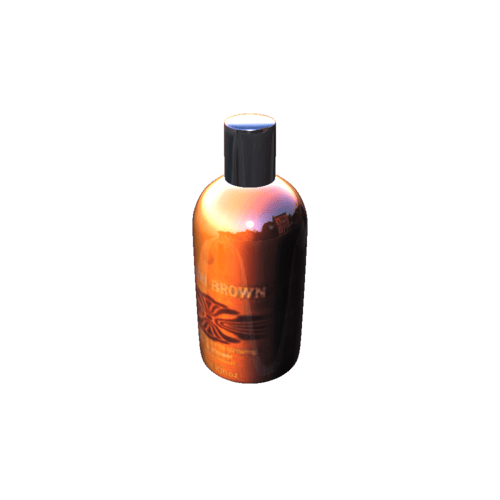} &
\includegraphics[]{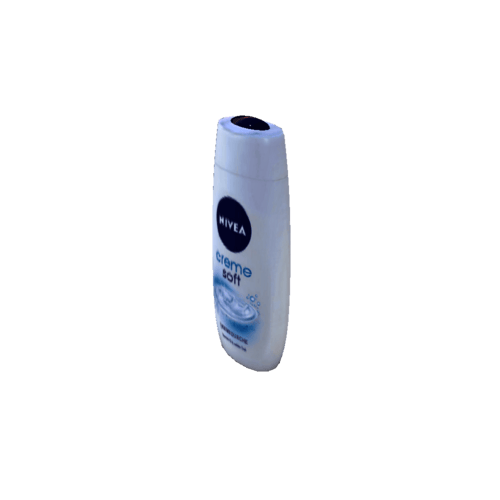} &
\includegraphics[]{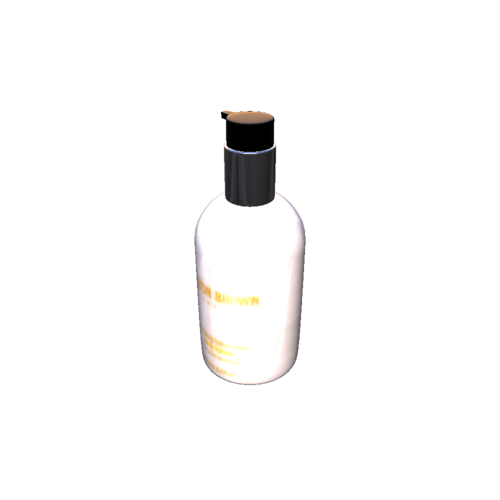} &
\includegraphics[]{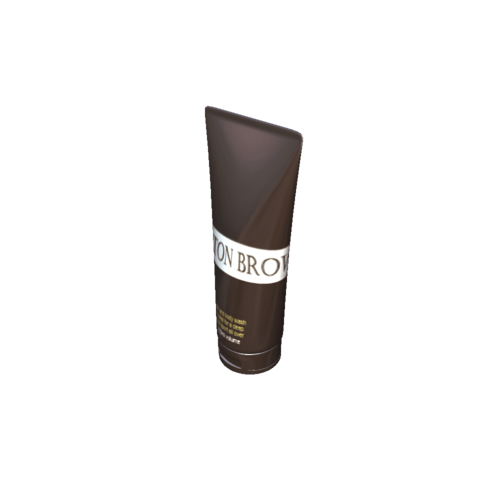} &
\includegraphics[]{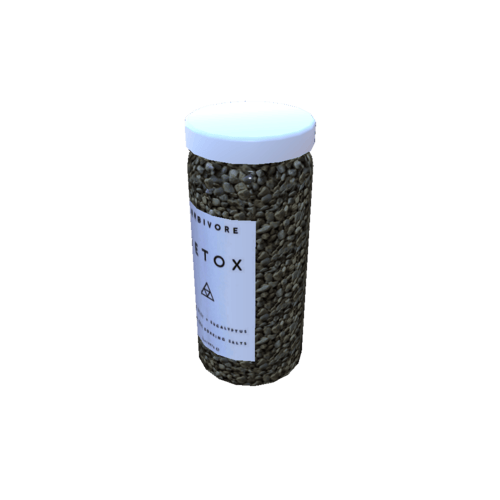} &
\includegraphics[]{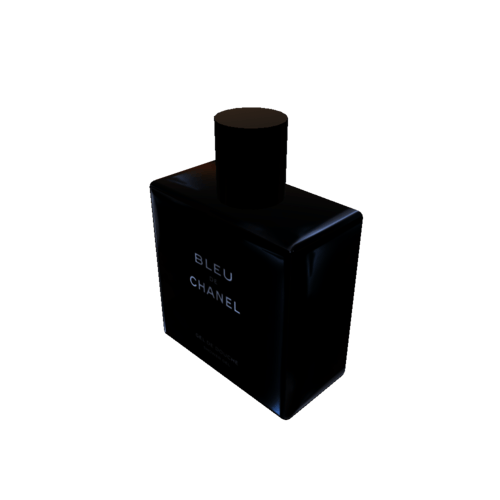} \\
radio &
\includegraphics[]{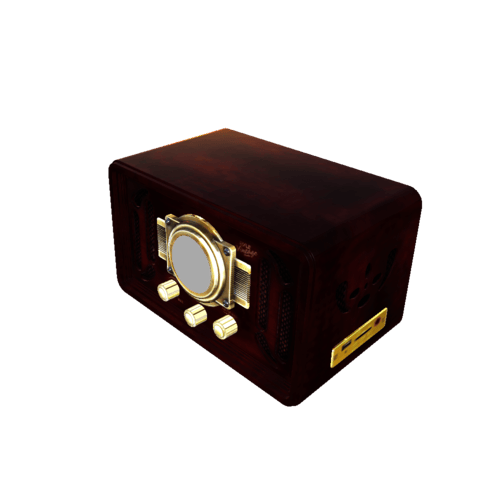} &
\includegraphics[]{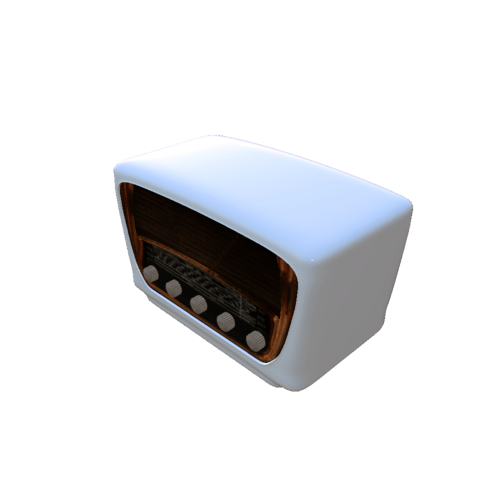} &
\includegraphics[]{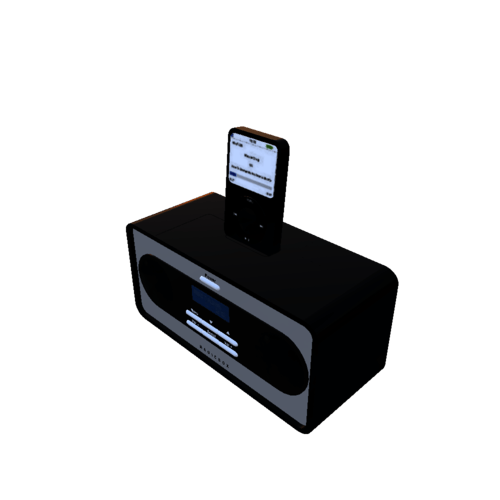} &
\includegraphics[]{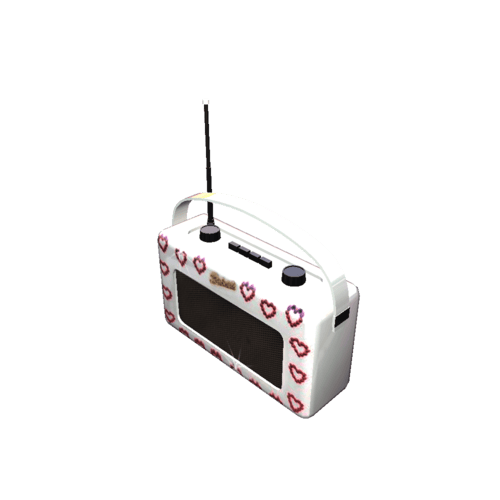} &
\includegraphics[]{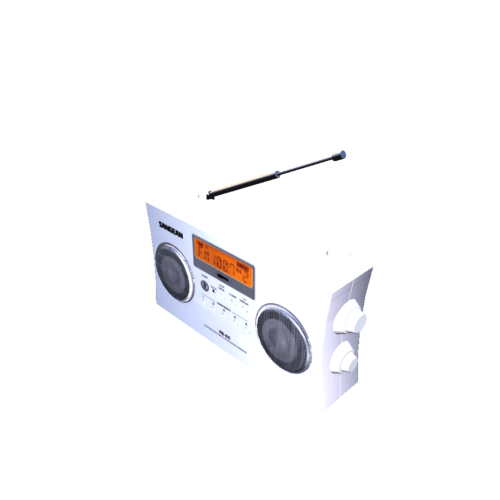} &
\includegraphics[]{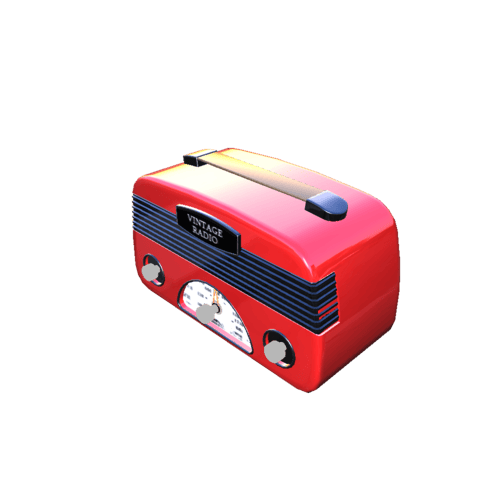} &
\includegraphics[]{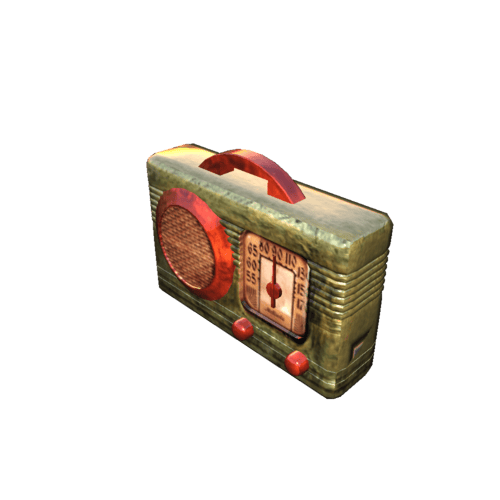} &
\includegraphics[]{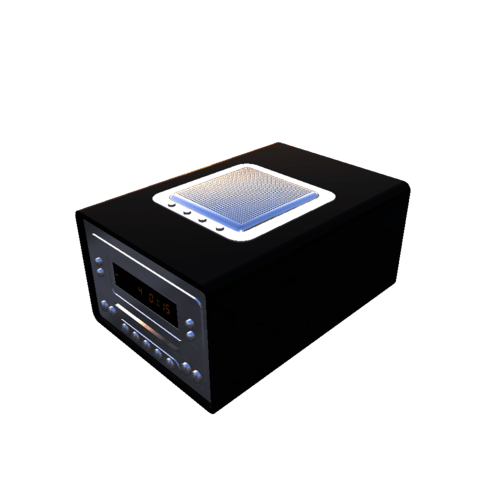} \\
mirror &
\includegraphics[]{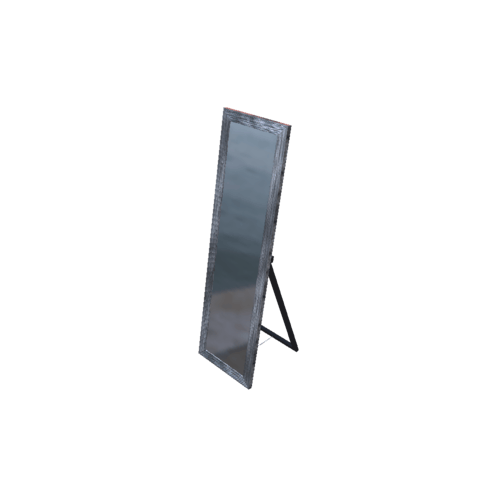} &
\includegraphics[]{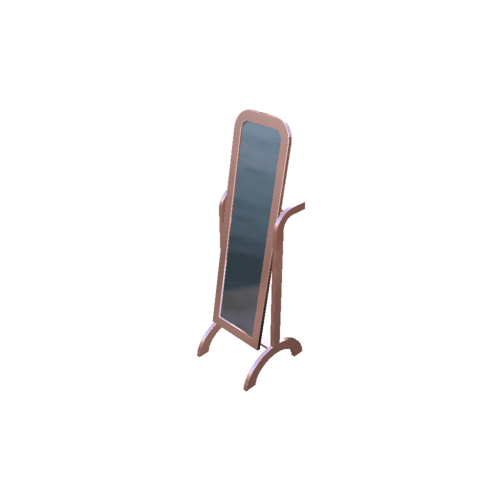} &
\includegraphics[]{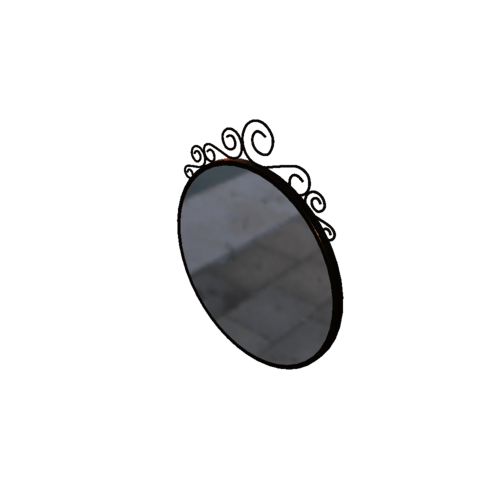} &
\includegraphics[]{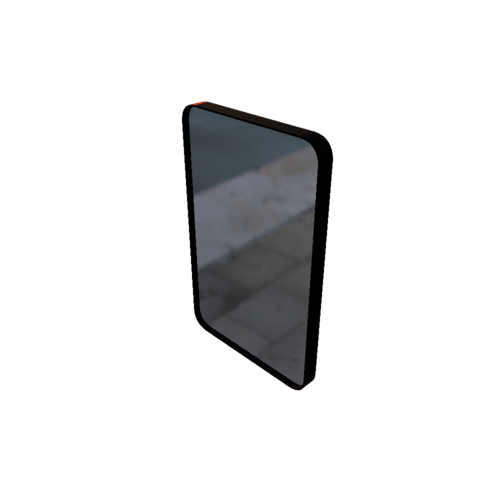} &
\includegraphics[]{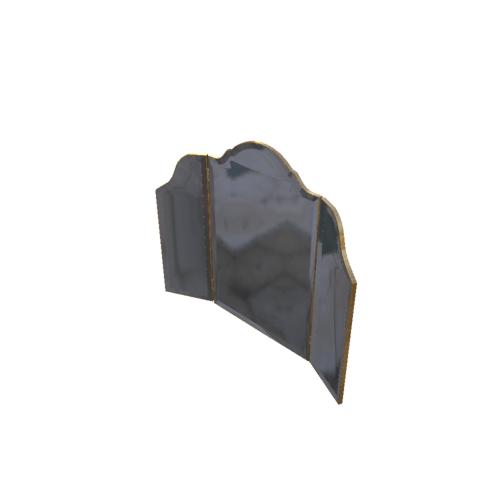} &
\includegraphics[]{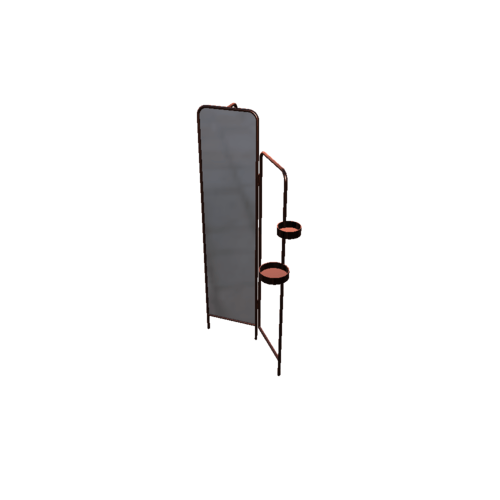} &
\includegraphics[]{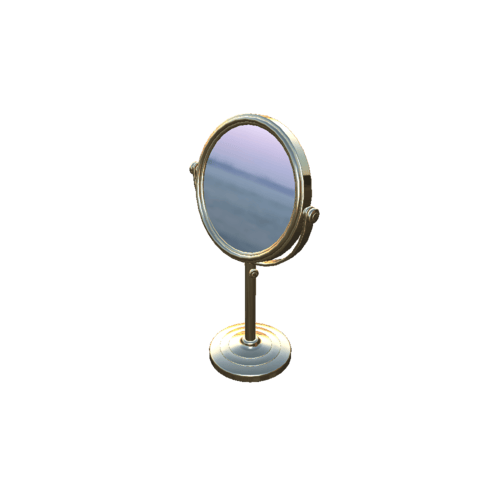} &
\includegraphics[]{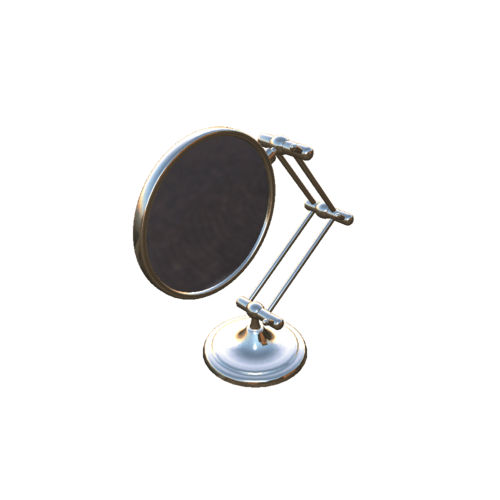} \\
clock &
\includegraphics[]{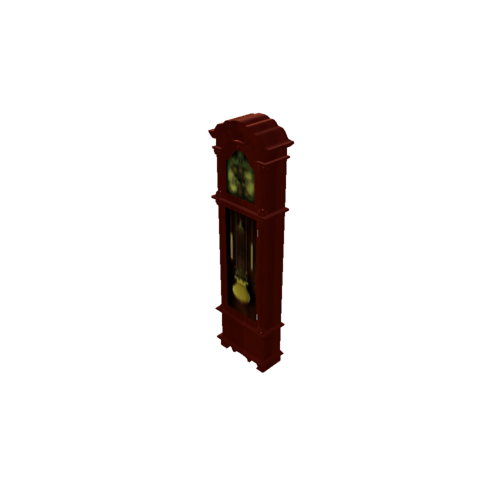} &
\includegraphics[]{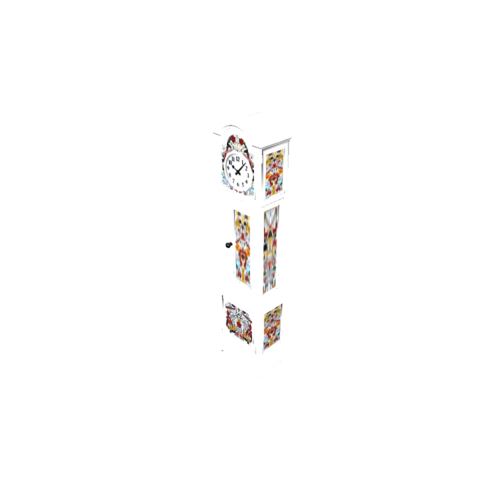} &
\includegraphics[]{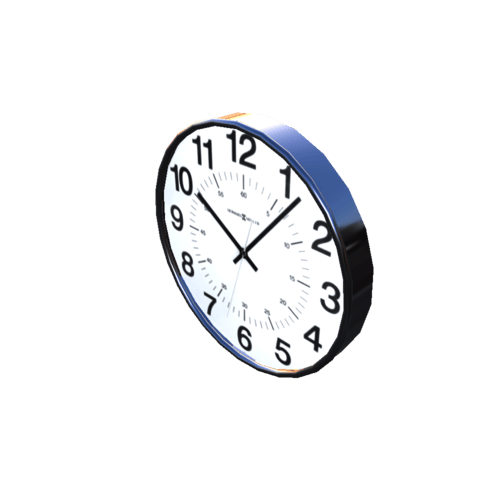} &
\includegraphics[]{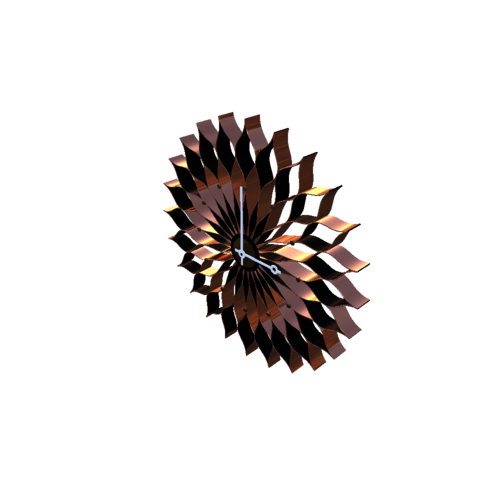} &
\includegraphics[]{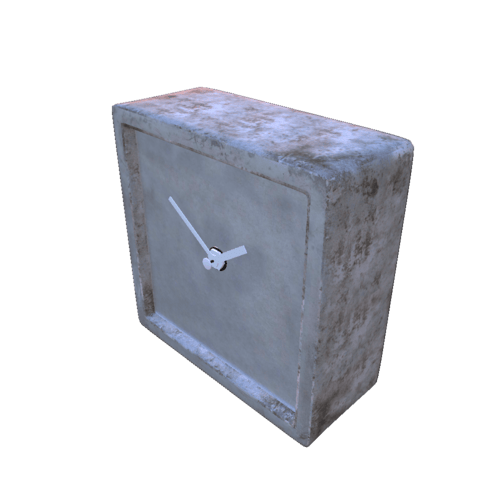} &
\includegraphics[]{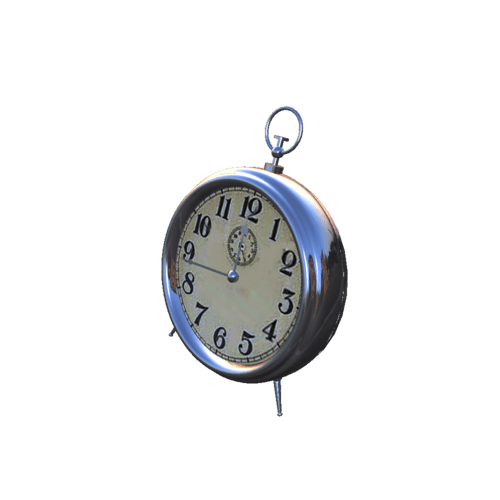} &
\includegraphics[]{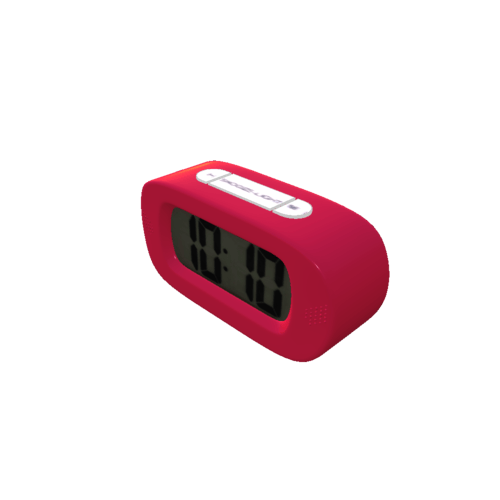} &
\includegraphics[]{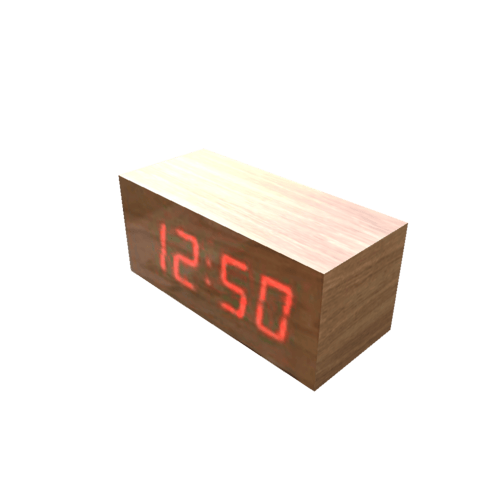} \\

\bottomrule
\end{tabularx}
\vspace{1pt}
\caption{
\textbf{Example objects from several categories.}
The \ourdataset-200 dataset contains a broad variety of object categories, each with a diverse set of object instances within each category.
Note the variety of lamp categories (table lamps, floor lamps, wall lamps), each exhibiting diversity of object instances with different physical sizes, geometry, and fine-grained appearance.
}
\label{fig:vis-fp-objects}
\end{figure*}
\begin{figure*}
\setkeys{Gin}{width=\linewidth}
\begin{tabularx}{\linewidth}{@{}YYY@{}}
\toprule
 ProcTHOR \cite{deitke2022procthor} & \ourdataset & HM3DSem \cite{yadav2022hm3dsem} \\
\includegraphics[]{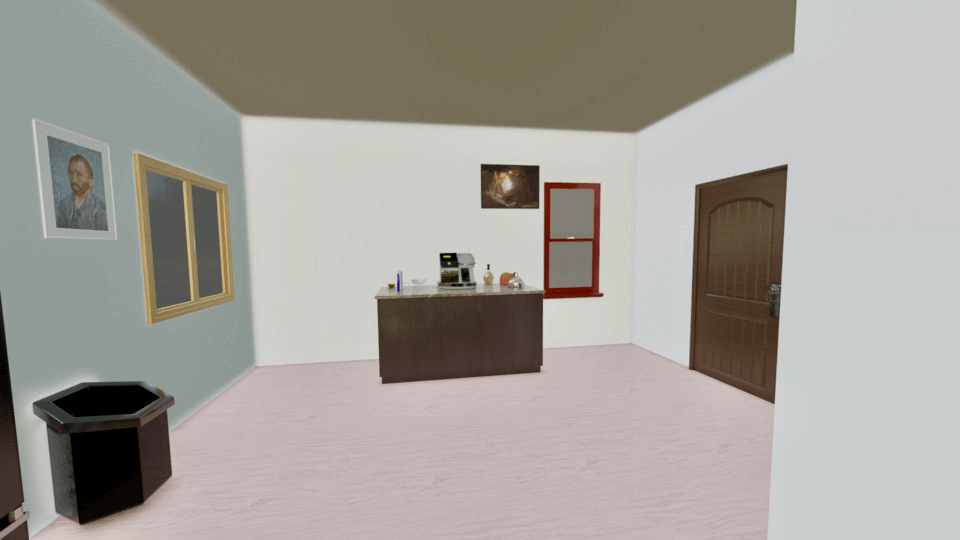} & \includegraphics[]{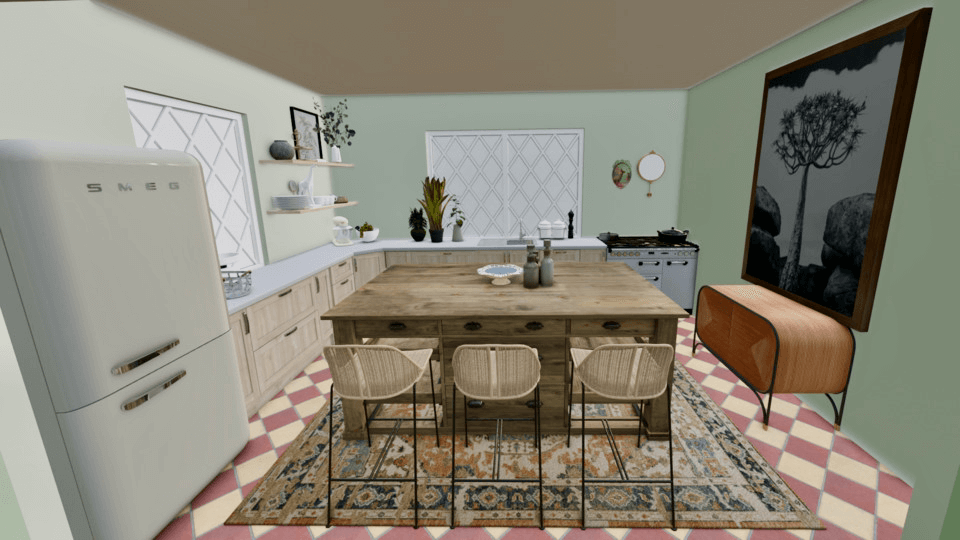} & \includegraphics[]{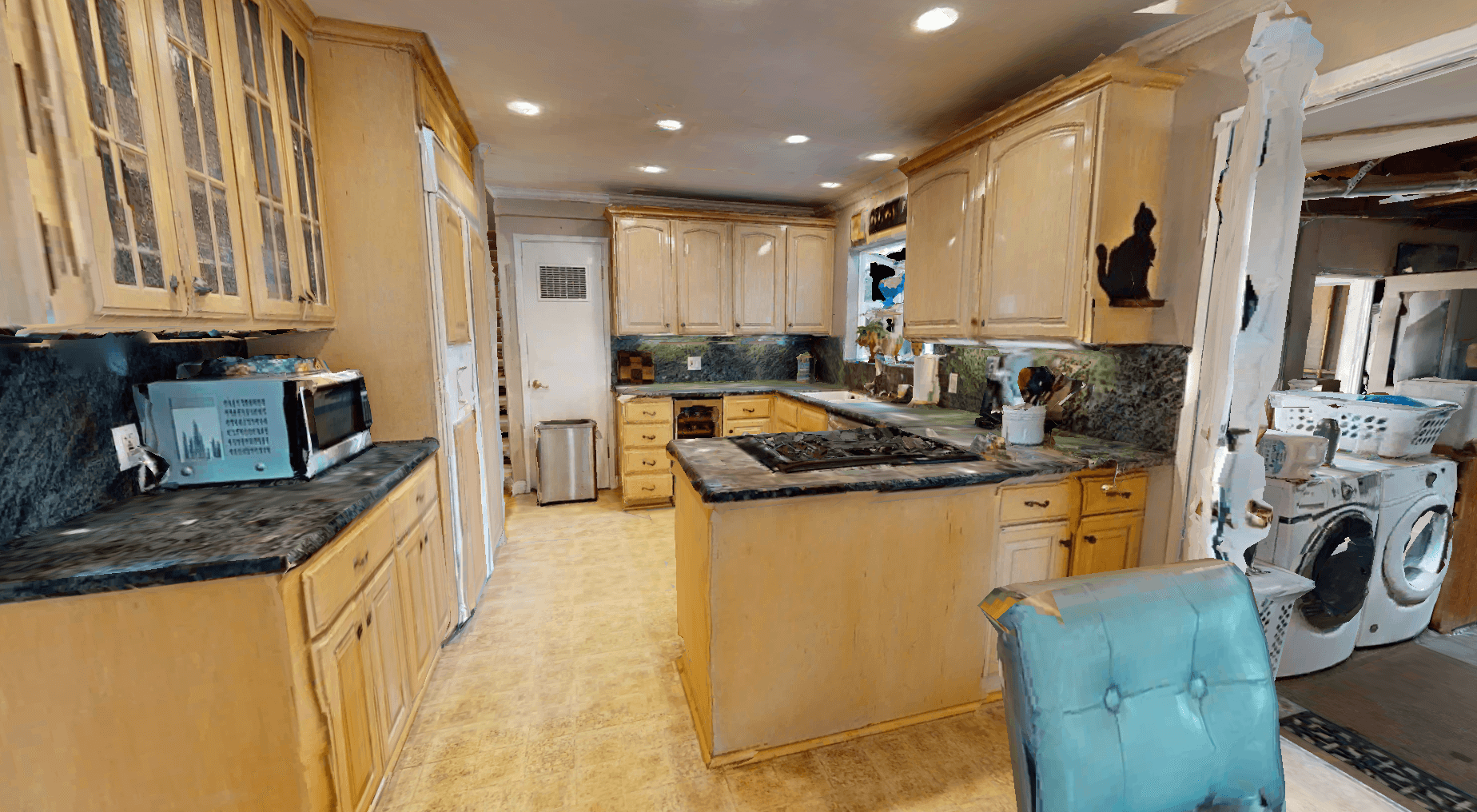} \\

\includegraphics[]{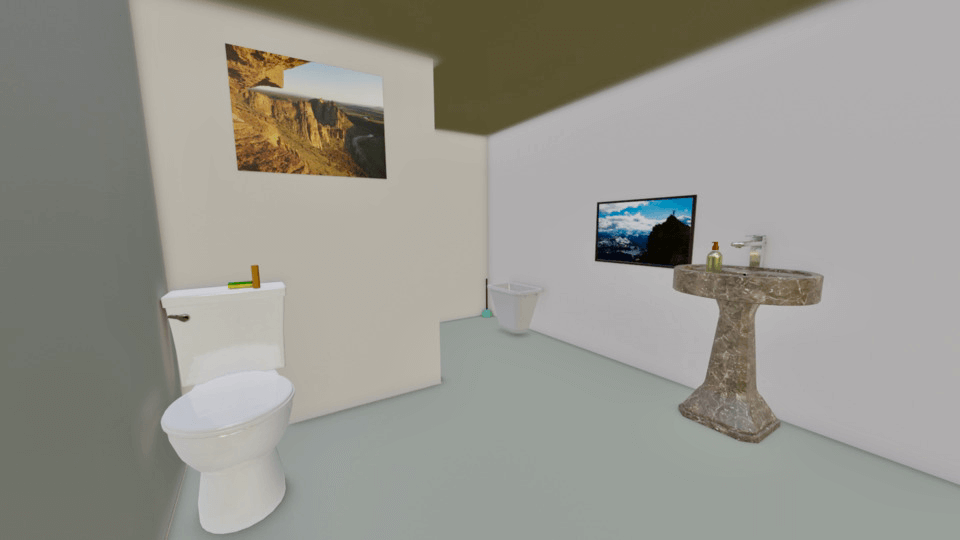} & \includegraphics[]{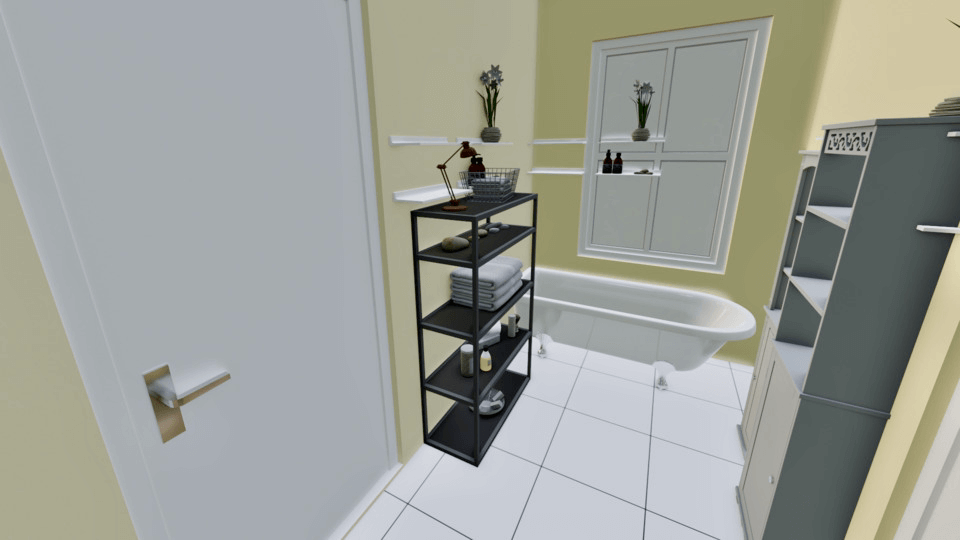} & \includegraphics[]{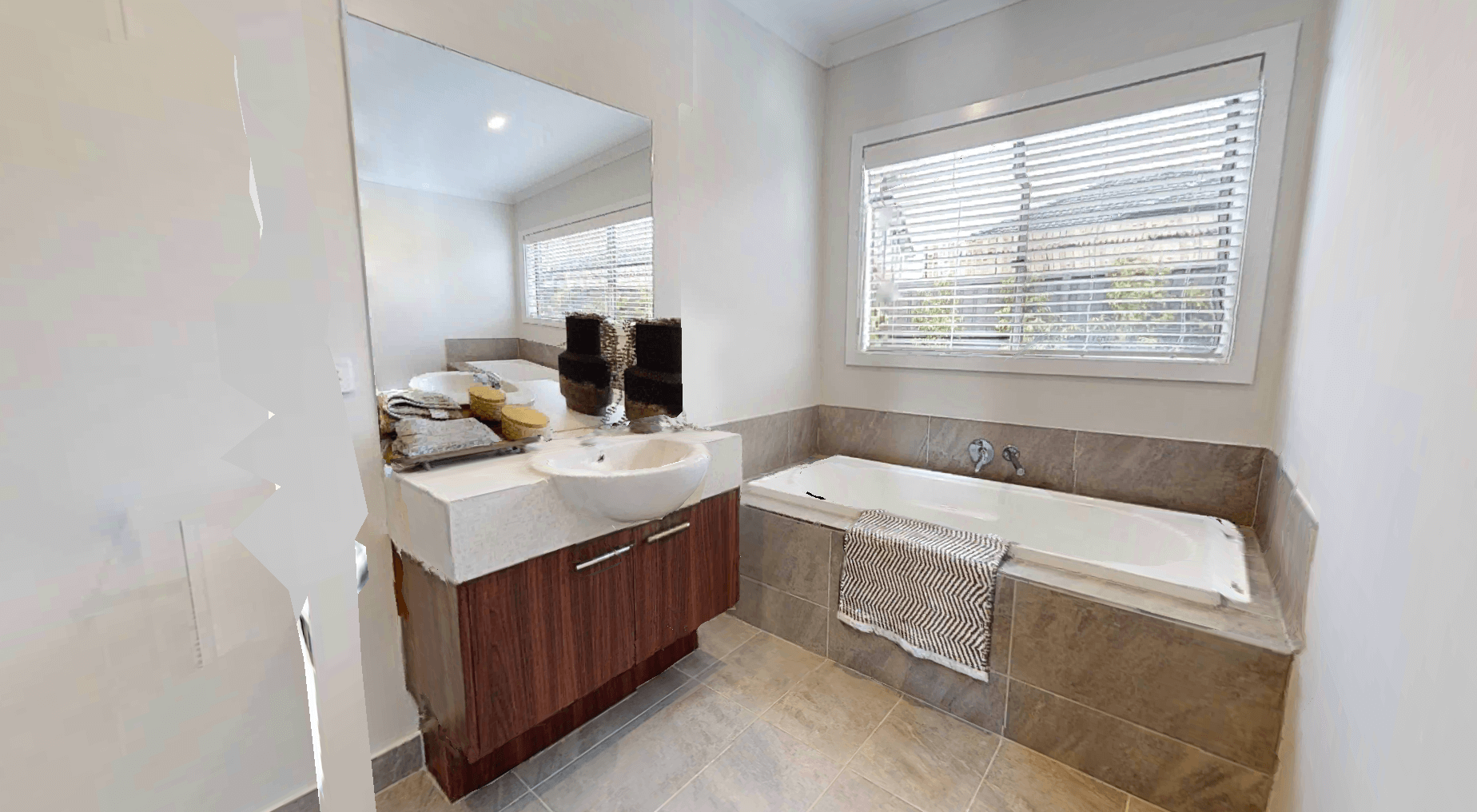} \\

\includegraphics[]{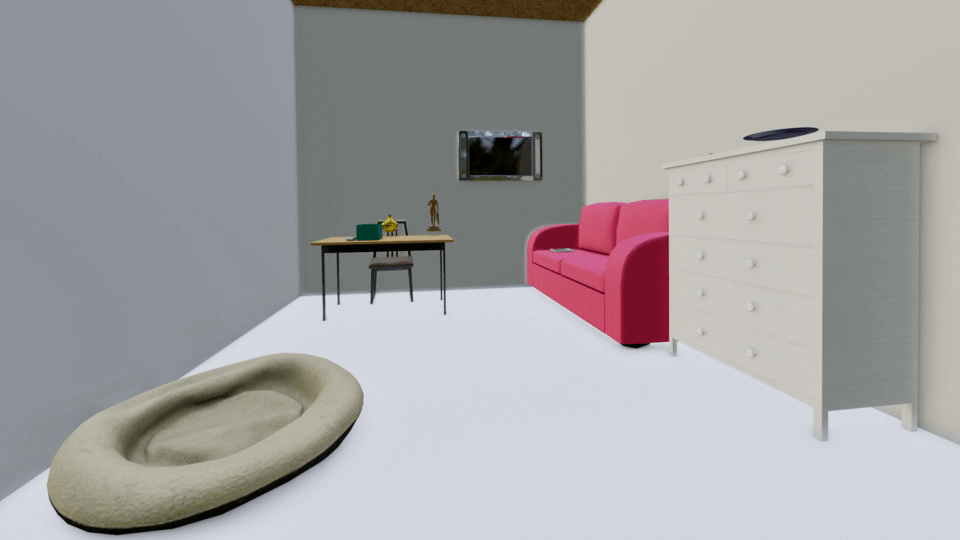} & \includegraphics[]{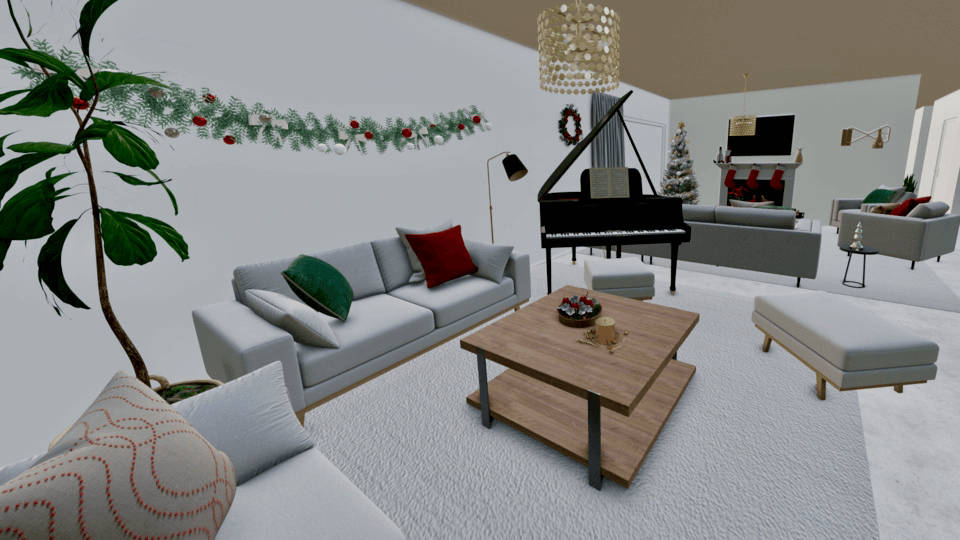} & \includegraphics[]{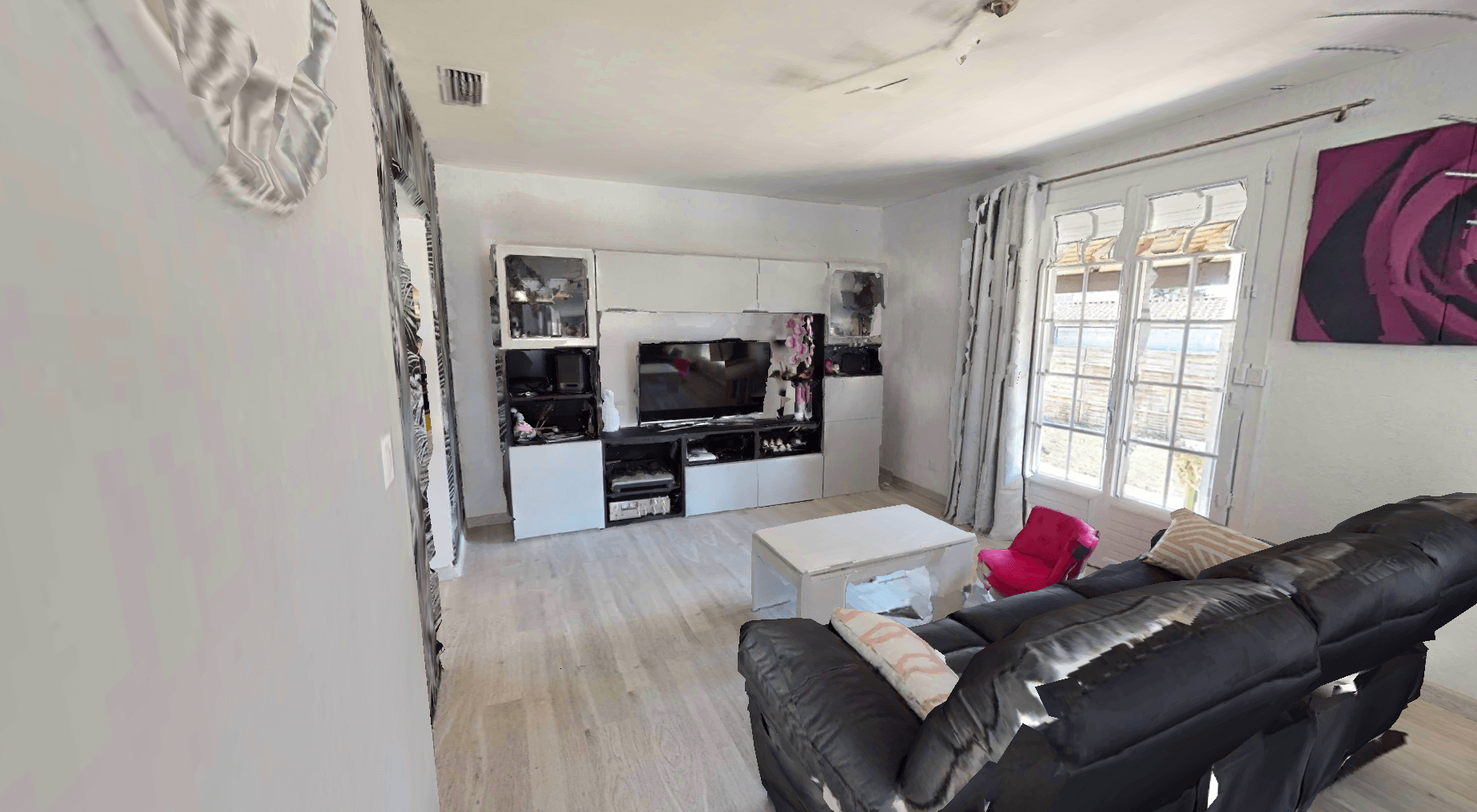} \\

\includegraphics[]{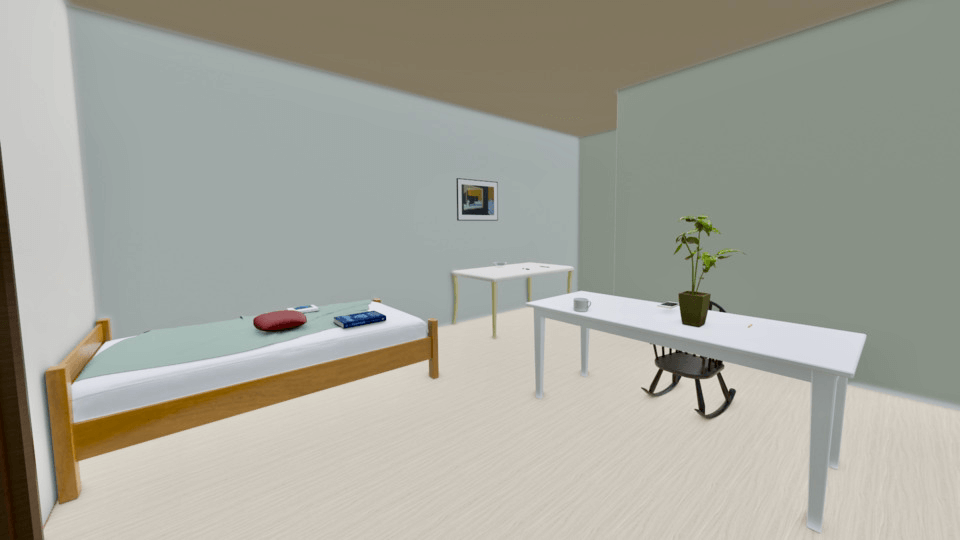} & \includegraphics[]{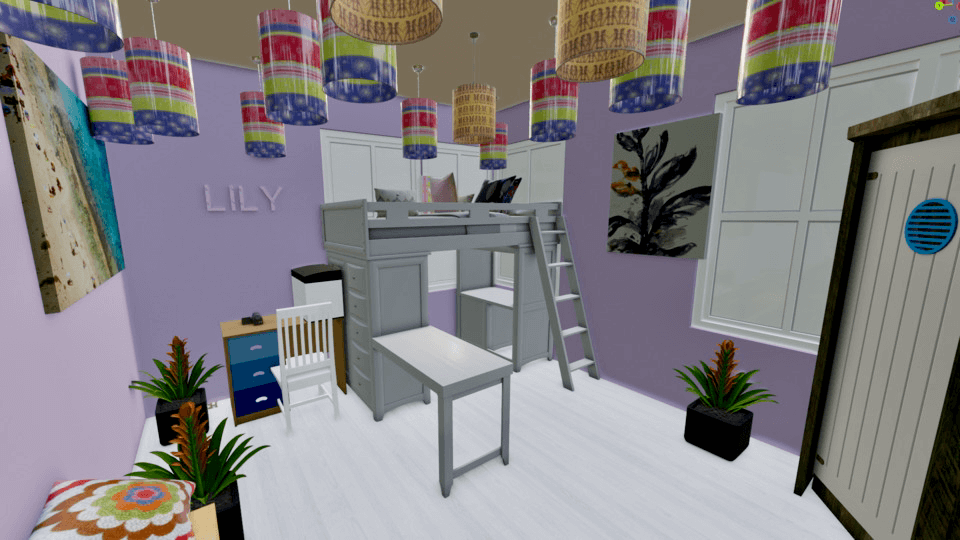} & \includegraphics[]{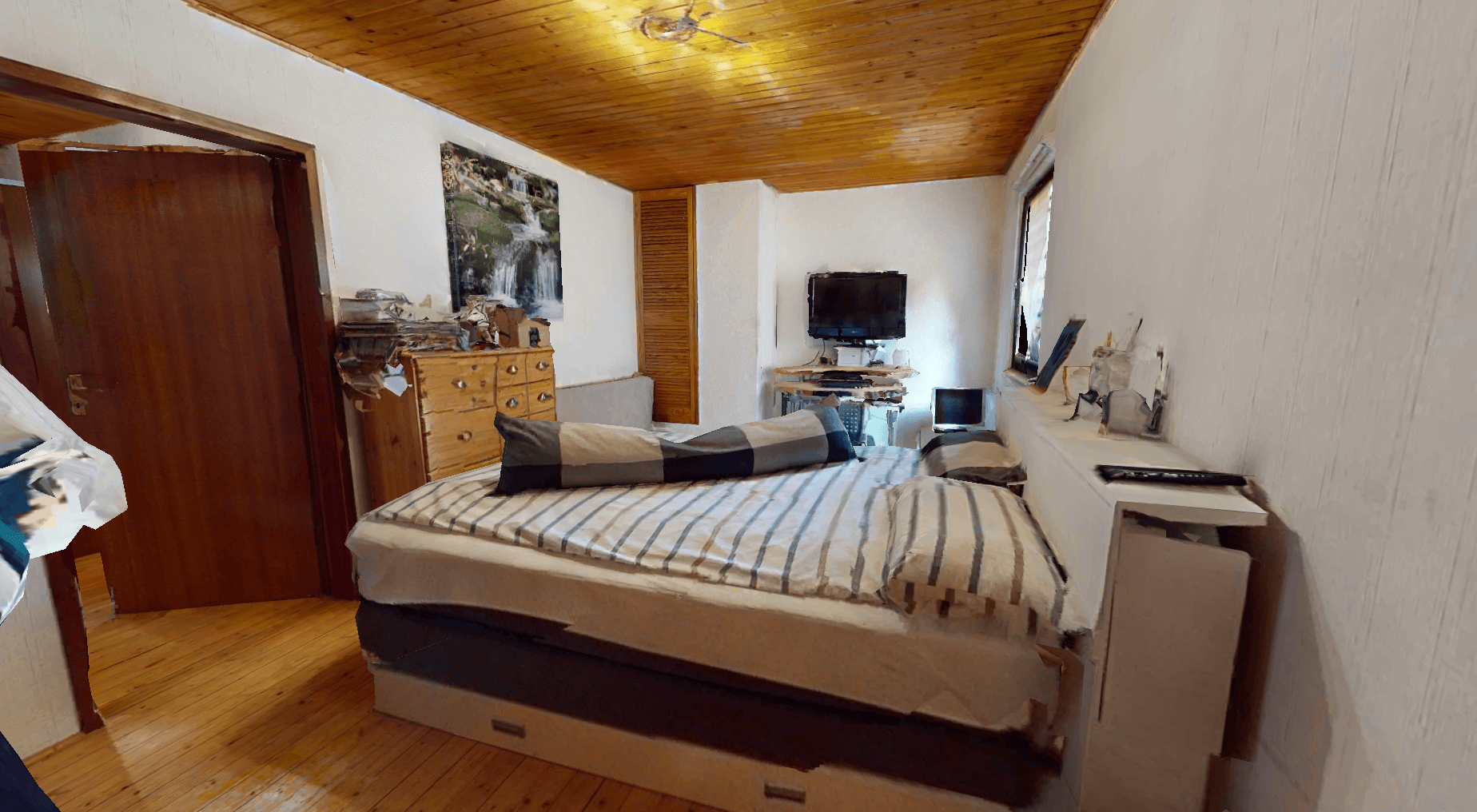} \\

\includegraphics[]{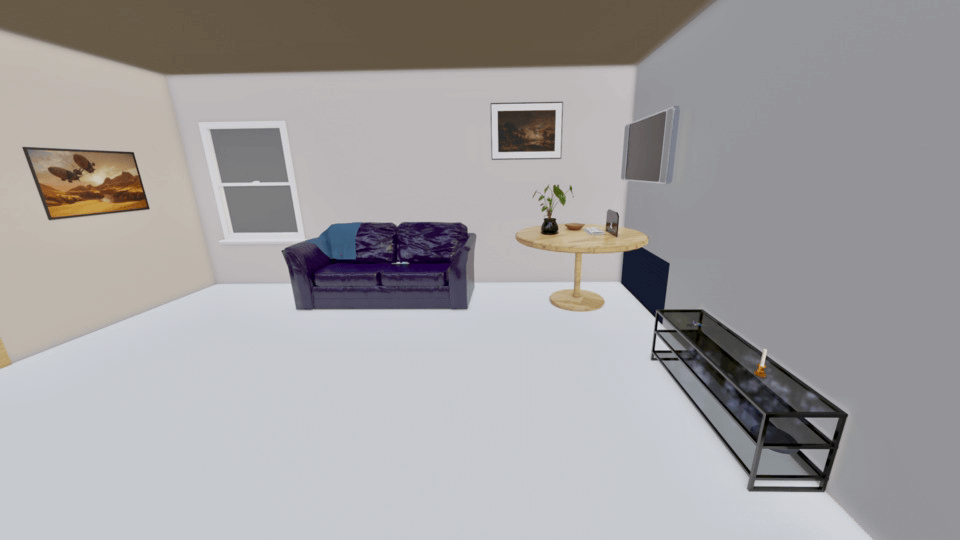} & \includegraphics[]{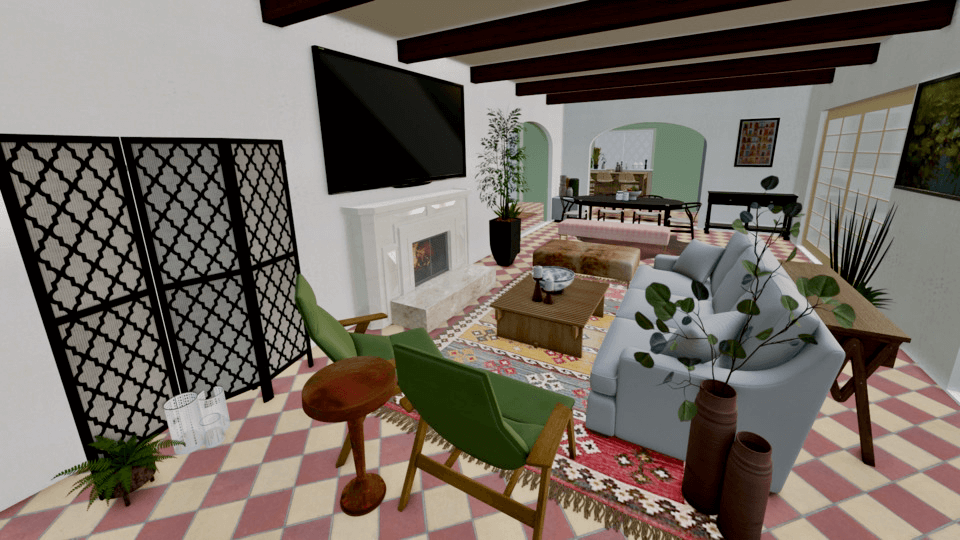} & \includegraphics[]{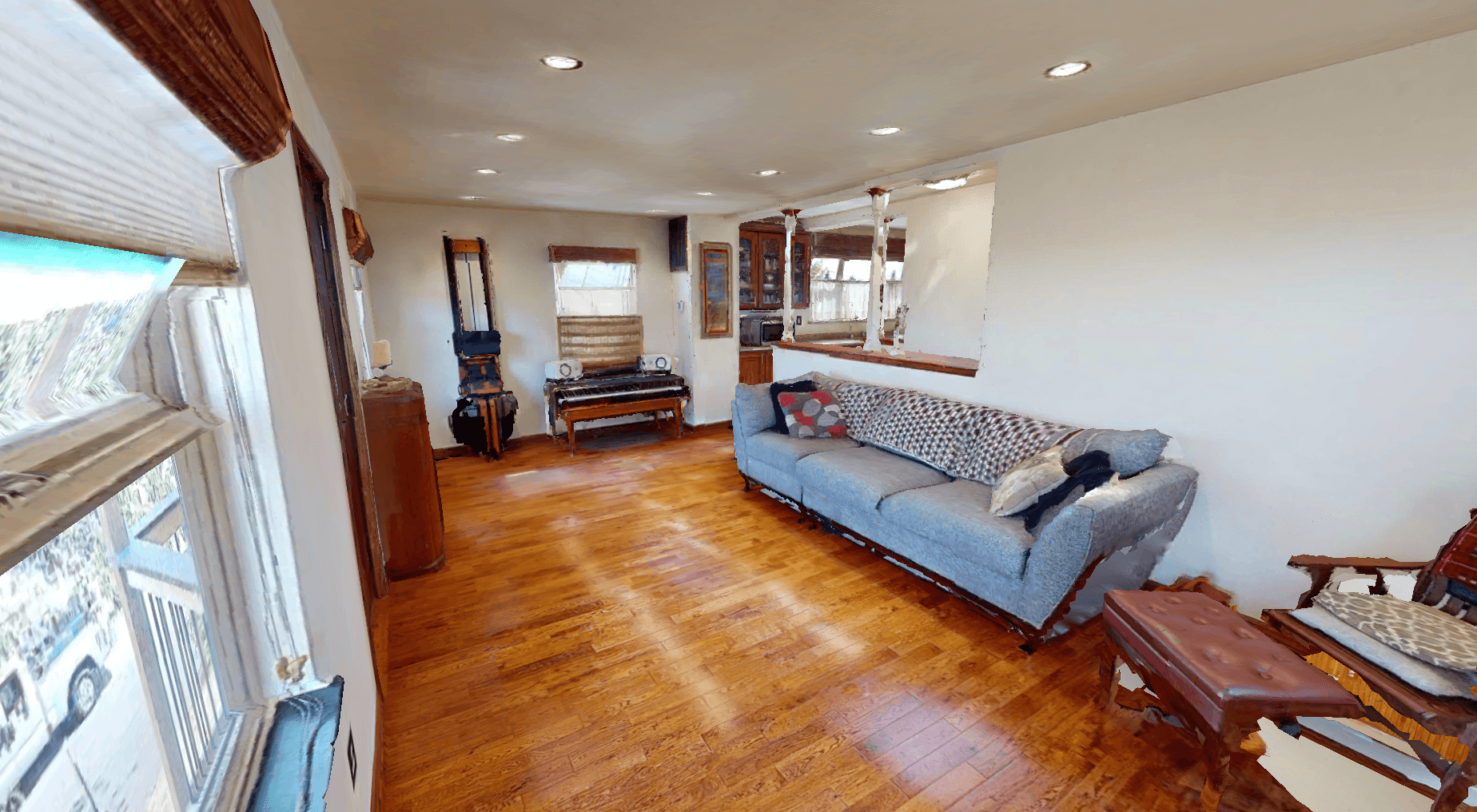} \\

\includegraphics[]{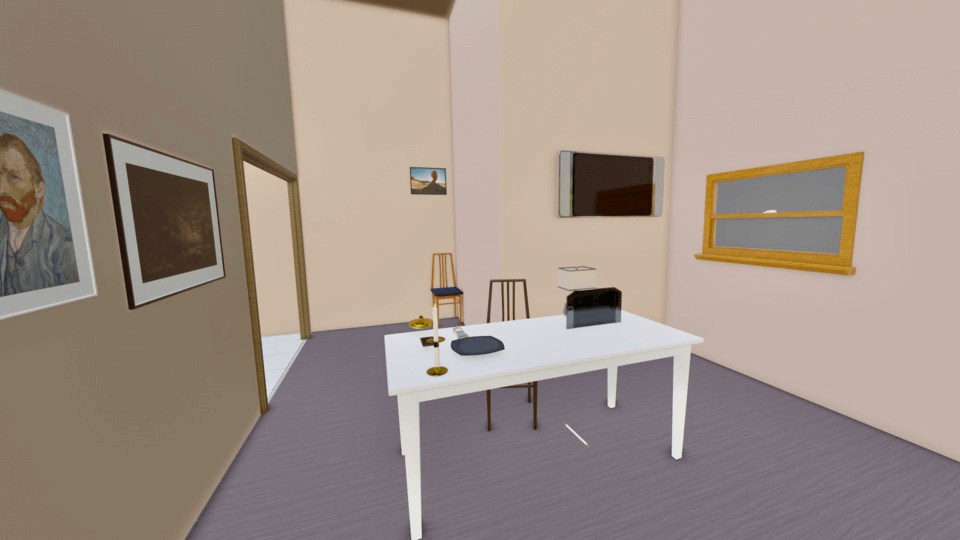} & \includegraphics[]{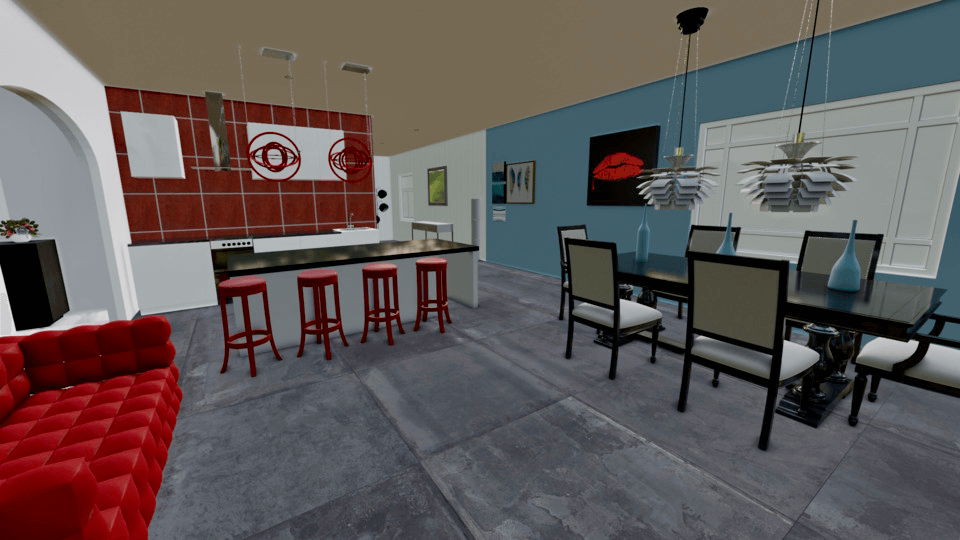} & \includegraphics[]{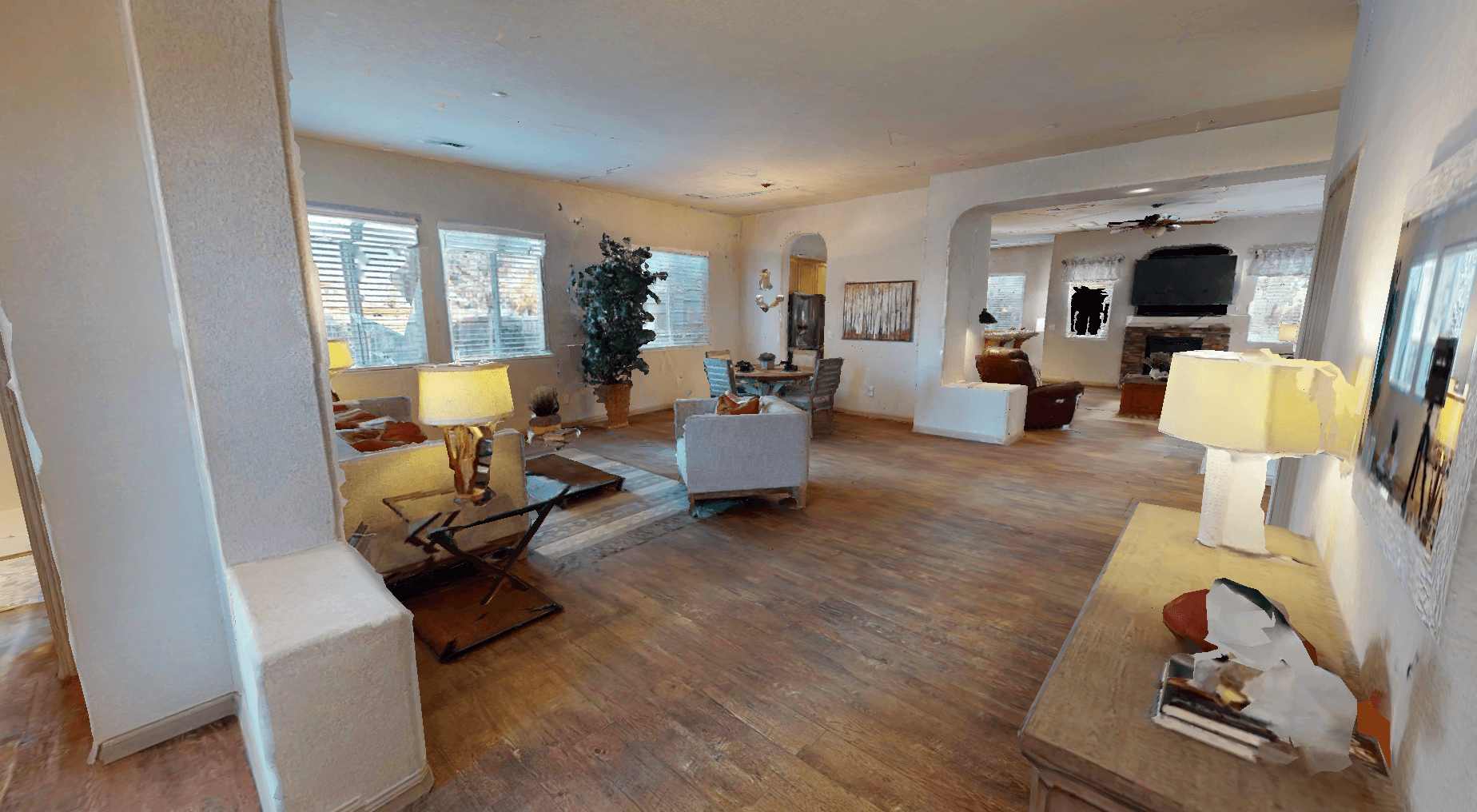} \\
\bottomrule
\end{tabularx}
\vspace{1pt}
\caption{
\textbf{Qualitative comparison of first-person views from ProcTHOR \cite{deitke2022procthor}, \ourdataset and HM3DSem \cite{yadav2022hm3dsem}.}
The \ourdataset scenes exhibit a richer diversity of objects and are more realistically populated than the ProcTHOR scenes.
Images are rendered using Blender's Eevee renderer with all parameters at default settings.
The ProcTHOR scenes and \ourdataset scenes are shaded with ambient occlusion and screen-space reflections, while the HM3DSem scenes are rendered without shading.
}
\label{fig:qual-firstperson}
\end{figure*}
\begin{figure*}
\setkeys{Gin}{width=\linewidth}
\begin{tabularx}{\linewidth}{@{}YYY@{}}
\toprule
 ProcTHOR \cite{deitke2022procthor} & \ourdataset & HM3DSem \cite{yadav2022hm3dsem} \\
\includegraphics[trim={5cm 0 5cm 0},clip]{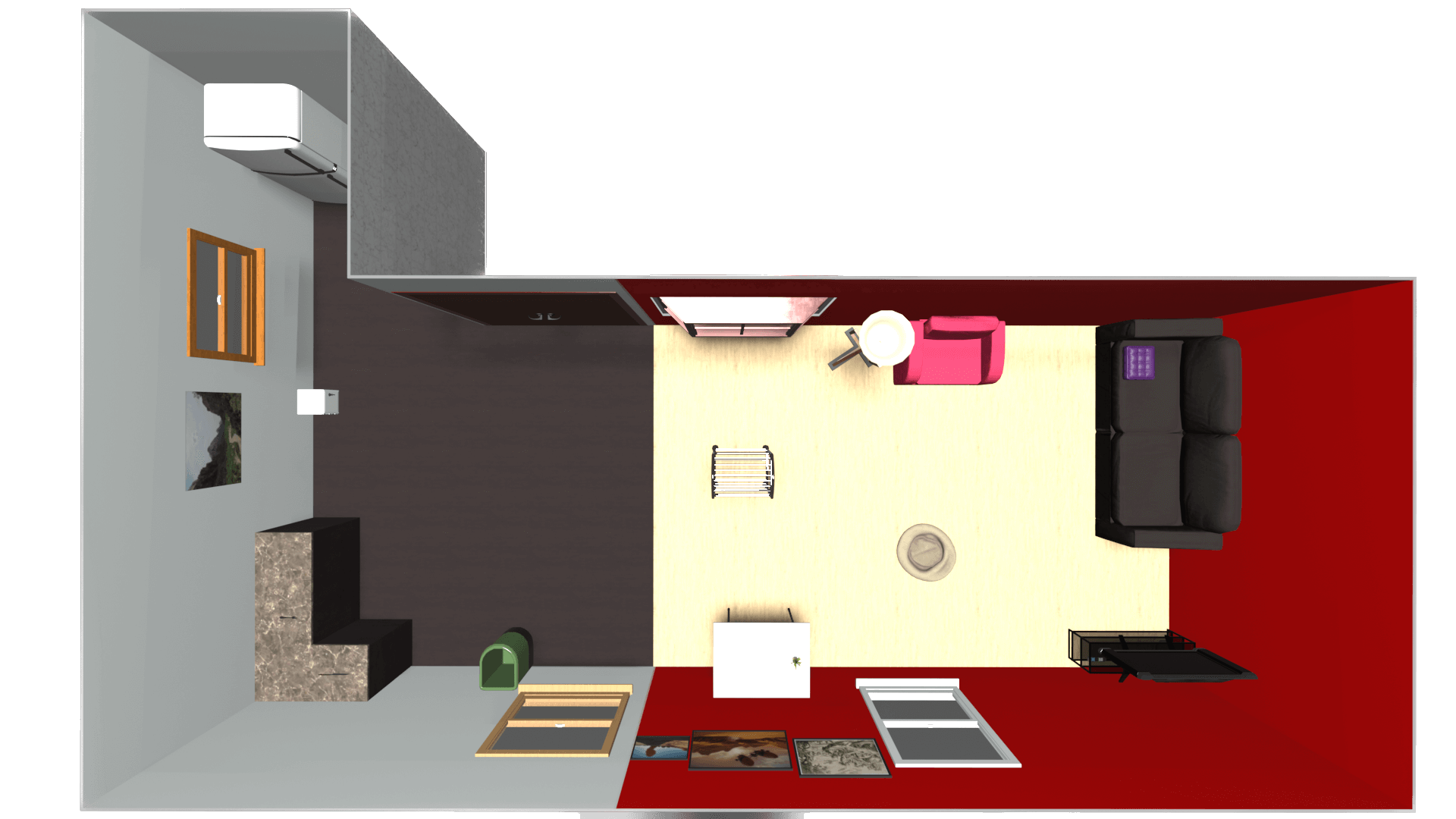} & \includegraphics[trim={5cm 0 5cm 0},clip]{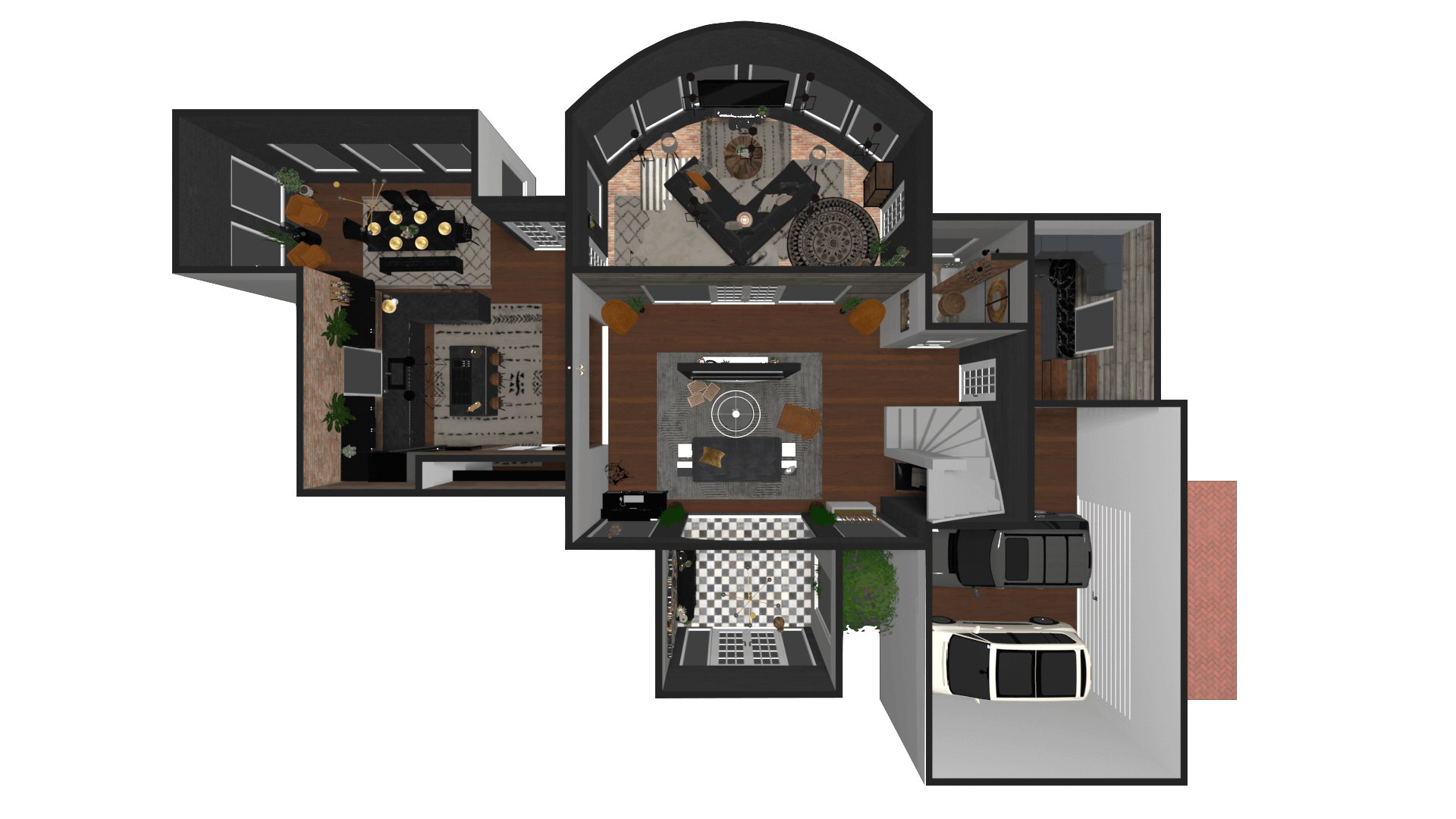} & \includegraphics[]{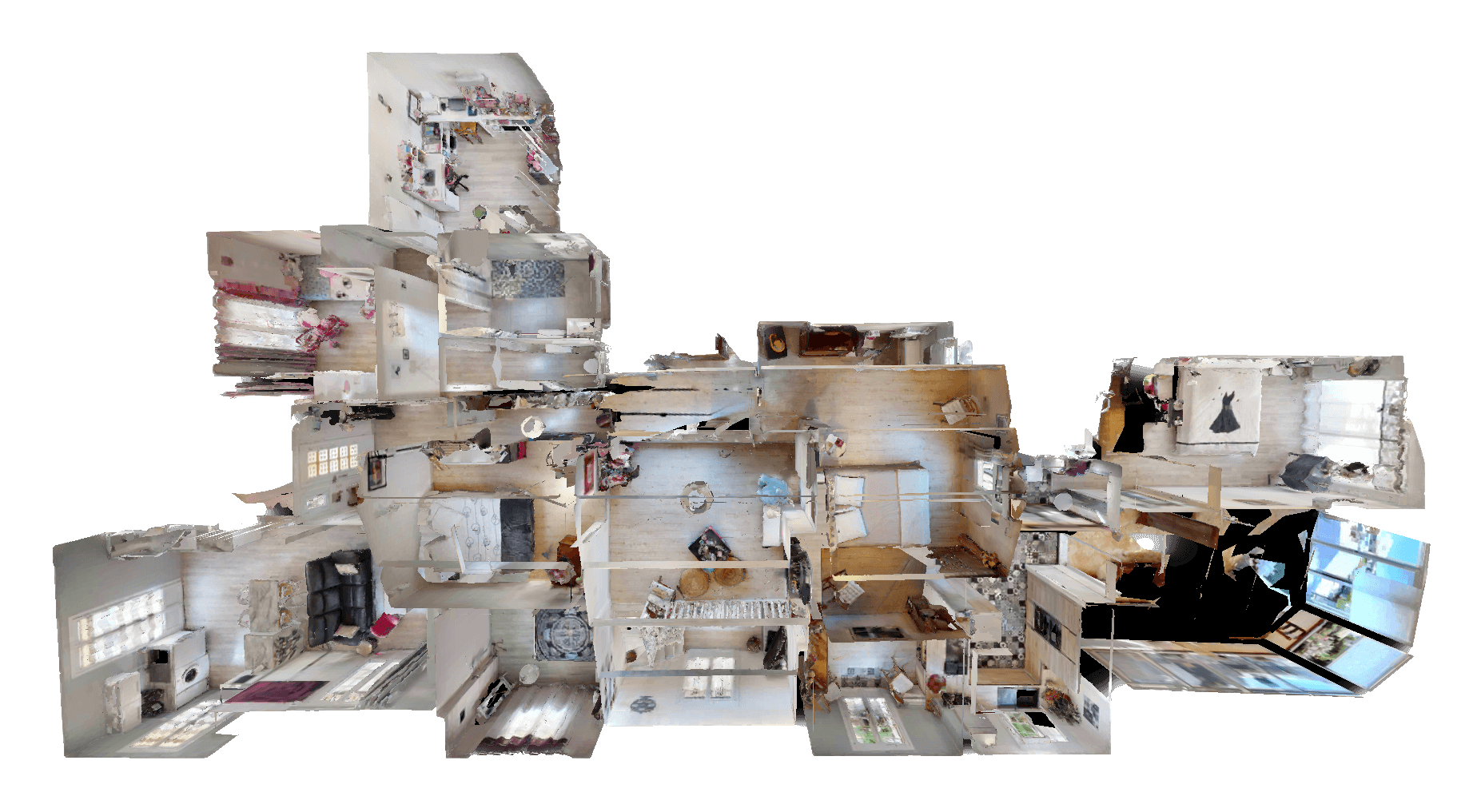} \\

\includegraphics[trim={10cm 0 10cm 0},clip]{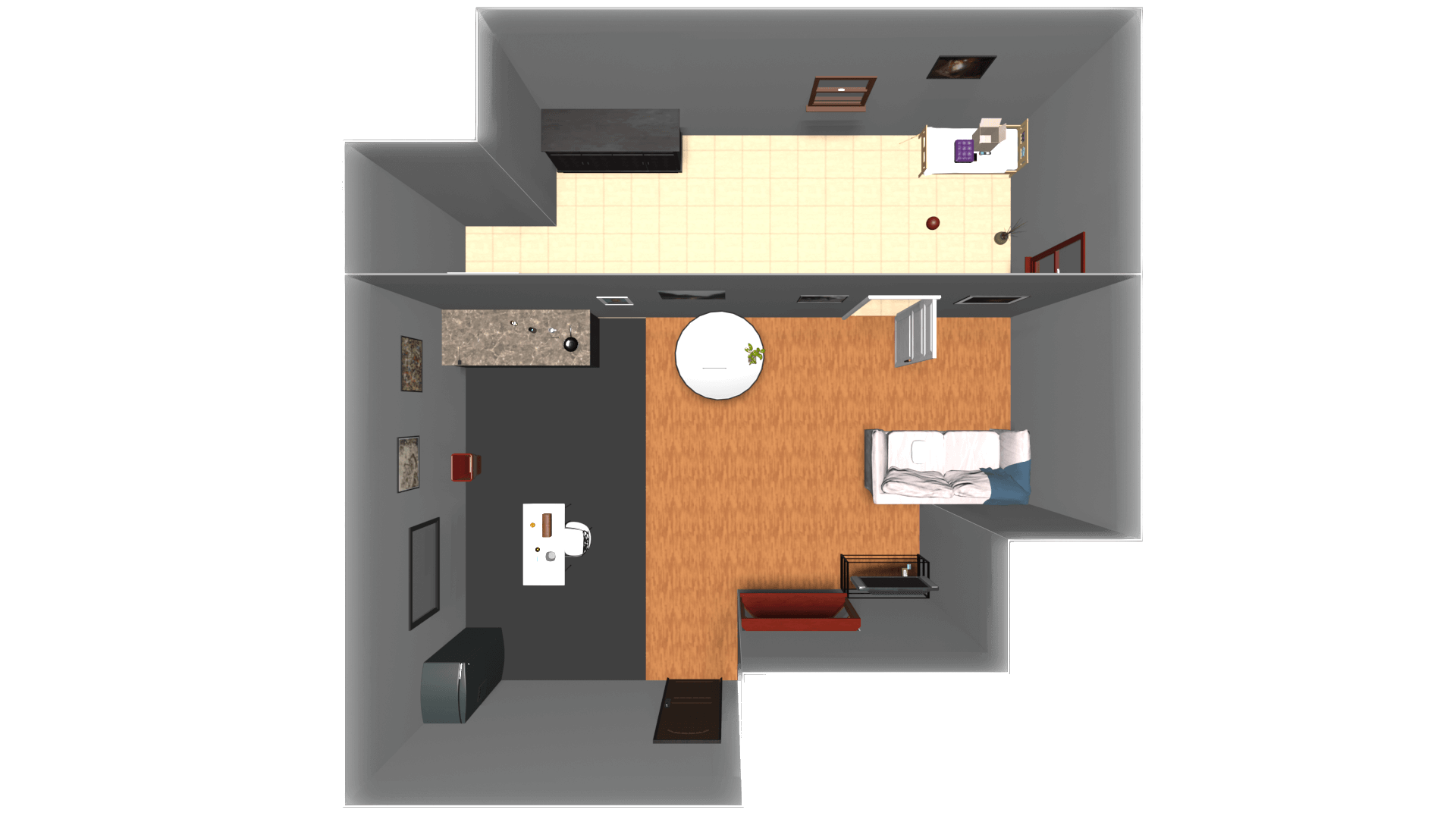} & \includegraphics[trim={7cm 0 7cm 0},clip]{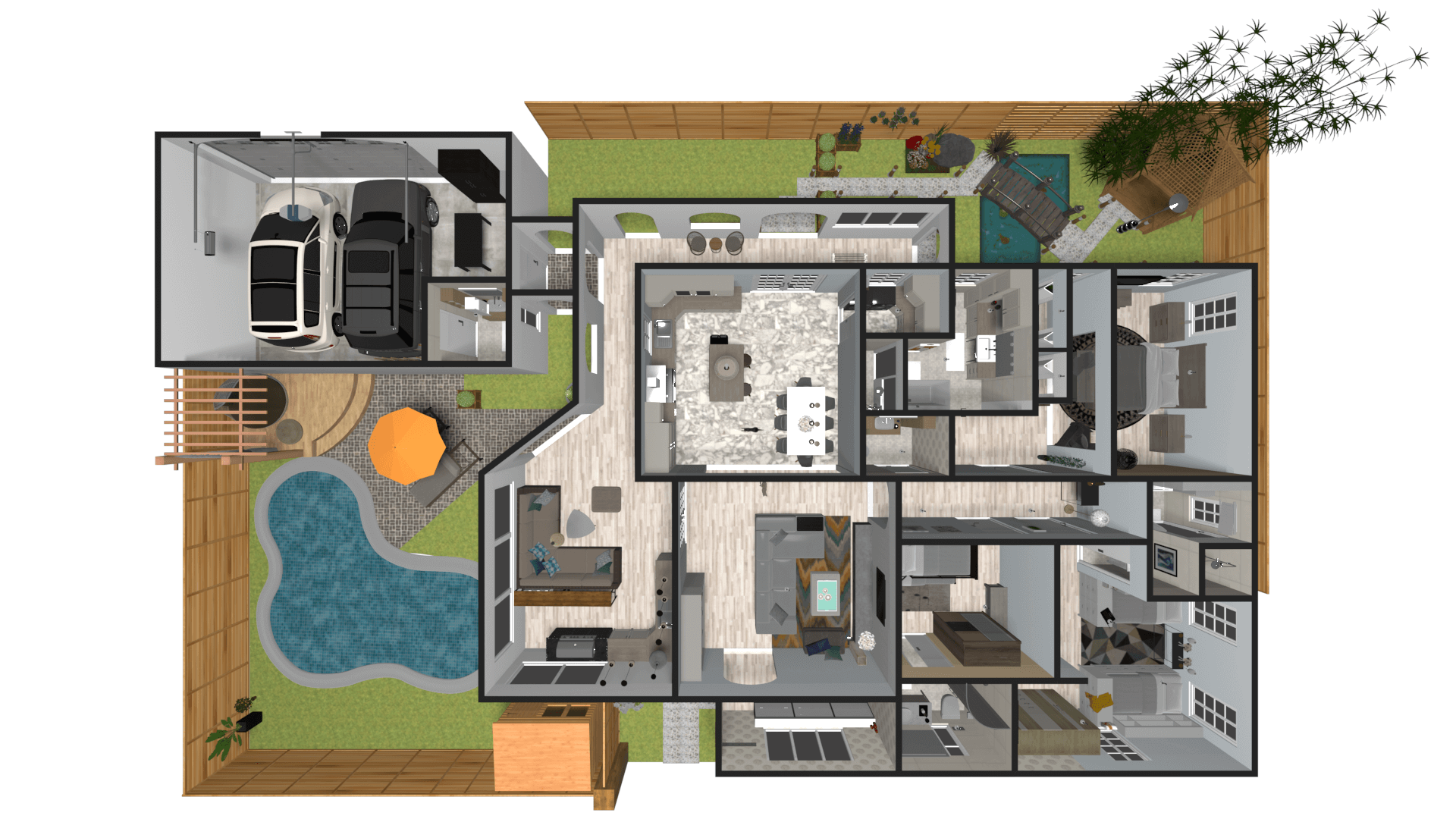} & \includegraphics[trim={7cm 0 3cm 0},clip]{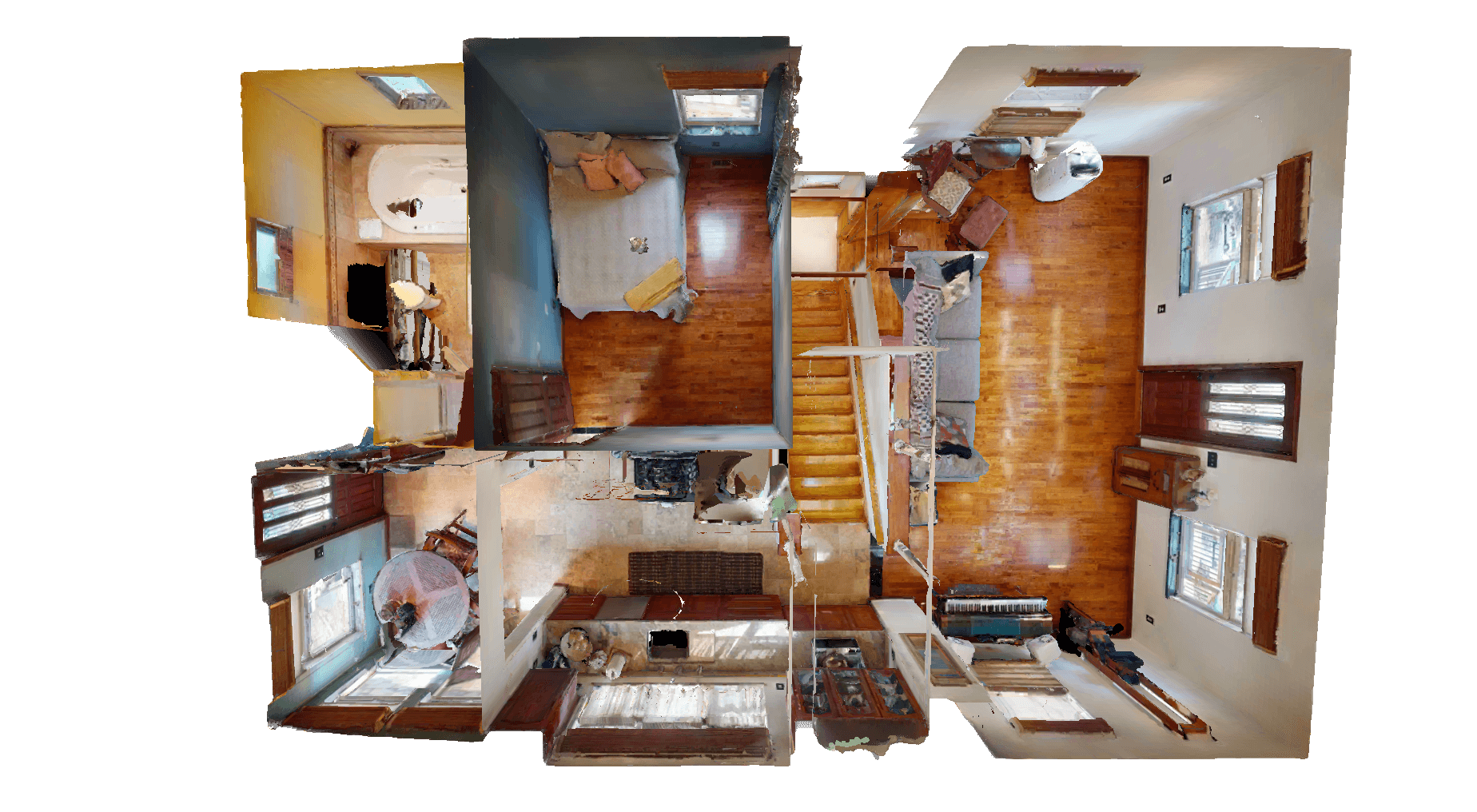} \\

\includegraphics[trim={7cm 0 0cm 0},clip]{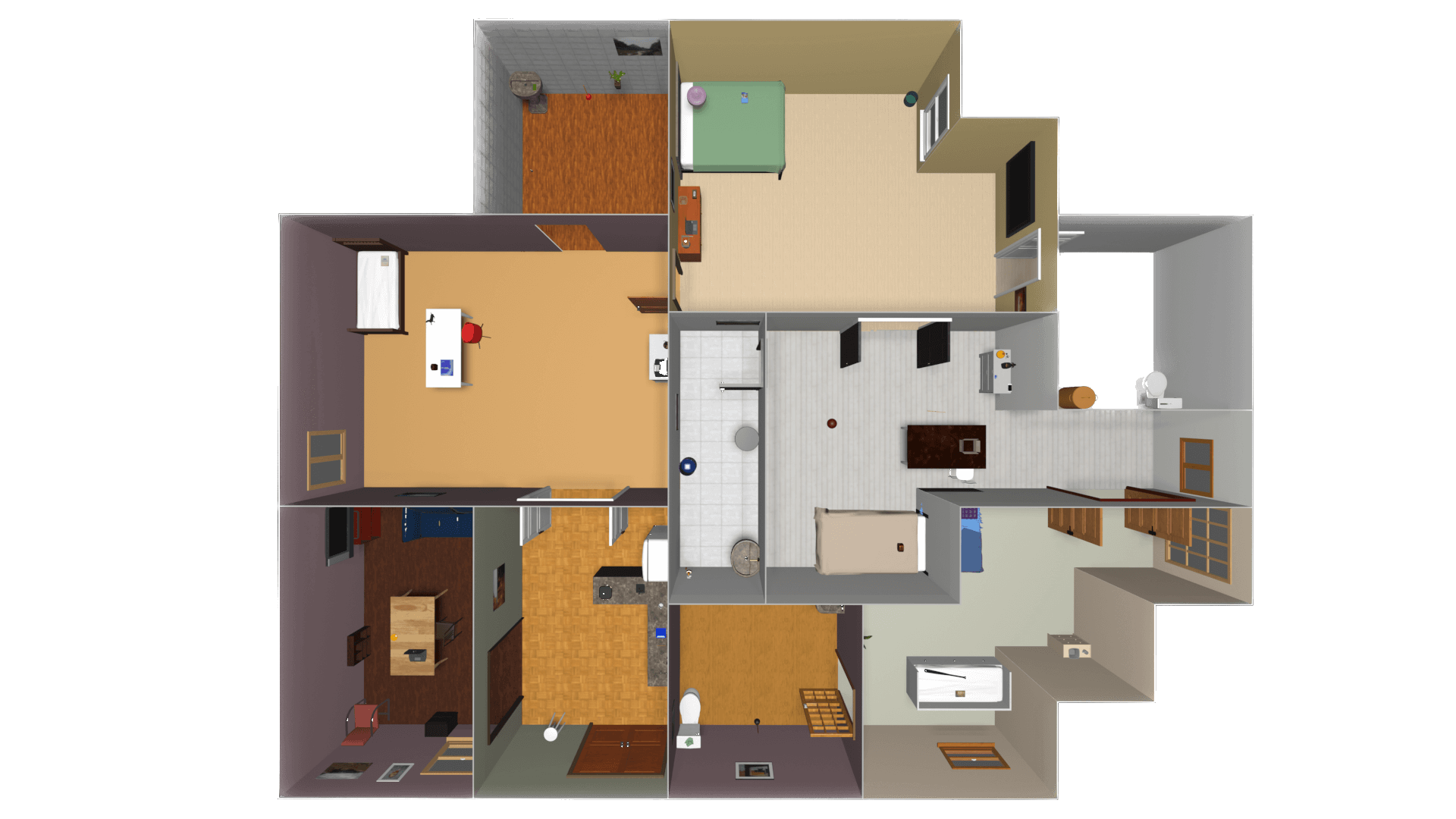} & \includegraphics[trim={5cm 0 5cm 0},clip]{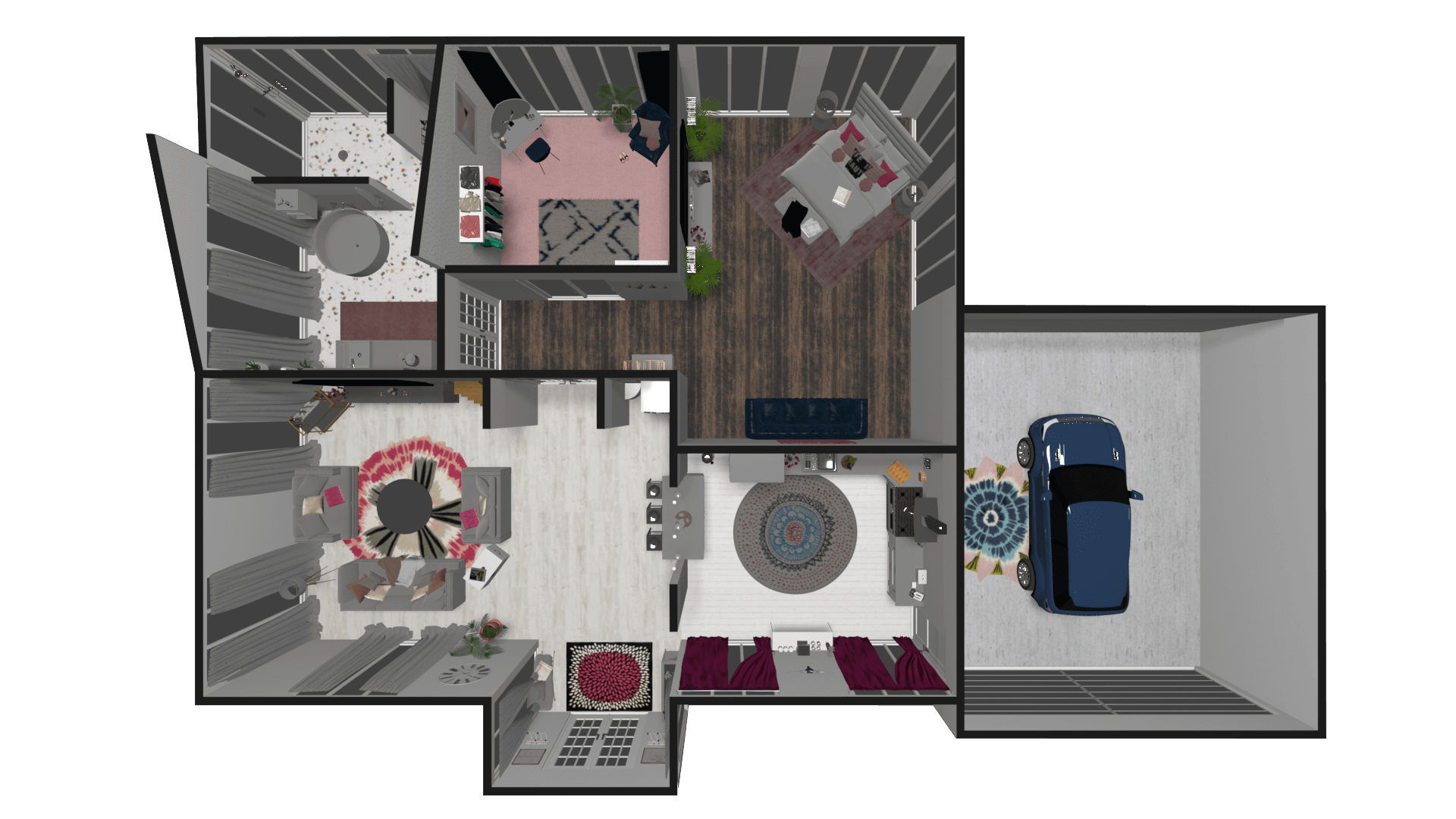} & \includegraphics[trim={10cm 0 10cm 0},clip]{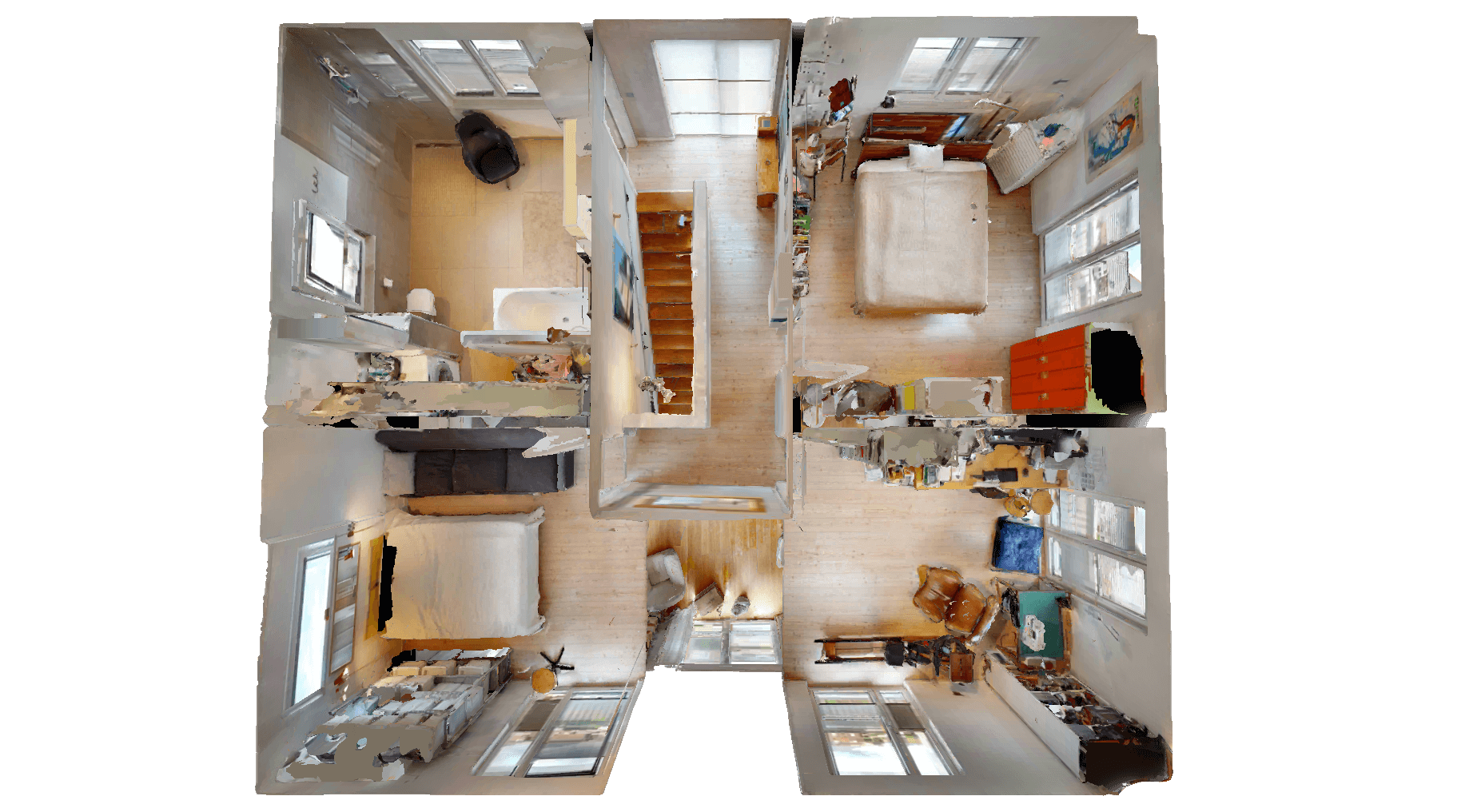} \\

\includegraphics[trim={10cm 0 10cm 0},clip]{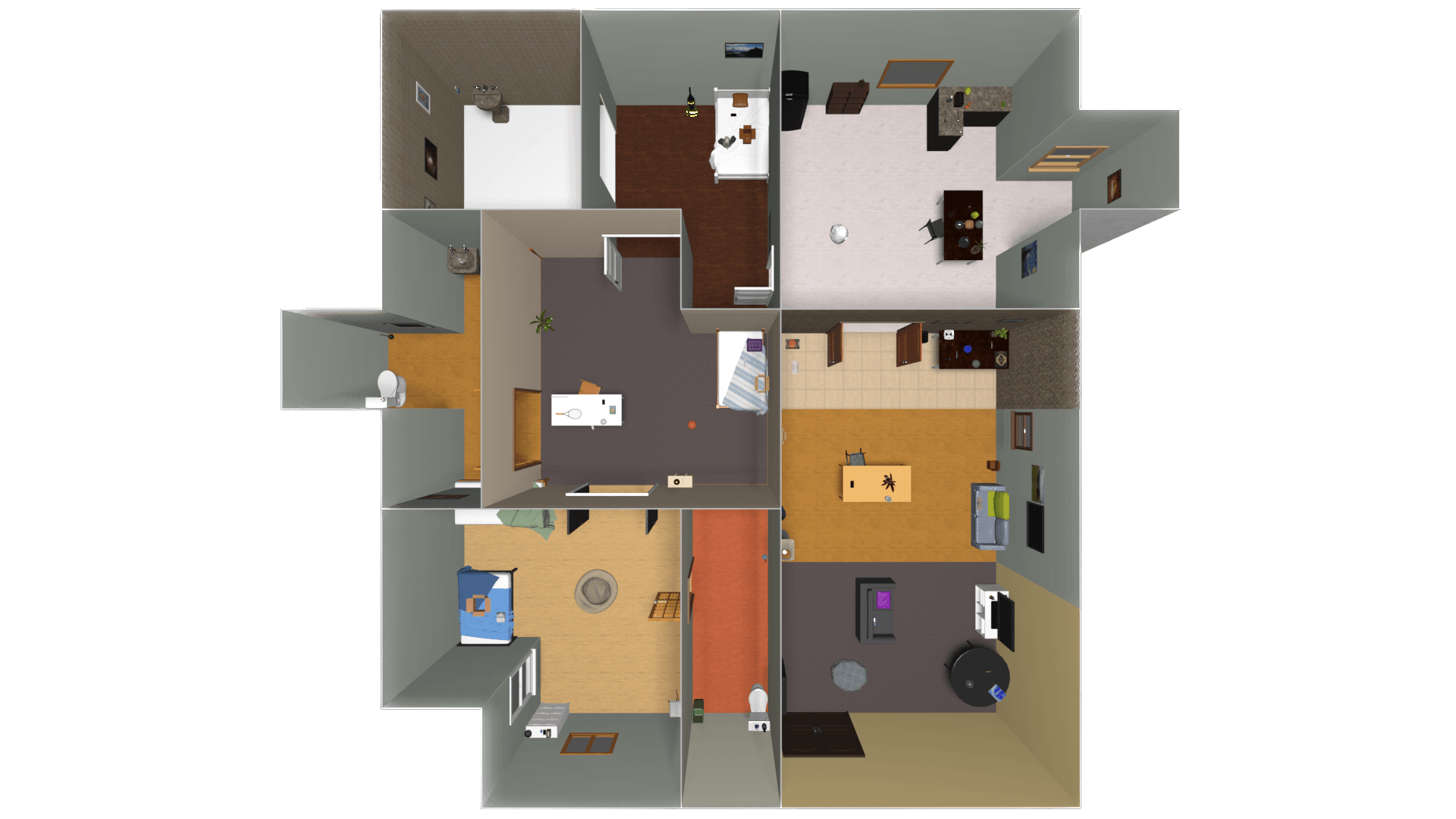} & \includegraphics[trim={10cm 0 10cm 0},clip]{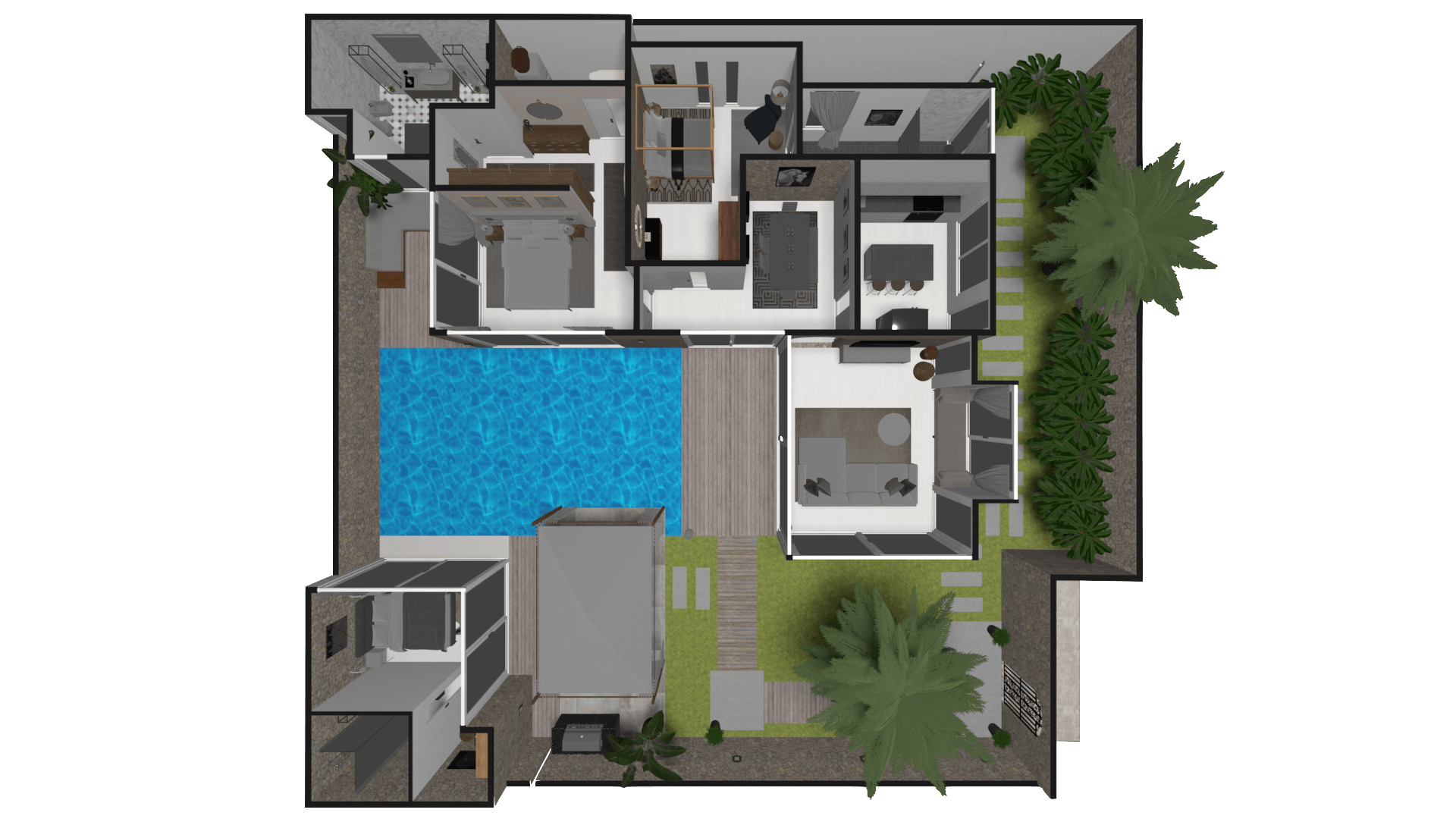} & \includegraphics[trim={10cm 0 10cm 0},clip]{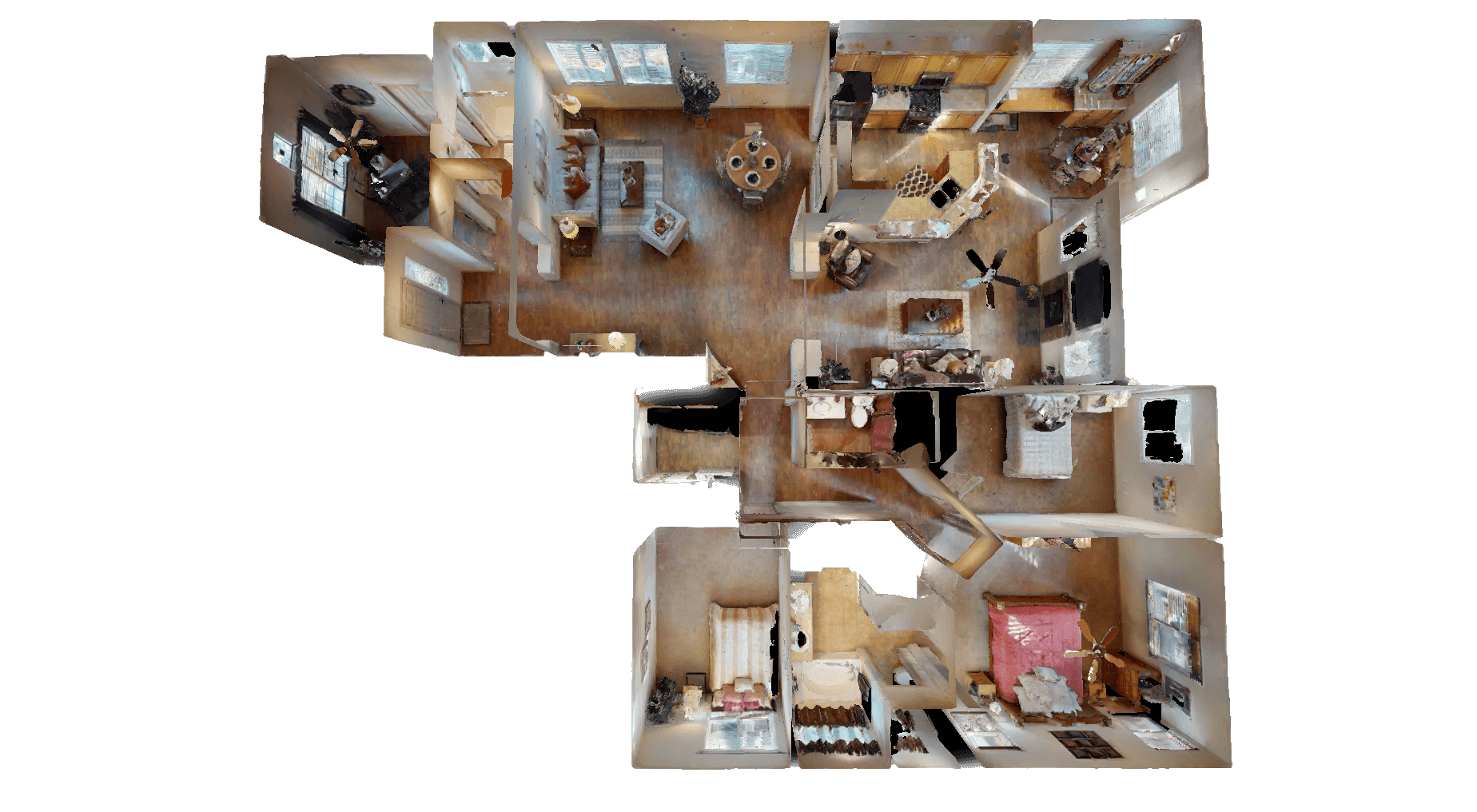} \\
\bottomrule
\end{tabularx}
\vspace{1pt}
\caption{
\textbf{Top-down views of scenes from ProcTHOR \cite{deitke2022procthor}, \ourdataset and HM3DSem \cite{yadav2022hm3dsem}.}
Compared to ProcTHOR, the \ourdataset scenes exhibit more realistic architectural layouts with corridors between rooms, non-rectilinear wall outlines and densely populated rooms.
These characteristics bring \ourdataset closer to real-world environments as captured in the HM3DSem dataset.
}
\label{fig:qual-topdown}
\end{figure*}

\subsection{\ourdataset qualitative visualizations}

In \Cref{fig:vis-fp-objects} we show example object instances for a number of categories from \ourdataset.
We see that \ourdataset exhibits a rich diversity of object geometry, apperance, and physical sizes across many categories.
This diversity is beneficial for experiments studying generalization of perception capabilities for embodied AI agents.
These objects also populate the scenes in \ourdataset in a way that produces more realistic environments.
We show first-person views in \Cref{fig:qual-firstperson} and top-down views in \Cref{fig:qual-topdown}.
Overall, we see that \ourdataset scenes exhibit more realistic architectural layouts than ProcTHOR \cite{deitke2022procthor}, and come closer to real-world scans from HM3DSem \cite{yadav2022hm3dsem} in terms of the richness and density of objects populating each room.

\subsection{AI2-THOR datasets in Habitat}
\label{sec:supp-data-a2}

To construct rigorous experiments comparing between the \ourdataset and ProcTHOR \cite{deitke2022procthor} datasets, we port and optimize ProcTHOR assets in the same fashion as \ourdataset so that they can be efficiently used in the Habitat simulator platform \cite{savva2019habitat}.
We built on top of the AI2-THOR Unity interface \cite{kolve2017ai2} to export all Unity prefab objects and scene assets to glTF format using UnityGLTF\footnote{\href{https://github.com/KhronosGroup/UnityGLTF}{github.com/KhronosGroup/UnityGLTF}}.
We also export a corresponding JSON-format metadata file with each asset to record information such as the semantic category label, position and orientation of the object.
For iTHOR \cite{kolve2017ai2}, RoboTHOR \cite{deitke2020robothor} and ArchitecTHOR, we filter the structural objects in the scene so that we leave only architectural objects (walls, floors, ceilings).
We then zero-center all exported objects, and re-orient the objects to standardized object-centric coordinates.
Subsequently, we can use the position and orientation information to correctly place the objects wherever they are observed in the original scene.
Note that we re-use assets across scenes to reduce on-disk and in-memory size.
Since ProcTHOR \cite{deitke2022procthor} has procedurally generated architectures, we construct the geometry of the architecture with the specified textures from the ProcTHOR scene layout specification JSON format, and create a glTF asset for each scene architecture.
All doors are exported using the AI2-THOR Unity interface in opened state to allow for navigation between rooms.
This porting of the AI-2THOR assets to Habitat format enables us to take advantage of the faster simulation speeds provided by Habitat~\cite{szot2021habitat} and run experiments with any combination of iTHOR, RobotTHOR, ArchitectTHOR, and ProcTHOR scenes.

\begin{figure}
\centering
\includegraphics[width=\linewidth]{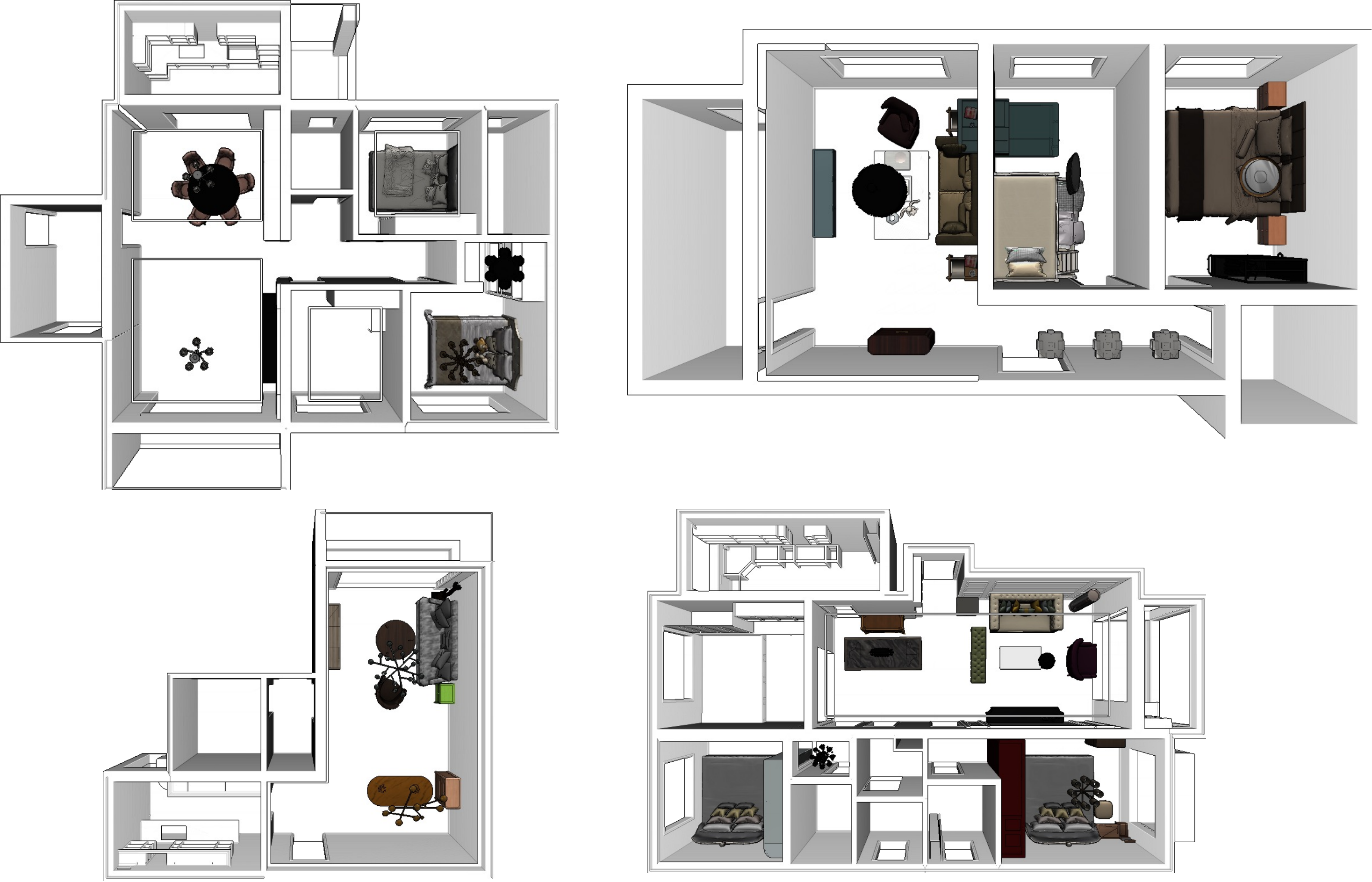}
\caption{
\textbf{3D-FRONT~\cite{fu20203dfront} scenes are sparsely populated.}
Algorithmic object replacement was used to place object instances, and some room types are unfortunately left empty (e.g., kitchens, bathrooms, closets).
This is due to limited rights to release the original 3D assets used in the scenes.
}
\label{fig:3df-top-down}

\end{figure}

\begin{figure}
\centering
\includegraphics[width=\linewidth]{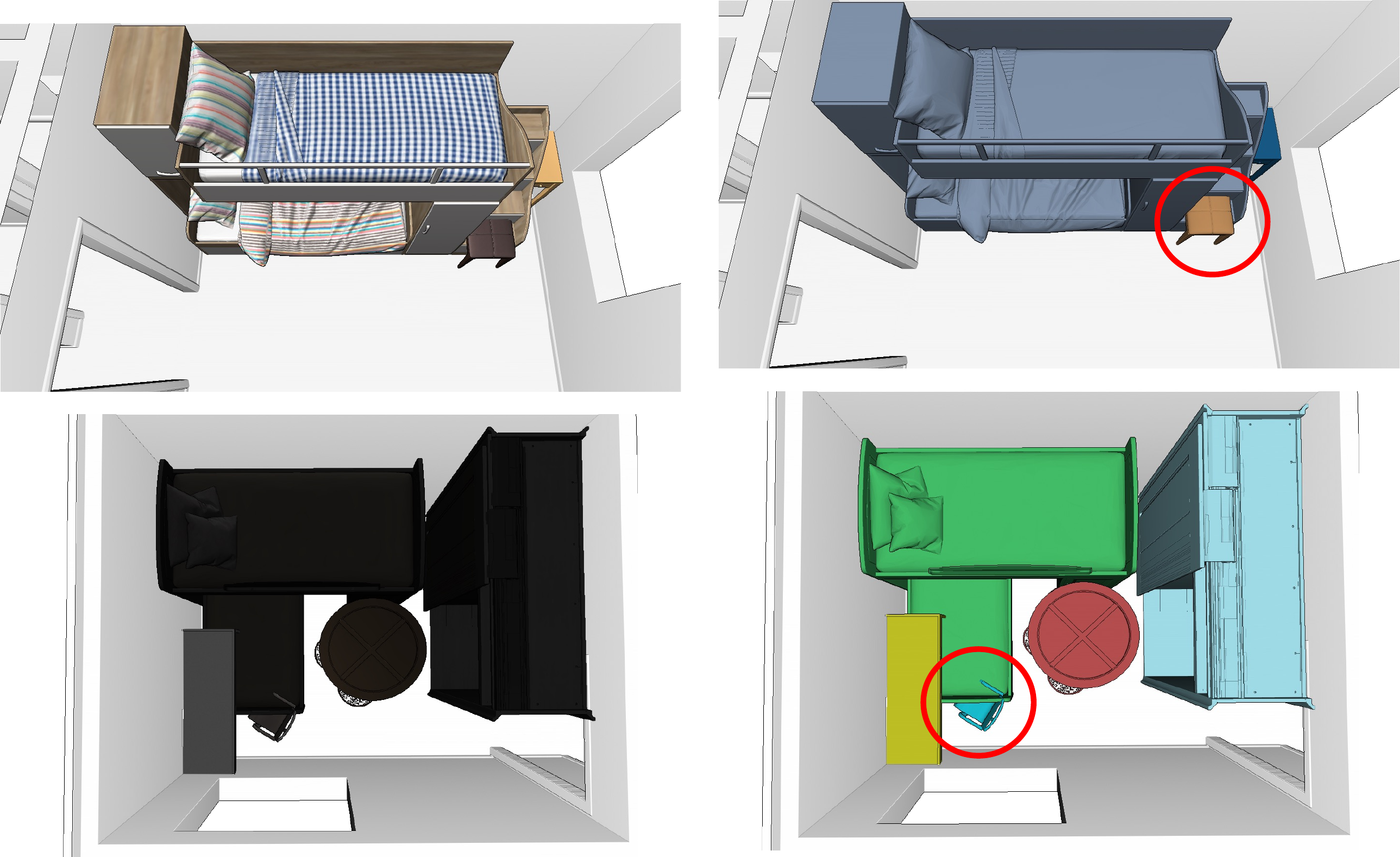}
\caption{
\textbf{Object inter-penetrations in the 3D-FRONT dataset.}
The algorithmic object replacement unfortunately produces cases such as the ones shown here (see red circles).
We show the scene with original object textures (left) and semantically colored by object category (right).
}
\label{fig:3df-overlap}
\end{figure}

\subsection{Why not use 3D-FRONT?}

3D-FRONT~\cite{fu20203dfront} is a popular dataset for 3D scene generation research.
However, the scenes are sparsely populated.
Due to limited rights in releasing the original 3D assets for the scenes, 3D models in the scenes for the 3D-FRONT dataset are replaced with 3D models from the 3D-FUTURE~\cite{fu20203dfuture} dataset.
This limited the presence of object in rooms to a subset of room categories (e.g., living rooms, dining rooms, bedrooms) and left other rooms (e.g., kitchens, bathrooms, and closets) empty.
There is also a limited number of placements of smaller objects on top of larger furniture objects (other than un-decomposed multi-object combinations), and no wall decoration objects.
Another impact of using replaced assets is that there are scenes with objects that are interpenetrating each other.
All of these factors make the 3D-FUTURE dataset much less well-suited for semantic indoor navigation where we expect the scenes to be well-populated, and objects to be realistically placed.

\section{Episode generation details}
\label{sec:supp-episode-gen}

\mypara{Viewpoint sampling.}
In each scene, for each indoor goal object instance (e.g., bed, chair, couch), we first sample a grid of prospective viewpoints around the object and then reject viewpoints that do not have a navigable position below them, those that are too far (distance > 1m) from the object's bounding box, and those that are outdoors or outside the house (in \ourdataset scenes). After snapping the valid viewpoints from the previous step to a nearby navigable position, we remove all the viewpoints snapped to small (radius < 1.5m) navigable islands (e.g., found on tabletops, counters, beds). Next, we spawn the agents on these valid viewing positions (facing the object) and compute the object's visibility at these positions by measuring the frame coverage, i.e. the fraction of image pixels belonging to object. We reject positions from where frame coverage is less than 0.1\%.  After getting valid viewpoints, we uniformly sample random starting positions (one per episode) across the scene such that: 1) the geodesic distance between each starting position and the goal instance is at least 1 meter but less than 30 meters; and 2) a fraction of the nearly straight-line episodes (ratio of geodesic distance to Euclidean distance < 1.05) are rejected. 

\begin{figure}
\centering
\setkeys{Gin}{width=\linewidth}
\noindent
\begin{tabularx}{\linewidth}{@{}YYYY@{}}
ProcTHOR & \ourdataset \\
\includegraphics[]{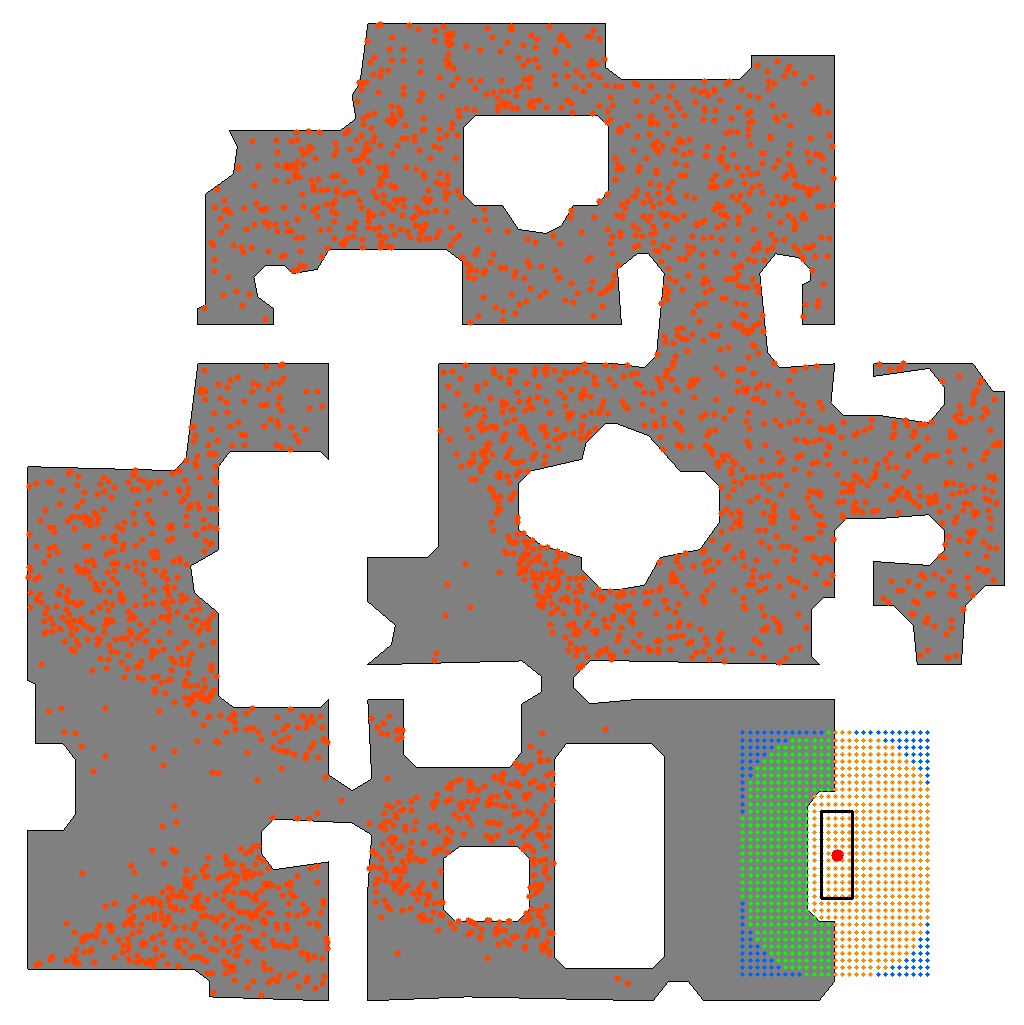} & \includegraphics[]{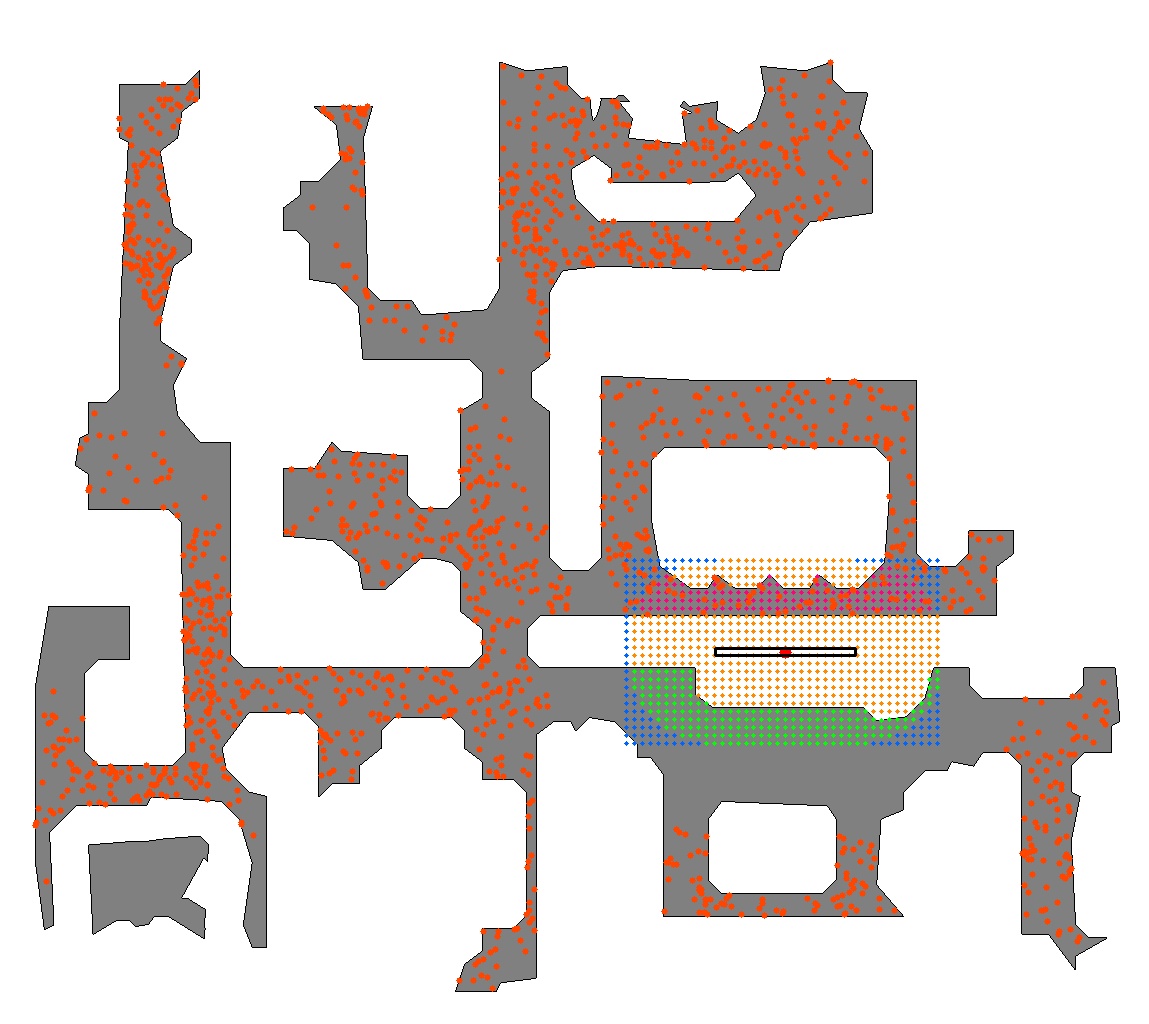}\\

\end{tabularx}
\caption{
\textbf{ObjectNav episode generation visualization.}
We show generated episode goal and starting positions for TVs in example scenes from \ourdataset and ProcTHOR. The goal object is outlined with a black box, the valid viewpoints are shown in green, and the episode start positions are shown with orange points.
}
\label{fig:episodes_viz}
\end{figure}

We present sample episode visualizations for two goal TV instances in \ourdataset and ProcTHOR through a top-down map in \Cref{fig:episodes_viz}. The bounding box of the goal TV instance is outlined with a black box and the viewpoints are shown in green (valid), blue (invalid due to being far from the object), and yellow (invalid due to being unnavigable) pixels. The orange pixels denote the episode starting positions.

Note that \ourdataset also has scene regions outside the house (e.g., backyards, gardens, balconies). We restrict all episodes to indoor regions as we focus on indoor-only navigation.
Both the goal object and episode start position are required to be inside the house, and doors leading to the exterior are closed.
A handful of scenes do not have a clear distinction between indoors and outdoors and are therefore excluded from episode generation.
For this reason, we generate training episodes for 122 scenes out of the 125 scenes in the training set.

\begin{figure}
\centering
\setkeys{Gin}{width=\linewidth}
\noindent
\begin{tabularx}{\linewidth}{@{}YY@{}}
Train & Validation \\
\includegraphics[]{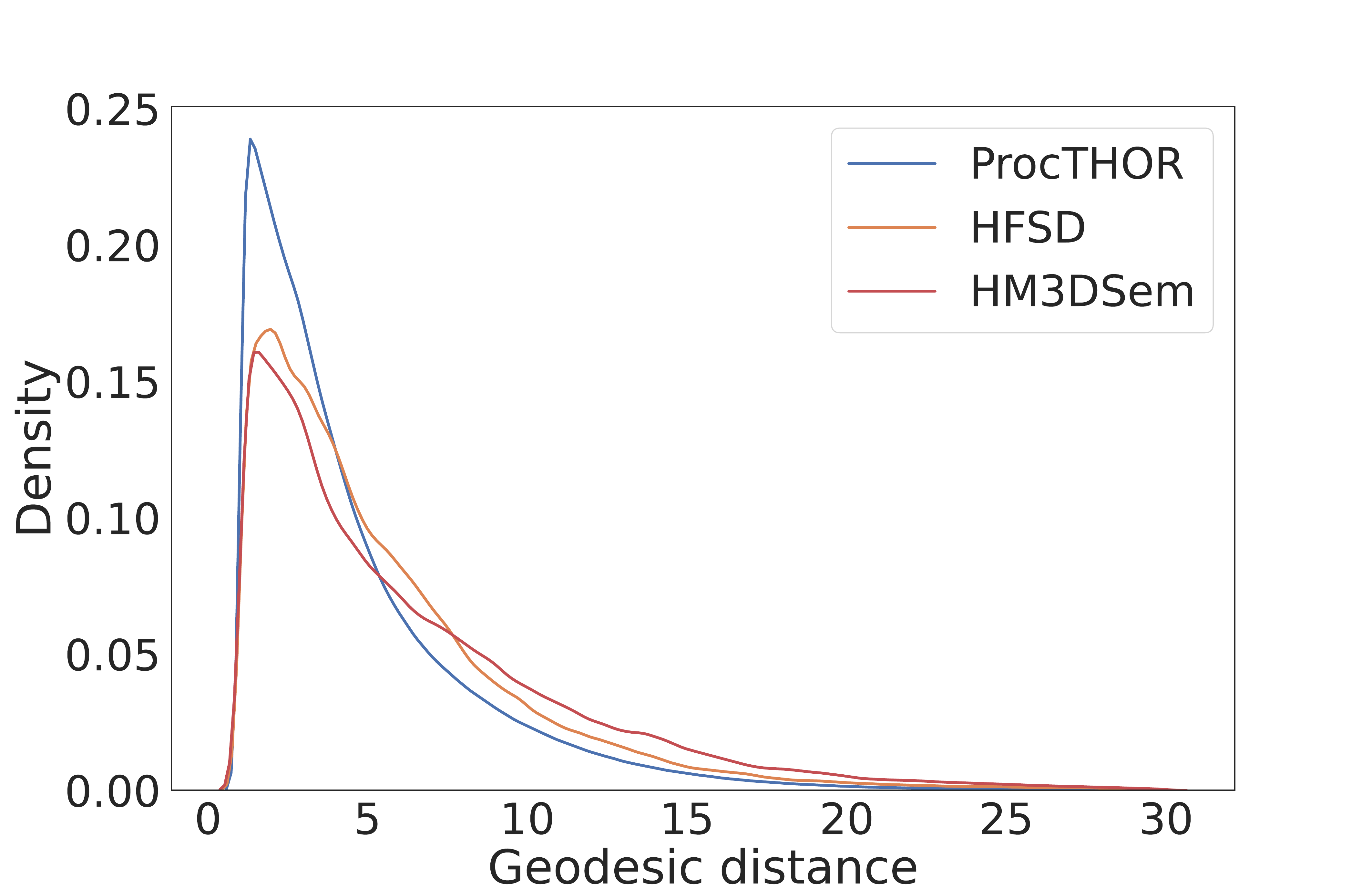} & \includegraphics[]{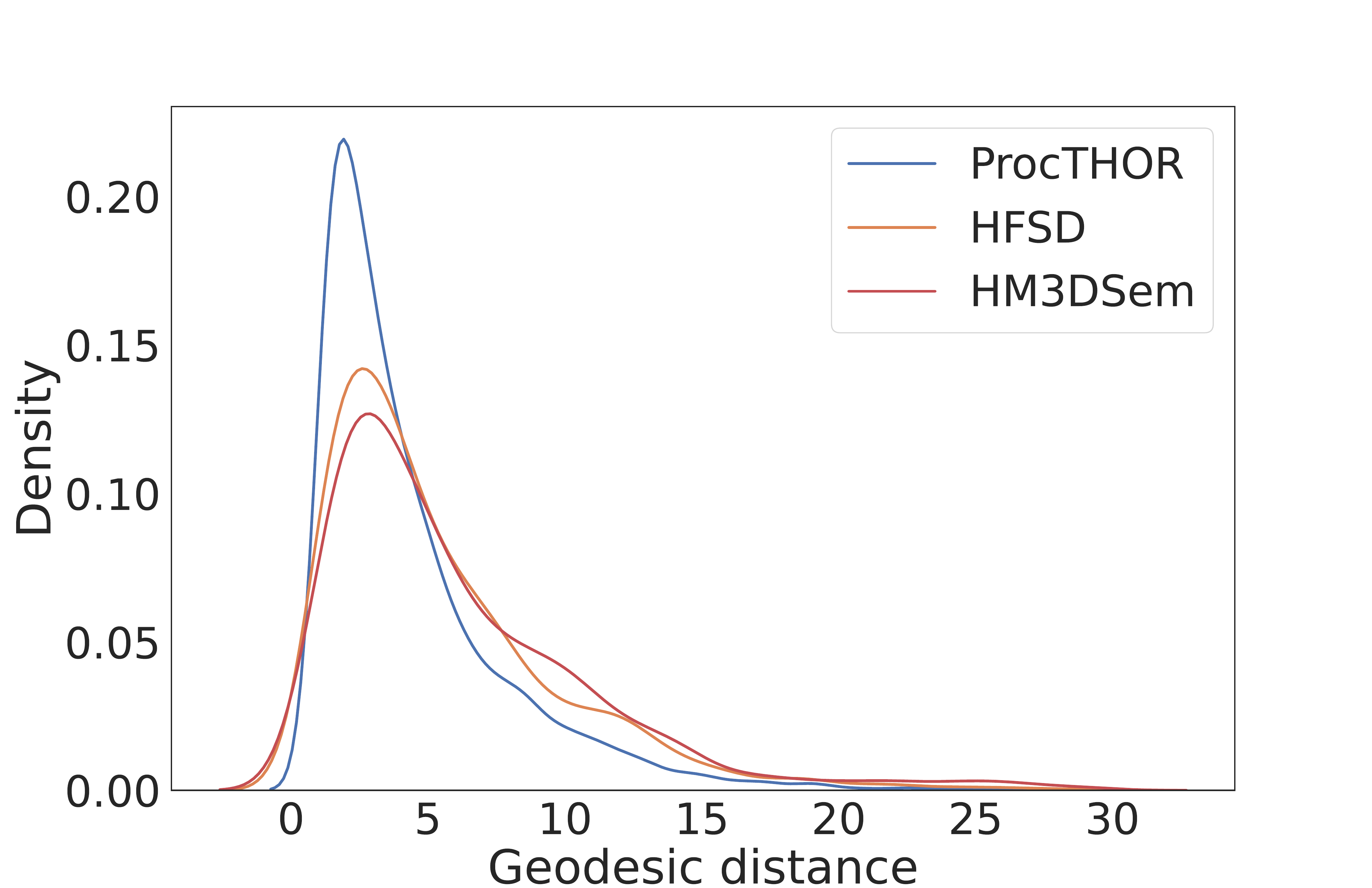}\\
\end{tabularx}
\caption{
\textbf{Geodesic distance distribution of episode datasets.}
We compare the distribution of geodesic distances across the episode datasets of ProcTHOR~\cite{deitke2022procthor}, \ourdataset, and HM3DSem~\cite{yadav2022hm3dsem}.
Note how the differences in scene size distributions lead to significantly higher numbers of (easier) low geodesic distance episodes in ProcTHOR compared to fairly similar distributions between \ourdataset and HM3DSem.
}
\label{fig:geod_dist}
\end{figure}

The inherent scene size distribution differences between ProcTHOR, \ourdataset, and HM3D are also reflected in the distributions of episode geodesic distance that emerge in episodes generated from each of these scene datasets. See \Cref{fig:geod_dist} for a comparison.
ProcTHOR has a high number of (easier) low geodesic distance episodes (due to a good number of small 1-3 room houses), with an exponential decay in the number of episodes as the distance increases (in bigger houses with more rooms). On the other hand, \ourdataset and HM3D have more similar distributions for both train and validation episode datasets.

\begin{figure}
\vspace{-0.5cm}
\includegraphics[width=\linewidth]{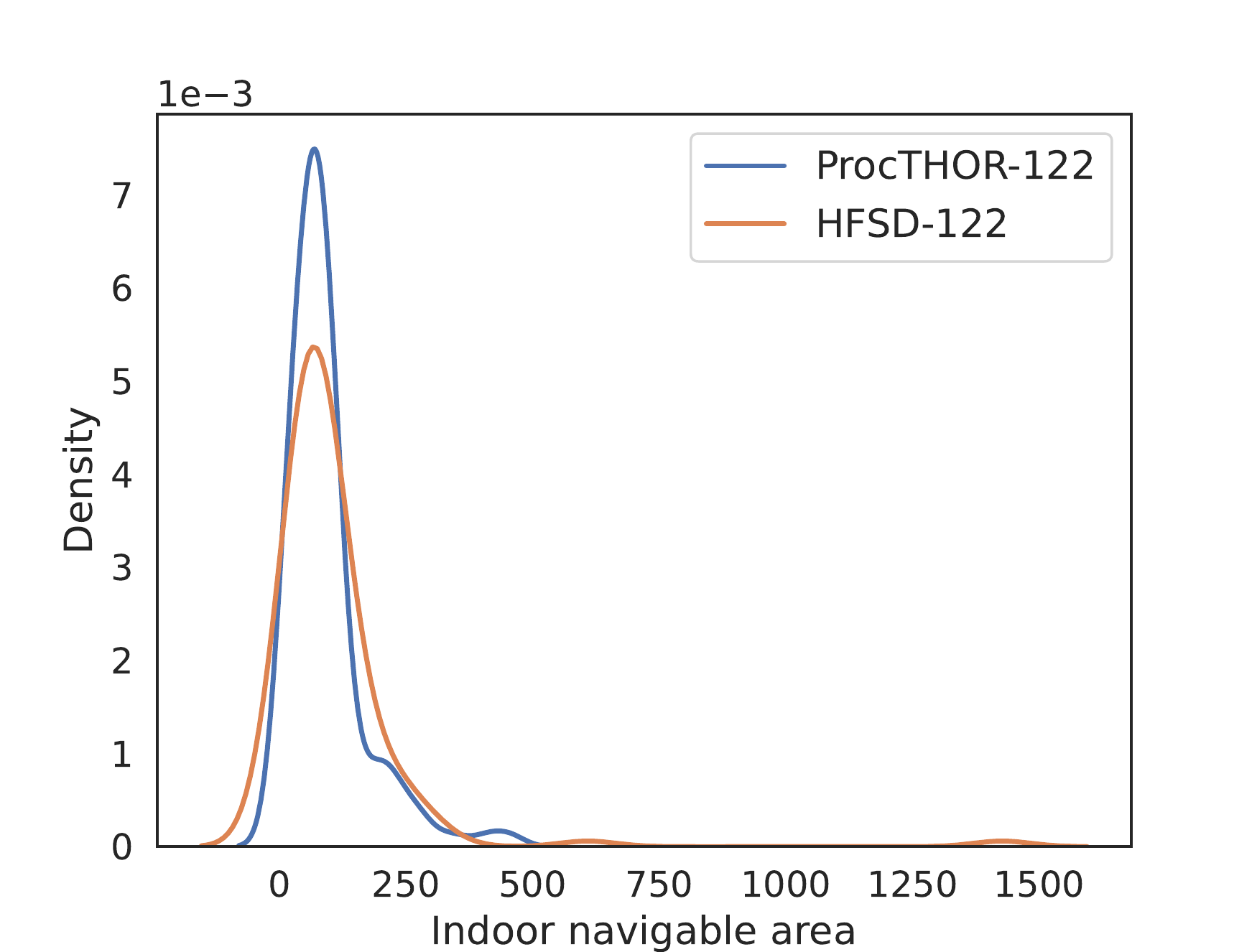}
\caption{
\update{\textbf{Navigable area distributions for comparable scale datasets.}}
We prepare a 122-scene subset of ProcTHOR-10K \cite{deitke2022procthor} that matches \ourdataset's training set in scale in terms of number of scenes and the navigable area distribution.
We refer to this scene dataset as ProcTHOR-122.
}
\label{fig:fp_vs_pt_122_scale_dist}
\end{figure}

\section{Analysis \& experiment details}
\label{sec:supp-expr-details}

\mypara{Hierarchical clustering algorithm details.}
Given the object co-occurrence matrix $C$ we obtain as described in the main paper, we first compute the dissimilarity matrix $D = 1 - C$.
Then, we compute the distance for each unique pair $D(i, j)$, constructing an $n * (n-1) / 2$-dimensional vector. %
This vector is then used for hierarchical clustering with the farthest point algorithm to compute the distance between clusters and output a linkage matrix.
We use SciPy's hierarchical clustering implementation to do this and form flat clusters.
Cutting into flat cluster requires a distance threshold (maximum distance between clusters).
We use a threshold $t = 0.8$ to compute the similarity scores reported in the main paper.

\paragraph{ProcTHOR-122.}
We disentangle scene dataset scale and scene dataset realism by creating a version of the ProcTHOR-10K dataset \cite{deitke2022procthor} that matches the scale of \ourdataset, as measured by number of scenes and navigable area.
We sample 122 scenes from ProcTHOR-10K, matching the navigable area distribution of \ourdataset's training dataset as closely as possible.
The navigable area distributions of these scale-matched scene datasets are in \Cref{fig:fp_vs_pt_122_scale_dist}.

\paragraph{\ourdataset-60 and ProcTHOR-60.}
To measure the impact of scene dataset scale, we also create variants of \ourdataset and ProcTHOR-122 with 60 scenes.
We do this by randomly sampling 60 scenes out of 122.
We refer to these scene dataset variants as \ourdataset-60 and ProcTHOR-60.

\update{\section{Training plots and finetuning results}}
\label{sec:train-and-ft-results}

\begin{figure}
\centering
\setkeys{Gin}{width=\linewidth}
\noindent
\begin{tabularx}{\linewidth}{@{}YYYY@{}}
\includegraphics[]{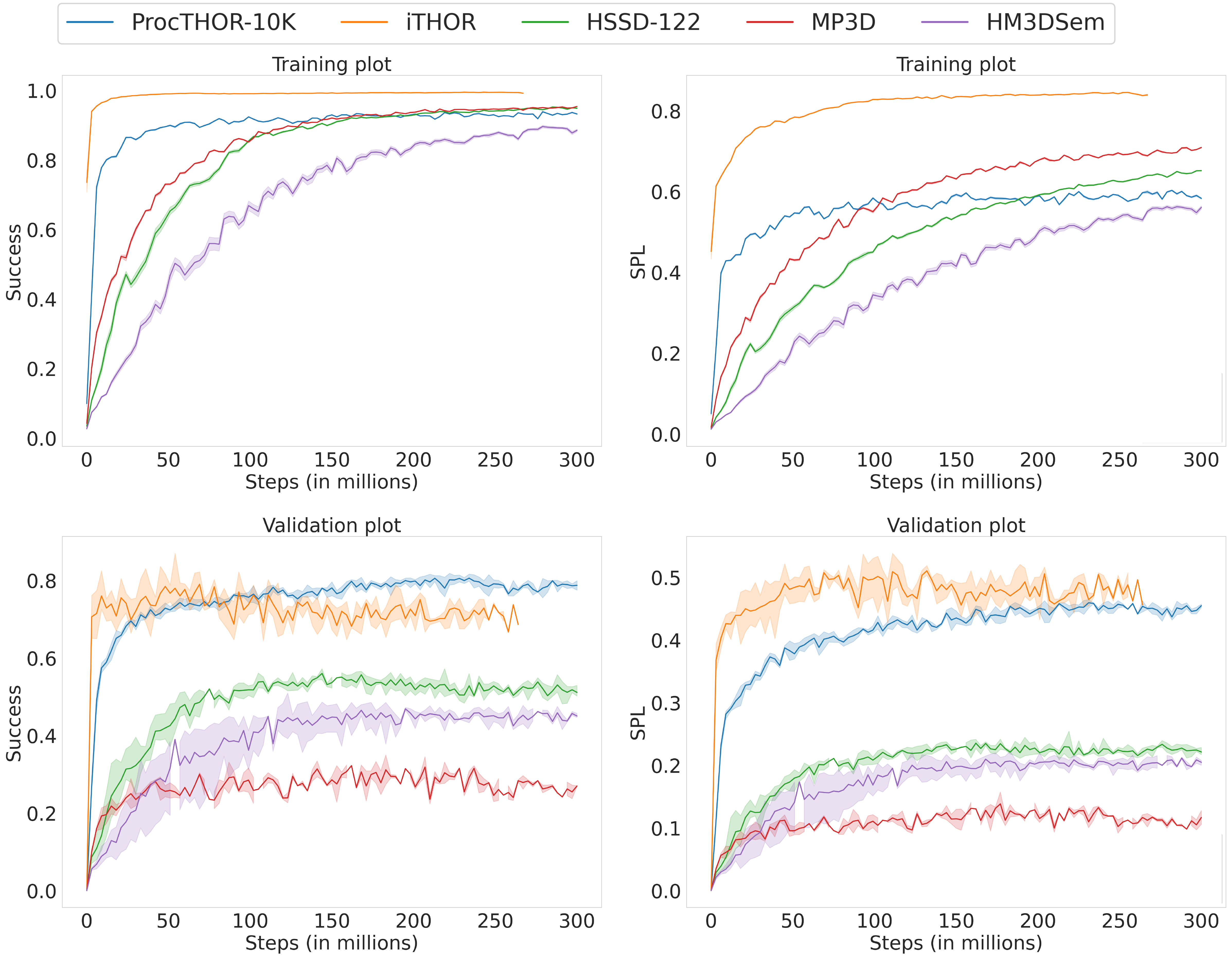}

\end{tabularx}
\caption{\update{
\textbf{Agent training plots.}
We provide plots of agent performance during training on each dataset's training set and validation set.
These plots correspond to the zero-shot agents presented in \Cref{tab:zero_6_28} of the main paper.
Each agent is trained on the indicated dataset (iTHOR, ProcTHOR, \ourdataset, HM3DSem, and MP3D) to convergence.
The plots show results from three independent training runs.
Validation set performance for all agents saturates by approximately 200M steps of experience.
Note that agents differ on when they reach convergence, with iTHOR agents doing so significantly faster likely due to the simplicity of the one-room scenes in the dataset.
}}
\label{fig:training_curves}
\end{figure}

\begin{figure}
\centering
\setkeys{Gin}{width=\linewidth}
\noindent
\begin{tabularx}{\linewidth}{@{}YYYY@{}}
\includegraphics[]{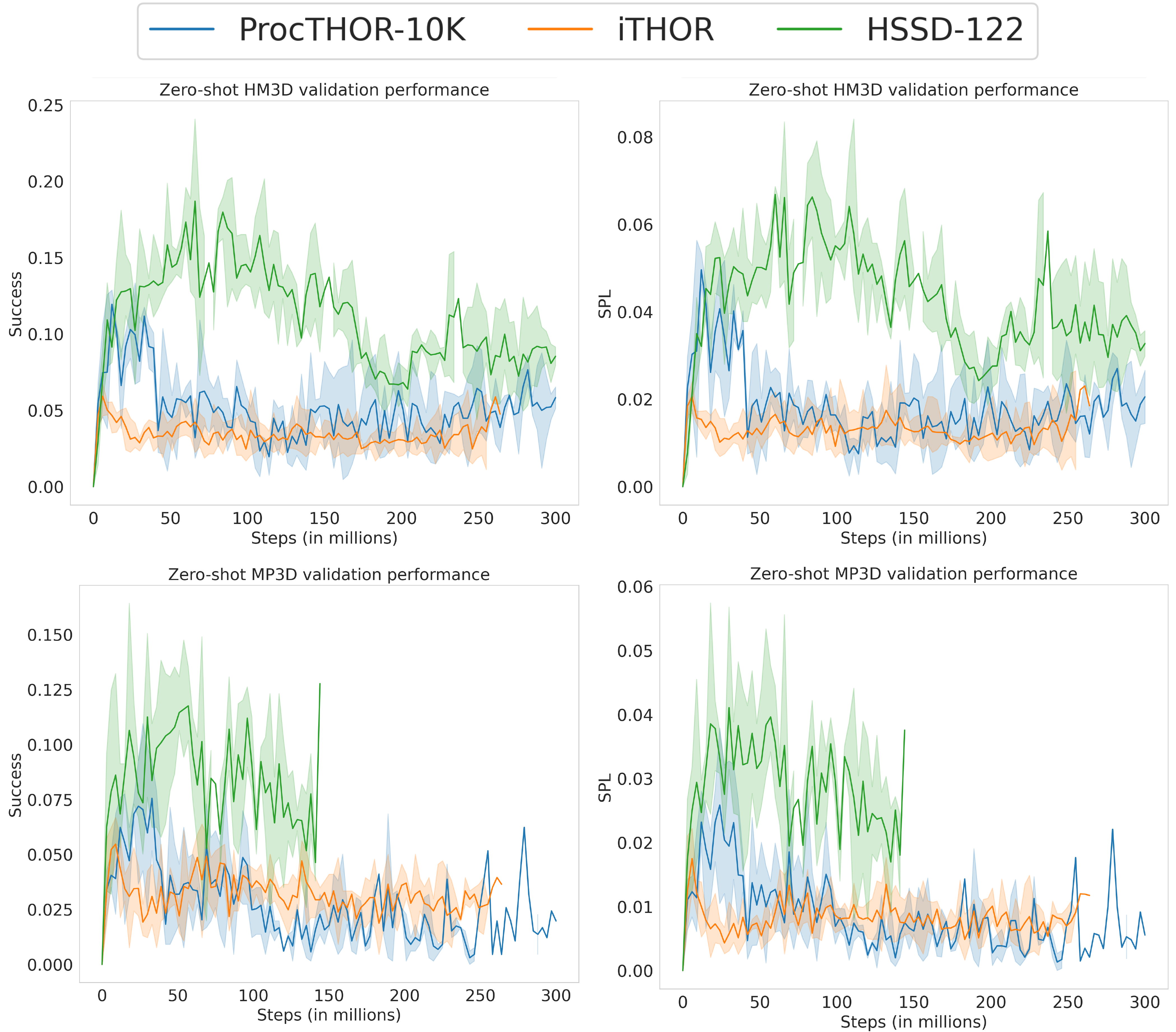}

\end{tabularx}
\caption{\update{
\textbf{Zero-shot evaluation on HM3DSem and MP3D.}
We plot zero-shot success and SPL on HM3DSem (top row) and MP3D (bottom row) validation scenes for agents pretrained on each synthetic scene dataset across number of training steps. Each line summarizes performance for agent checkpoints from three independent training runs evaluated zero-shot on HM3DSem and MP3D. We see that \ourdataset-pretrained agents perform better throughout compared to both ProcTHOR and iTHOR-pretrained agents. 
}}
\label{fig:zero-shot_eval}
\end{figure}

\update{
\paragraph{Training and evaluation plots.}
In the main paper we reported zero-shot performance of agents trained on different datasets in \Cref{tab:zero_6_28}.
Here, we present the agent training plots as well as validation set performance plots during training (on the same dataset's validation set).
\Cref{fig:training_curves} shows these plots.
All agents reach validation set convergence by approximately 200M steps of experience.
The results in the main paper use the agent checkpoint with highest validation set SPL from each training run.
We also plot zero-shot performance of agents on HM3DSem and MP3D validation datasets across number of training steps in \Cref{fig:zero-shot_eval}.
These plots show that overall \ourdataset-pretrained agents generalize better to real-world 3D scanned scenes than ProcTHOR and iTHOR-pretrained agents.
}

\update{
\paragraph{Finetuning results.}
In addition to the zero shot generalization experiments which were the focus of our work, we also present experiments with agents finetuned on the target dataset.
The agents are pre-trained on variants of the \ourdataset and ProcTHOR training datasets and then finetuned on the HM3DSem training set.
Specifically, for each agent, we finetune the agent checkpoint that has the best zero-shot performance on the HM3DSem validation set in terms of the SPL metric.
See \Cref{tab:finetuning_results} for a summary of the results.
In the table we compare these finetuned agents against the performance of an agent trained directly on HM3DSem.
We find that all agents converge to similar levels of performance after finetuning, irrespective of the pre-training dataset.
Performance in terms of the success metric ranges between 47.85 and 48.48, while the combined success and efficiency performance as measured by SPL ranges between 22.16 and 23.1, with HSSD agents retaining a lead over ProcTHOR agents.
This trend is not surprising as finetuning on the target dataset is expected to reduce performance gaps due to differences between the original training datasets.
}

\begin{table}
    \centering
    \begin{tabular}{@{} lcc @{}}
        \toprule
        Pre-training dataset & Success $\uparrow$ & SPL $\uparrow$ \\
        \midrule
        HM3DSem & 48.10 & 22.16 \\
        \midrule
        ProcTHOR-60 & 47.85 & 22.37 \\
        \ourdataset-60 & \textbf{48.13} & \textbf{22.76} \\
        \midrule
        ProcTHOR-122 & \textbf{48.48} & 22.79 \\
        \ourdataset-122 & 48.23 & \textbf{23.10} \\
        ProcTHOR-10K & 48.32 & 21.80 \\
        \bottomrule
    \end{tabular}
    \caption{
    \update{
    \textbf{Finetuned agent performance.}
    Performance of \ourdataset and ProcTHOR pre-trained agents on HM3DSem validation set scenes after finetuning on the HM3DSem training set scenes.
    All agents converge to comparable performance, though HSSD agents retain a small lead in combined success and efficiency (SPL).}
    }
    \label{tab:finetuning_results}
\end{table}

\begin{figure}
    \centering
    \includegraphics[width=\linewidth]{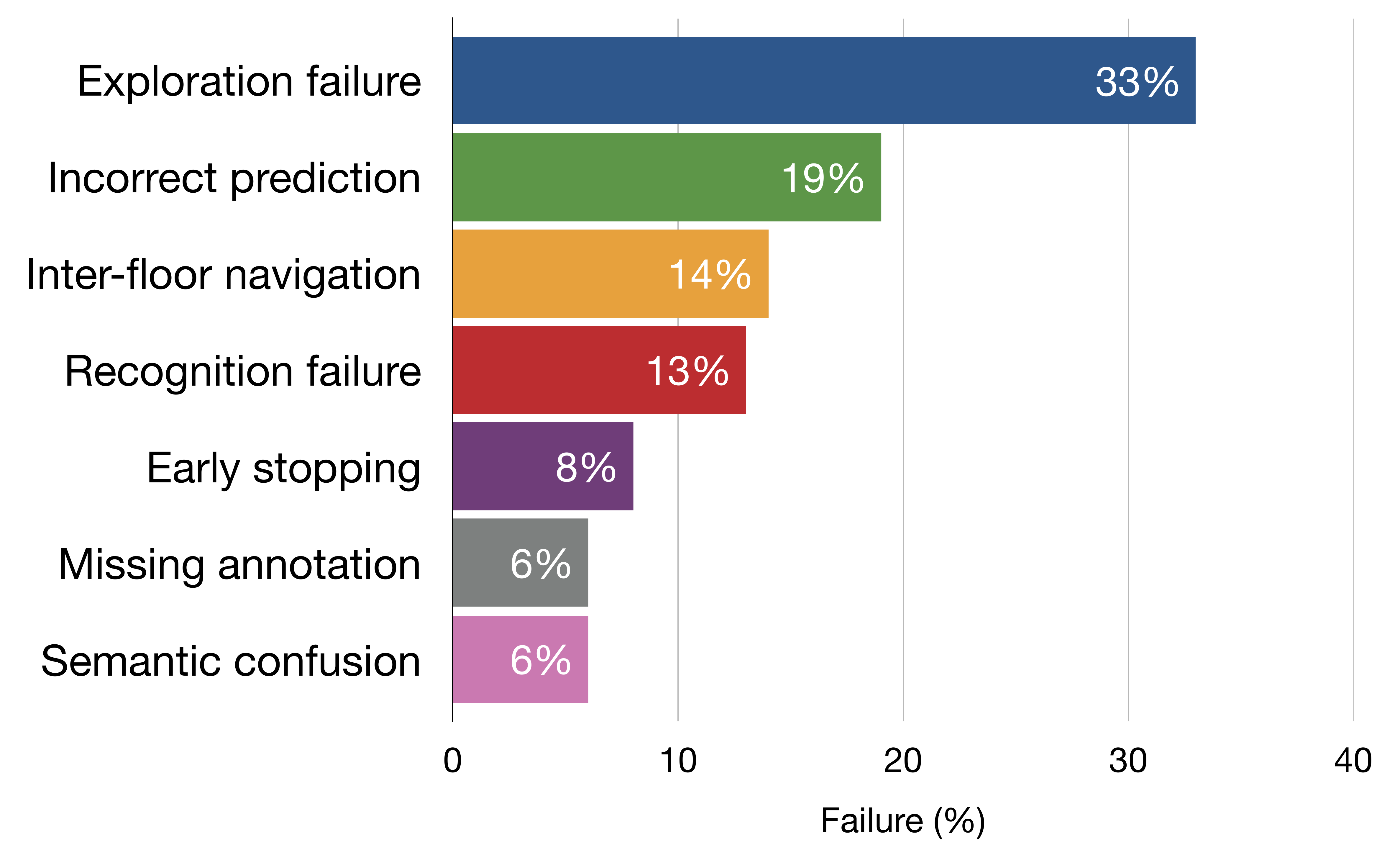}
    \caption{
    \update{\textbf{Failure case analysis.}}
    Breakdown of 100 randomly selected failure cases for agent pre-trained on \ourdataset, fine-tuned on the HM3DSem~\cite{yadav2022hm3dsem} train set, and evaluated on the HM3DSem val set. We see that agents are most likely to fail due to inefficacy in exploring the scene (i.e. exploration failures).}
    \label{fig:failure_analysis}
\end{figure}

\section{Agent failure case analysis}
\label{sec:supp-expr-error}

Inspired by \citet{ramrakhya2023pirlnav}, we analyze common failure cases when evaluating the \ourdataset-trained agent on the HM3DSem~\cite{yadav2022hm3dsem} val set (after fine-tuning on the HM3DSem train set).
We randomly sample 100 val set episodes where the agent failed to succeed and analyze the modes of failure by classifying the agent's performance into the following failure cases:

\mypara{Exploration failure} (33\%): agent does not explore some part of the house and therefore fails to come across the goal object. A common cause is excessive looping behavior in one part of the house, i.e. repeatedly visiting the same region.

\mypara{Incorrect prediction} (19\%): agent stops in front of an object that is not the goal (e.g. stopping in front of a green toy when the goal was a plant).

\mypara{Inter-floor navigation} (14\%): agent is spawned on a floor that does not have any goal instances. It needs to change floors to find the object.

\mypara{Recognition failure} (13\%): agent sees the object clearly when exploring, but does not navigate to it.

\mypara{Early stopping} (8\%): agent finds the object but stops a few centimeters too far from it.

\mypara{Missing annotation} (6\%): agent navigates to a valid goal object that unfortunately has not been annotated in the HM3DSem scene, causing the episode to be deemed unsuccessful.

\mypara{Semantic confusion} (6\%): agent navigates to an incorrect but semantically similar object category (e.g. navigating to an armchair instead of a sofa).

We plot the corresponding distribution of failure cases in \Cref{fig:failure_analysis}.
A major cause of failure is inability to effectively explore the scene.
Agents are likely to show better performance if annotations are improved and if objects can be found on the same floor.


\end{document}